\pdfoutput=1
\documentclass[sn-mathphys-num,iicol]{sn-jnl}
\usepackage{natbib}
\setcitestyle{numbers,square}
\usepackage{graphicx}
\usepackage{multirow}
\usepackage{amsmath,amssymb,amsfonts}
\usepackage{amsthm}
\usepackage{mathrsfs}
\usepackage[table]{xcolor}
\usepackage{manyfoot}
\usepackage{booktabs}
\usepackage{algorithm}
\usepackage{algpseudocode}
\usepackage{algorithmicx}
\usepackage{algpseudocode}
\usepackage{listings}
\usepackage{url}
\usepackage{mathtools}
\usepackage{balance}
\usepackage{makecell}
\usepackage{epsfig}
\usepackage{subcaption}
\usepackage{pifont}
\usepackage{arydshln}
\usepackage{hyperref}
\makeatletter
\g@addto@macro{\UrlBreaks}{\UrlOrds}
\makeatother
\hypersetup{
    colorlinks=true,
    linkcolor=teal,     
    citecolor=violet,    
    urlcolor=magenta     
}

\definecolor{verylightgray}{gray}{0.93}
\colorlet{mygreen}{green!50!black}

\begin{document}

\title[Article Title]{Light-VQA+: A Video Quality Assessment Model for Exposure Correction with Vision-Language Guidance}

\author[1]{\fnm{Xunchu} \sur{Zhou}}\email{zhou\_xc@sjtu.edu.cn}
\equalcont{These authors contributed equally to this work.}

\author*[1]{\fnm{Xiaohong} \sur{Liu}}\email{xiaohongliu@sjtu.edu.cn}
\equalcont{These authors contributed equally to this work.}

\author[1]{\fnm{Yunlong} \sur{Dong}}\email{dongyunlong@sjtu.edu.cn}

\author[1]{\fnm{Tengchuan} \sur{Kou}}\email{2213889087@sjtu.edu.cn}

\author[1]{\fnm{Yixuan} \sur{Gao}}\email{gaoyixuan@sjtu.edu.cn}

\author[1]{\fnm{Zicheng} \sur{Zhang}}\email{zzc1998@sjtu.edu.cn}

\author[1]{\fnm{Chunyi} \sur{Li}}\email{lcysyzxdxc@sjtu.edu.cn}

\author[2]{\fnm{Haoning} \sur{Wu}}\email{haoning001@e.ntu.edu.sg}

\author[1]{\fnm{Guangtao} \sur{Zhai}}\email{zhaiguangtao@sjtu.edu.cn}

\affil[1]{\orgname{Shanghai Jiao Tong University}, \orgaddress{Shanghai 200240}, \city{Shanghai}, \country{China}}

\affil[2]{\orgname{Nanyang Technological University}, \orgaddress{Singapore 6339798}, \country{Singapore}}

\abstract{
Recently, User-Generated Content (UGC) videos have gained popularity in our daily lives. However, UGC videos often suffer from poor exposure due to the limitations of photographic equipment and techniques. Therefore, Video Exposure Correction (VEC) algorithms have been proposed, Low-Light Video Enhancement (LLVE) and Over-Exposed Video Recovery (OEVR) included. Equally important to the VEC is the Video Quality Assessment (VQA). Unfortunately, almost all existing VQA models are built generally, measuring the quality of a video from a comprehensive perspective. As a result, Light-VQA, trained on LLVE-QA, is proposed for assessing LLVE. We extend the work of Light-VQA by expanding the LLVE-QA dataset into Video Exposure Correction Quality Assessment (\textbf{VEC-QA}) dataset with over-exposed videos and their corresponding corrected versions. In addition, we propose \textbf{Light-VQA+}, a VQA model specialized in assessing VEC. Light-VQA+ differs from Light-VQA mainly from the usage of the CLIP model and the vision-language guidance during the feature extraction, {followed by a new module referring to the Human Visual System (HVS) for more accurate assessment.} Extensive experimental results show that our model achieves the best performance against the current State-Of-The-Art (SOTA) VQA models on the VEC-QA dataset and other public datasets. Our code and dataset can be found at \url{https://github.com/SaMMyCHoo/Light-VQA-plus}.
}


\maketitle

\section{Introduction}

\begin{table*}[t]
\renewcommand{\arraystretch}{1.2}
\caption{
\textcolor{brown}{Which video enjoys the best visual perceptual quality in the listed examples?} The below 12 figures are the representative frames of 12 corrected videos obtained by applying different correction algorithms to corresponding original over-exposed and low-light videos. The concrete algorithms are listed below the figures. Then we use 5 SOTA VQA models (Simple-VQA~\cite{vqasimple}, FAST-VQA~\cite{vqafastvqa}, Max-VQA~\cite{vqamaxvqa}, Q-Align~\cite{vqaqalign}, Light-VQA~\cite{baselightvqa}) and the proposed Light-VQA+ to predict the quality of these videos. The Ground-Truth (GT) perceptual quality of enhanced videos, are obtained through a subjective experiment. It is evident that the results of Light-VQA+ are highly consistent with human perception as compared to others. [Key: \textcolor{mygreen}{\textbf{video with the best perceptual quality given by models or GT}}]
}
\label{intro}
\resizebox{\linewidth}{!}{
\begin{tabular}{c ccc c ccc||c ccc c ccc}
\toprule[1.5pt]
\multicolumn{8}{c||}{\textbf{Over-Exposed-Recovered Videos}} & \multicolumn{8}{c}{\textbf{Low-Light-Enhanced Videos}}
\\ \midrule[1pt]
\multicolumn{8}{c||}{
    \begin{tabular}{ccc}
        {
        \begin{minipage}[b]{0.33\columnwidth}
        \centering
        \caption*{(\uppercase\expandafter{\romannumeral1})}
		\includegraphics[width=1\linewidth]{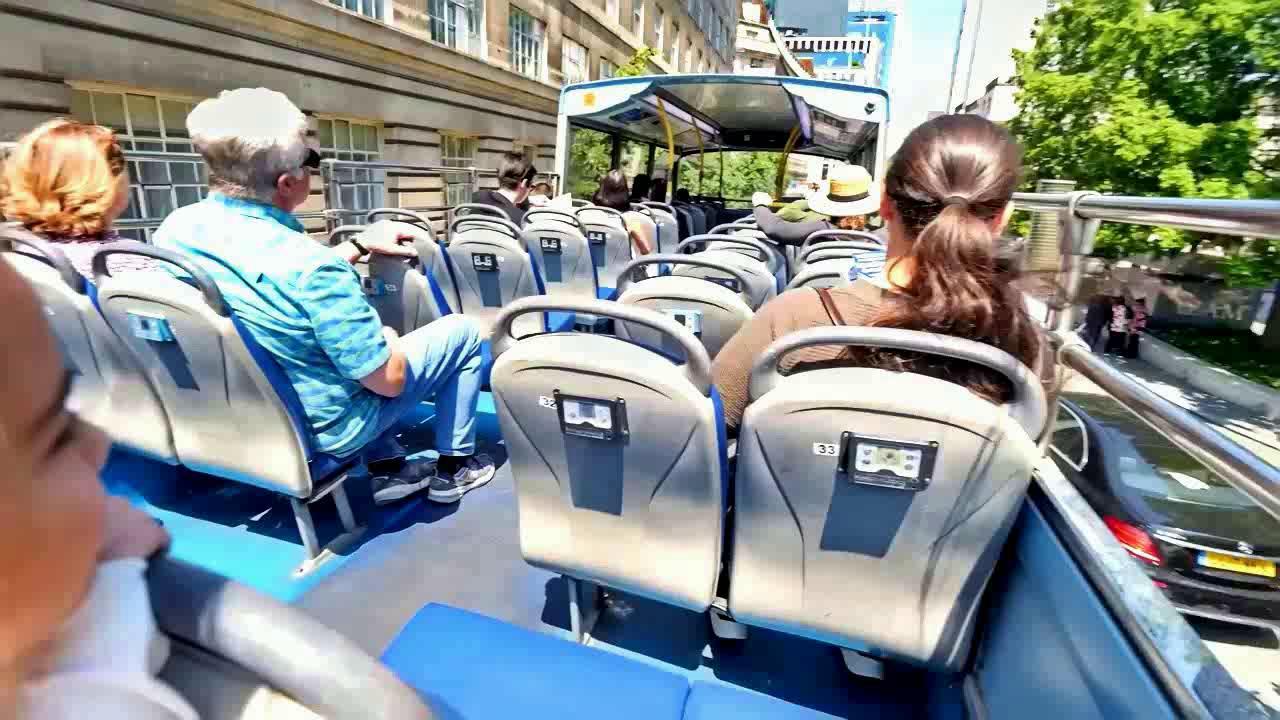}
        \caption*{DIEREC~\cite{algoDIEREC}}
        \end{minipage}
        }&{
        \begin{minipage}[b]{0.33\columnwidth}
        \centering
        \caption*{(\uppercase\expandafter{\romannumeral2})}
		\includegraphics[width=1\linewidth]{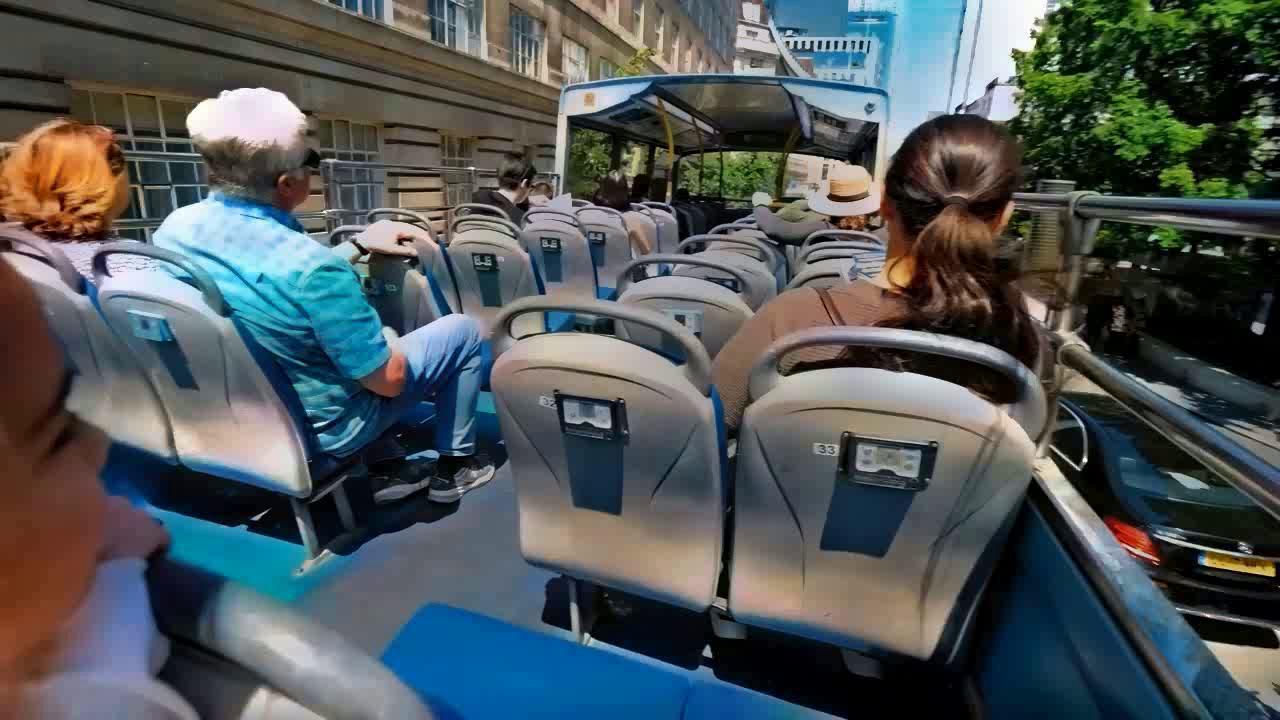}
        \caption*{LMSPEC~\cite{algoLMSPEC}}
        \end{minipage}
        }&{
        \begin{minipage}[b]{0.33\columnwidth}
        \centering
        \caption*{(\uppercase\expandafter{\romannumeral3})}
		\includegraphics[width=1\linewidth]{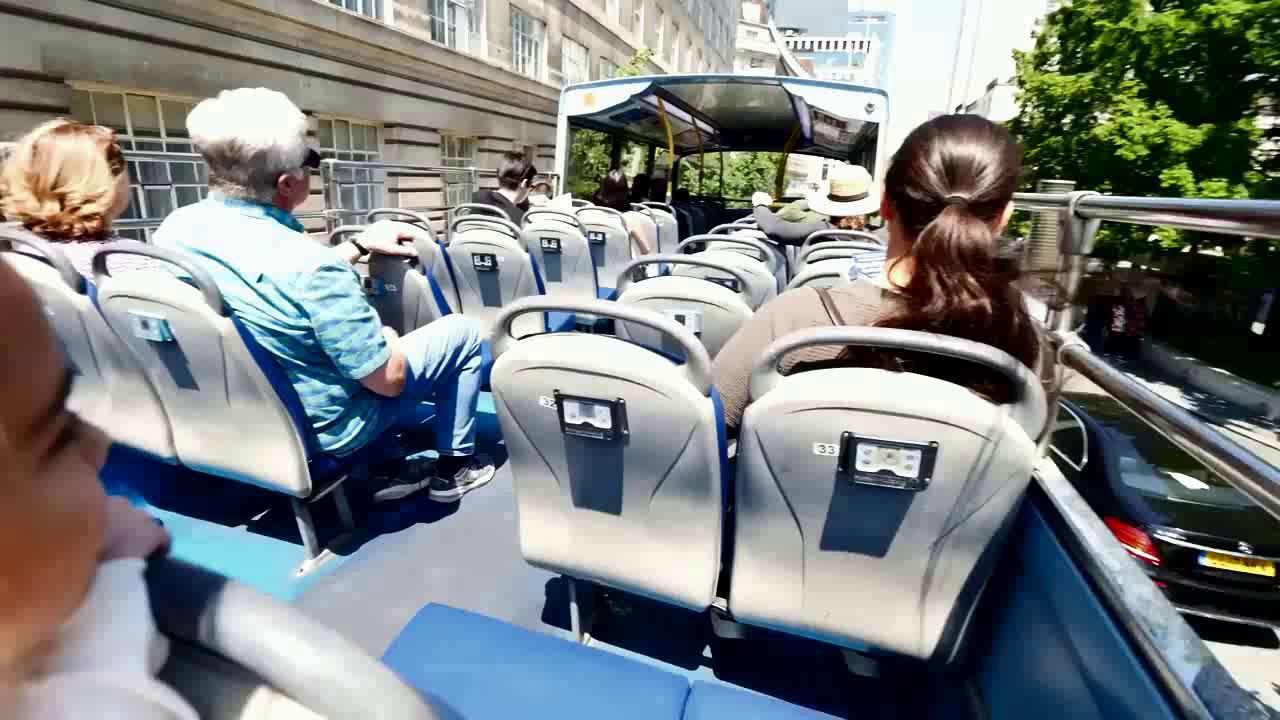}
        \caption*{ECMEIQ~\cite{algoECMEIQ}}
        \end{minipage}
        }
    \end{tabular}
}& \multicolumn{8}{c}{
    \begin{tabular}{ccc}
        {
        \begin{minipage}[b]{0.33\columnwidth}
        \centering
        \caption*{(\uppercase\expandafter{\romannumeral1})}
		\includegraphics[width=1\linewidth]{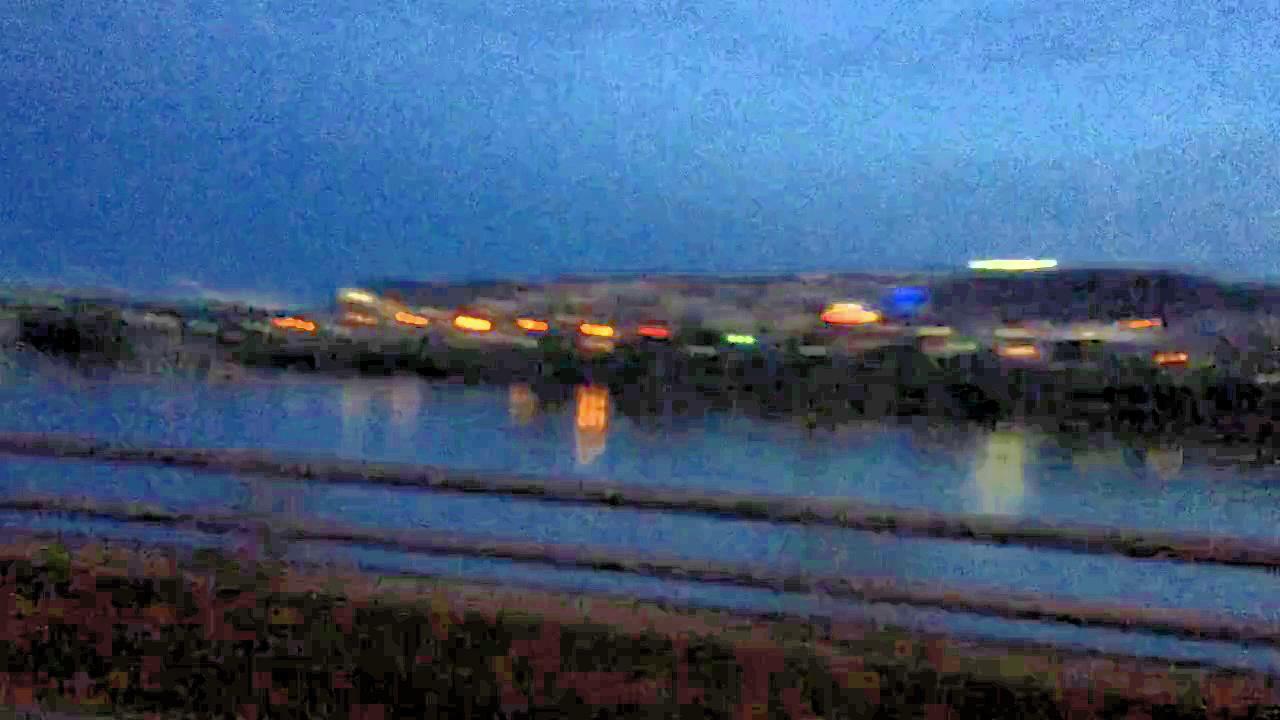}
        \caption*{AGCCPF~\cite{algoAGCCPF}}
        \end{minipage}
        }&{
        \begin{minipage}[b]{0.33\columnwidth}
        \centering
        \caption*{(\uppercase\expandafter{\romannumeral2})}
		\includegraphics[width=1\linewidth]{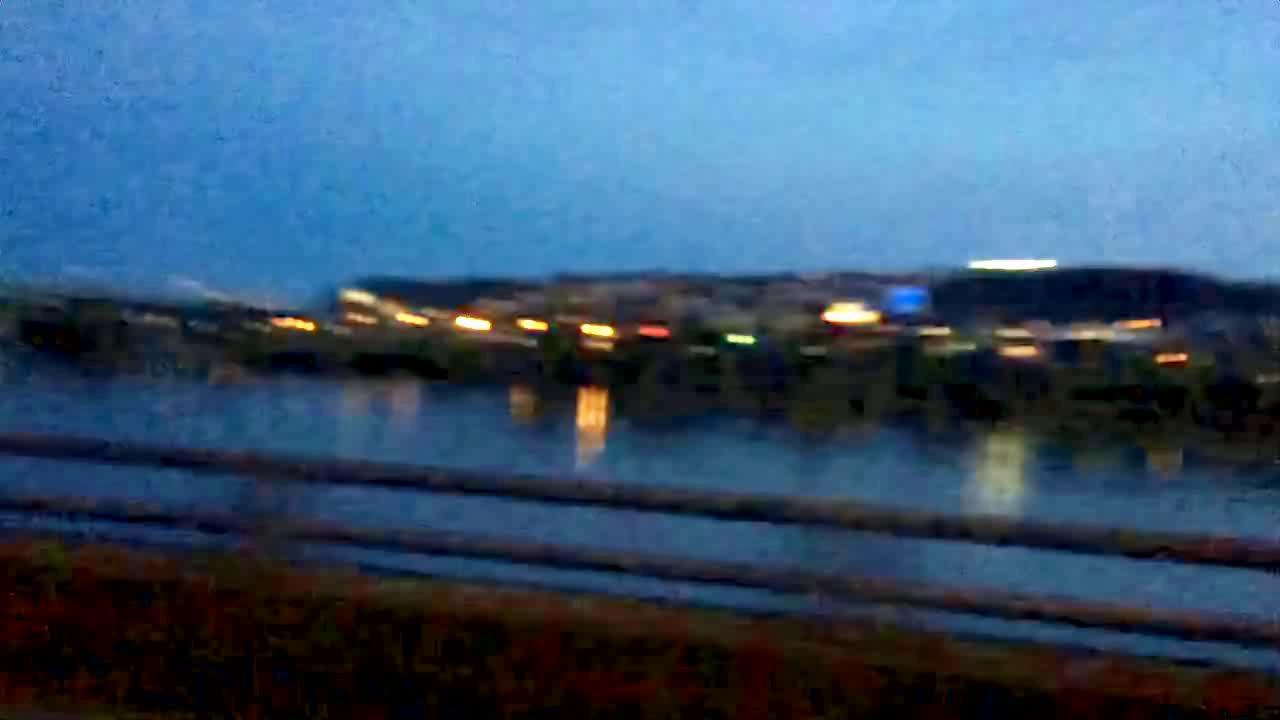}
        \caption*{MBLLEN~\cite{algoMBLLEN}}
        \end{minipage}
        }&{
        \begin{minipage}[b]{0.33\columnwidth}
        \centering
        \caption*{(\uppercase\expandafter{\romannumeral3})}
		\includegraphics[width=1\linewidth]{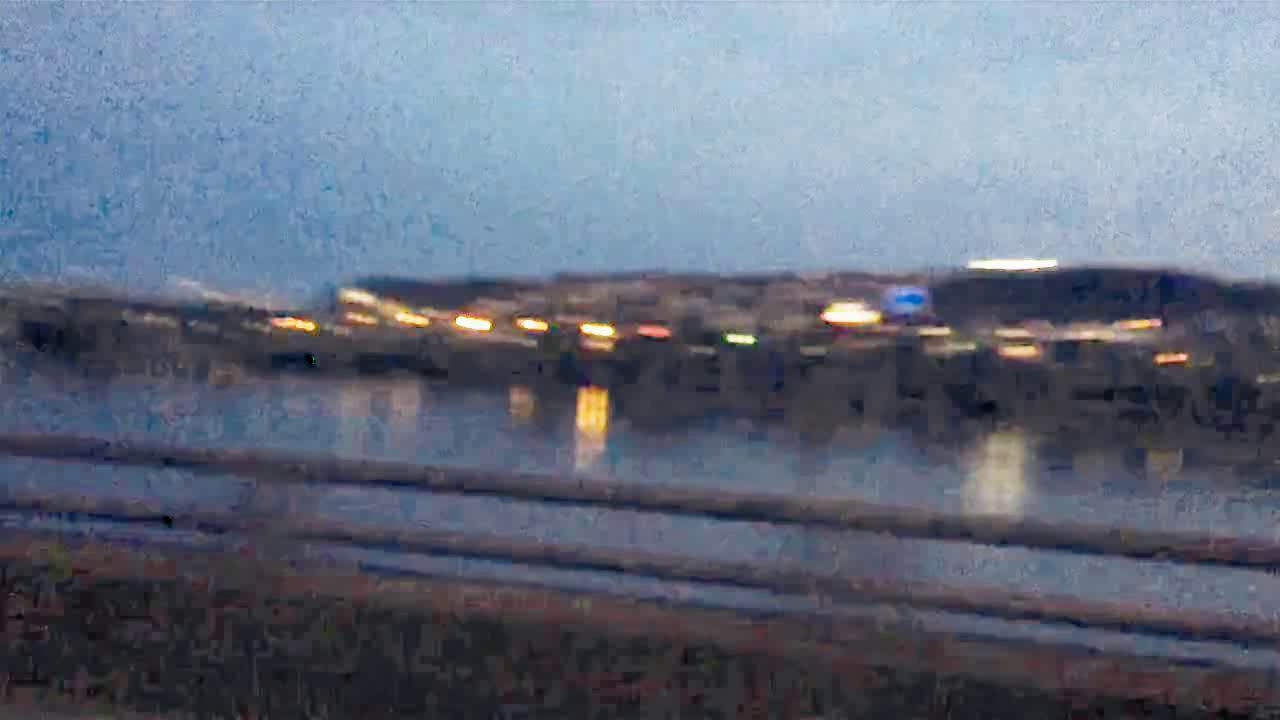}
        \caption*{DCC-Net~\cite{algoDCCNET}}
        \end{minipage}
        }
    \end{tabular}
} \\ 
\hline
 Ground-truth & {\uppercase\expandafter{\romannumeral1}} &\textcolor{mygreen}{\uppercase\expandafter{\romannumeral2}} & {\uppercase\expandafter{\romannumeral3}} & 
 Ground-truth & {\uppercase\expandafter{\romannumeral1}} &\textcolor{mygreen}{\uppercase\expandafter{\romannumeral2}} & {\uppercase\expandafter{\romannumeral3}} & 
 Ground-truth & {\uppercase\expandafter{\romannumeral1}} &\textcolor{mygreen}{\uppercase\expandafter{\romannumeral2}} & {\uppercase\expandafter{\romannumeral3}} & 
 Ground-truth & {\uppercase\expandafter{\romannumeral1}} &\textcolor{mygreen}{\uppercase\expandafter{\romannumeral2}} & {\uppercase\expandafter{\romannumeral3}} \\
\hdashline
 Simple-VQA~\cite{vqasimple}&{\textcolor{mygreen}{\ding{52}}}&{}&{}&
 Q-Align~\cite{vqaqalign}&&{\textcolor{mygreen}{\ding{52}}}&{}&
 Simple-VQA~\cite{vqasimple}&\textcolor{mygreen}{\ding{52}}&{}&{}&
 Q-Align~\cite{vqaqalign}&\textcolor{mygreen}{\ding{52}}&{}&{}\\
 
 Fast-VQA~\cite{vqafastvqa}&{\textcolor{mygreen}{\ding{52}}}&{}&{}&
 Light-VQA\cite{baselightvqa}&{}&{}&{\textcolor{mygreen}{\ding{52}}}&
 Fast-VQA~\cite{vqafastvqa}&\textcolor{mygreen}{\ding{52}}&{}&{}&
 Light-VQA\cite{baselightvqa}&{}&{}&\textcolor{mygreen}{\ding{52}}\\
 
 Max-VQA~\cite{vqamaxvqa}&{}&{}&\textcolor{mygreen}{\ding{52}}&
 \textbf{Light-VQA+}&{}&{\textcolor{mygreen}{\ding{52}}}&{}&
 Max-VQA~\cite{vqamaxvqa}&\textcolor{mygreen}{\ding{52}}&{}&{}&
 \textbf{Light-VQA+}&{}&{\textcolor{mygreen}{\ding{52}}}&\\

\midrule[1pt]
\multicolumn{8}{c||}{
    \begin{tabular}{ccc}
        {
        \begin{minipage}[b]{0.33\columnwidth}
        \centering
        \caption*{(\uppercase\expandafter{\romannumeral1}) }
		\includegraphics[width=1\linewidth]{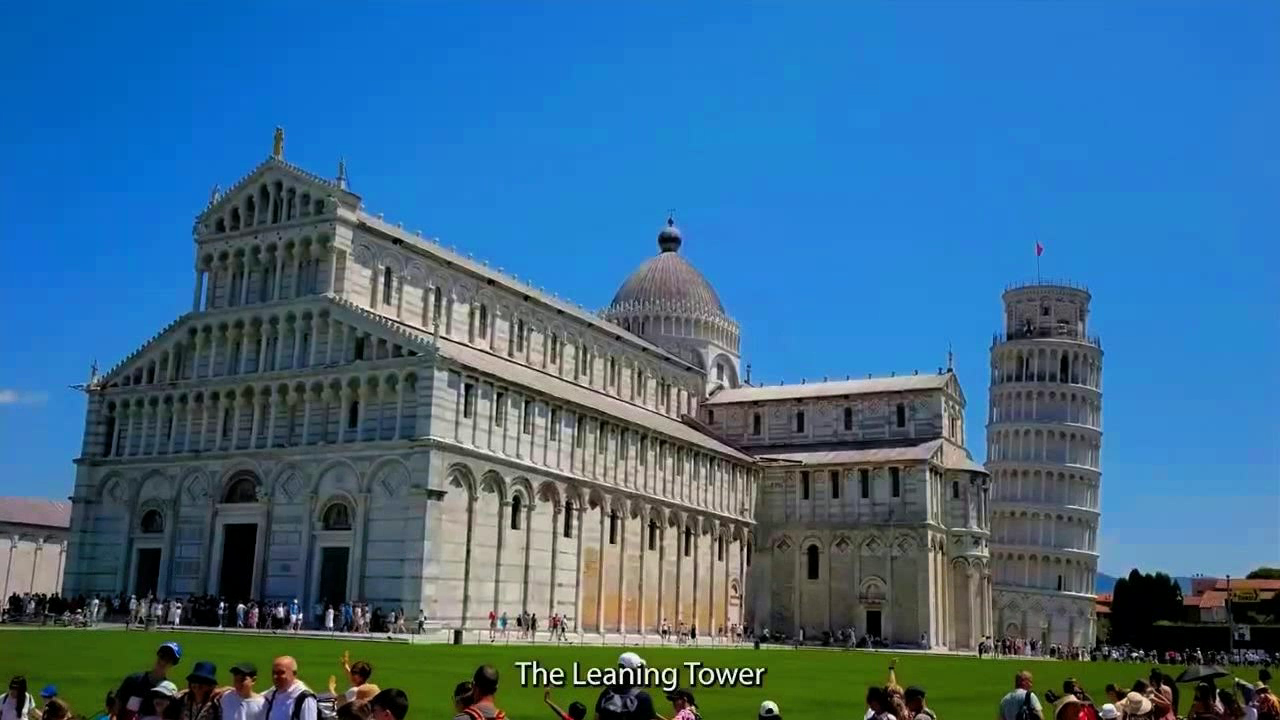}
        \caption*{LECVCM~\cite{algoLECVCM}}
        \end{minipage}
        }&{
        \begin{minipage}[b]{0.33\columnwidth}
        \centering
        \caption*{(\uppercase\expandafter{\romannumeral2})}
		\includegraphics[width=1\linewidth]{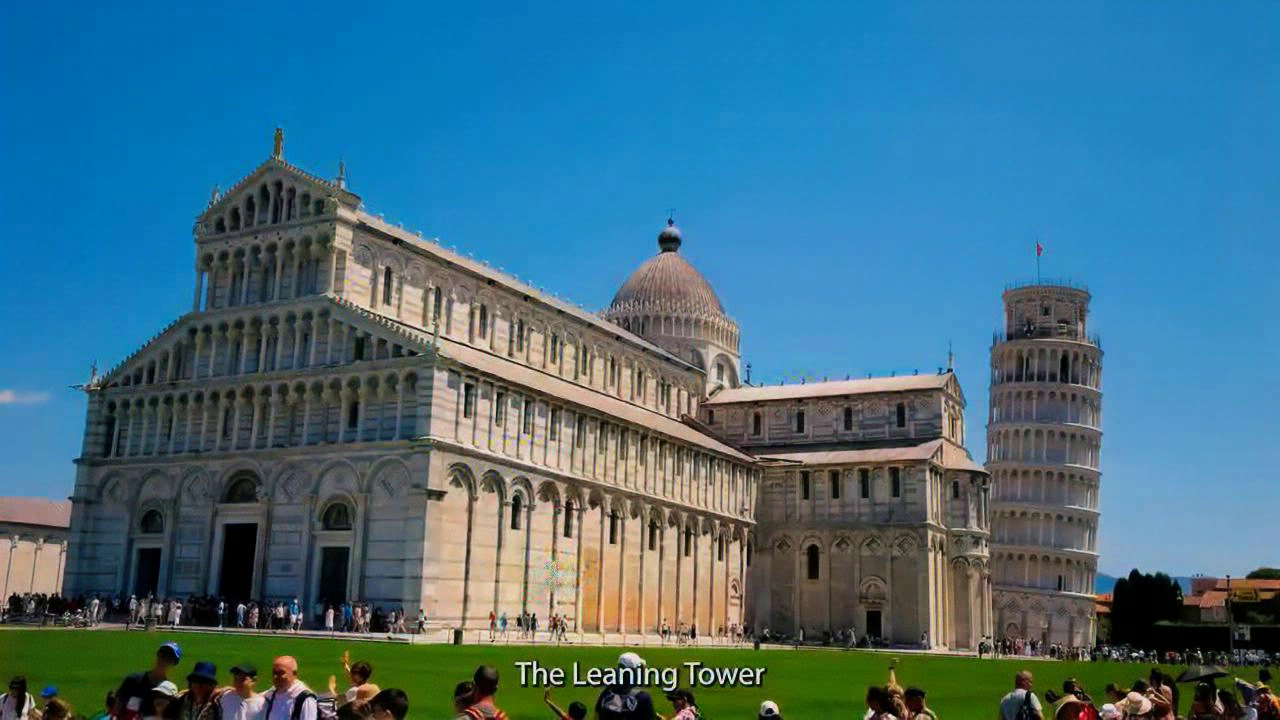}
        \caption*{Cap-Cut~\cite{algocapcut}}
        \end{minipage}
        }&{
        \begin{minipage}[b]{0.33\columnwidth}
        \centering
        \caption*{(\uppercase\expandafter{\romannumeral3})}
		\includegraphics[width=1\linewidth]{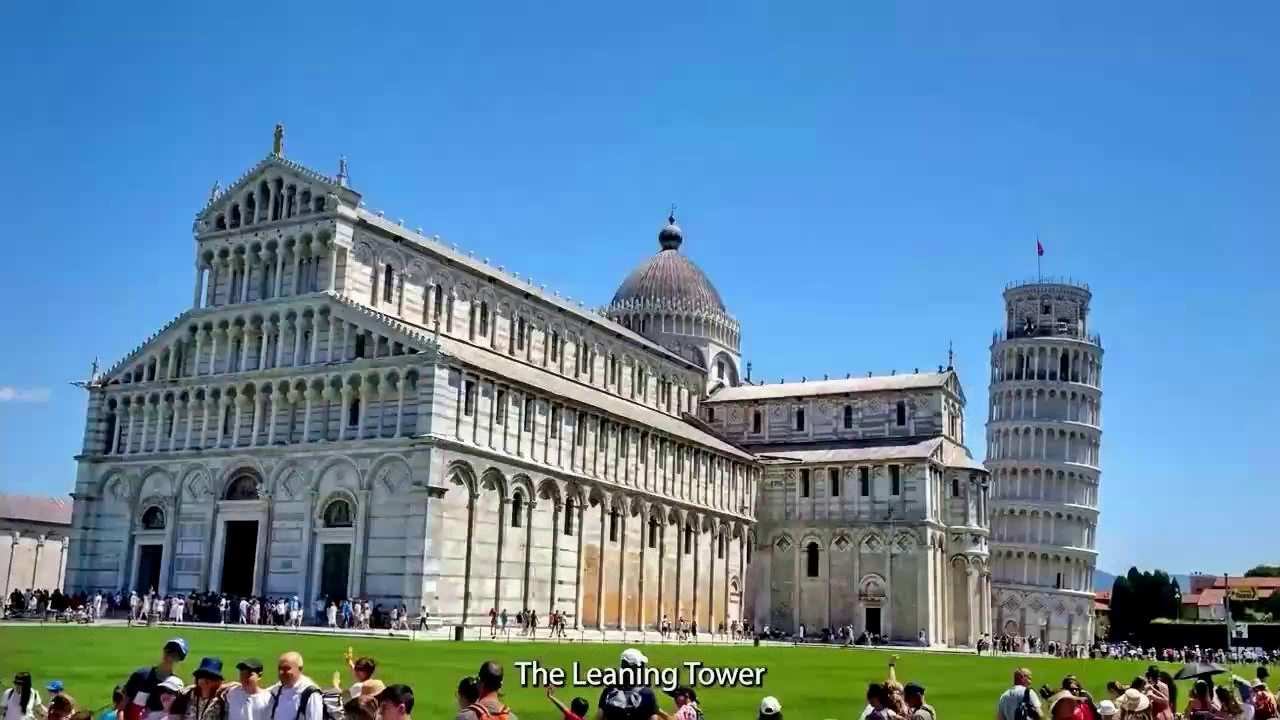}
        \caption*{PSE-Net~\cite{algopsenet}}
        \end{minipage}
        }
    \end{tabular}
}& \multicolumn{8}{c}{
    \begin{tabular}{ccc}
        {
        \begin{minipage}[b]{0.33\columnwidth}
        \centering
        \caption*{(\uppercase\expandafter{\romannumeral1})}
		\includegraphics[width=1\linewidth]{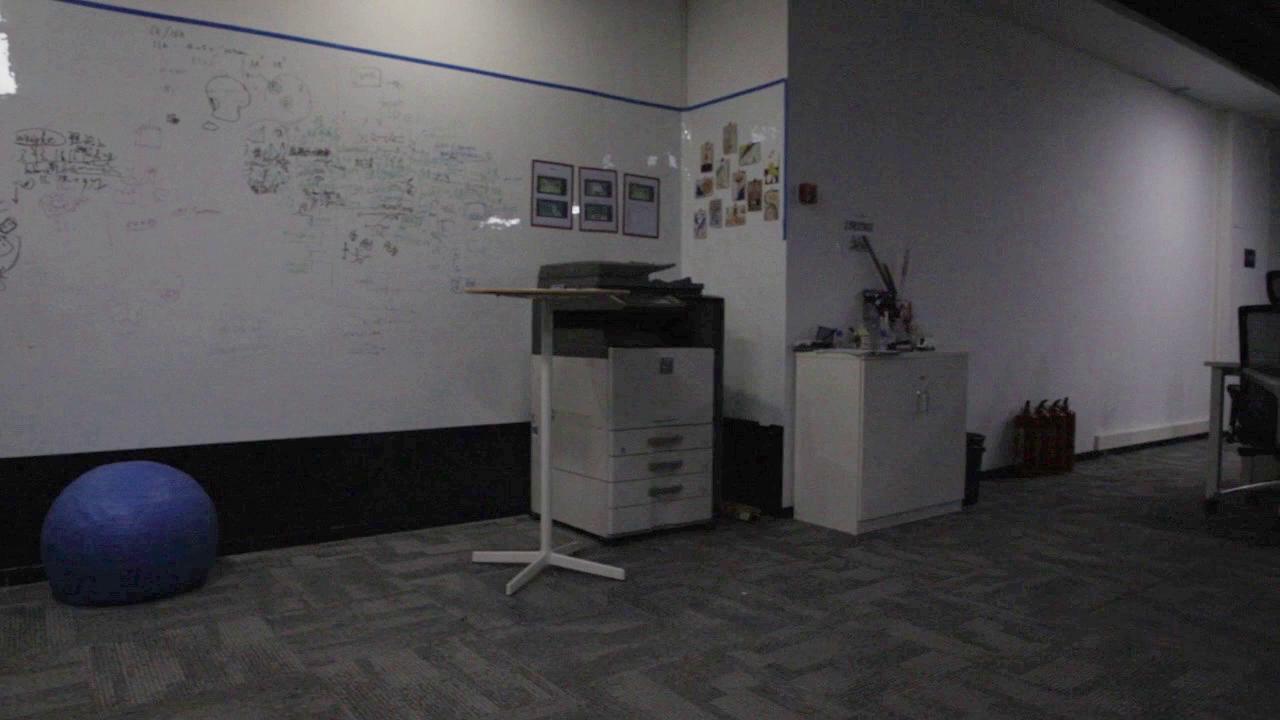}
        \caption*{GHE~\cite{algoGHE}}
        \end{minipage}
        }
        &
        {
        \begin{minipage}[b]{0.33\columnwidth}
        \centering
        \caption*{(\uppercase\expandafter{\romannumeral2})}
		\includegraphics[width=1\linewidth]{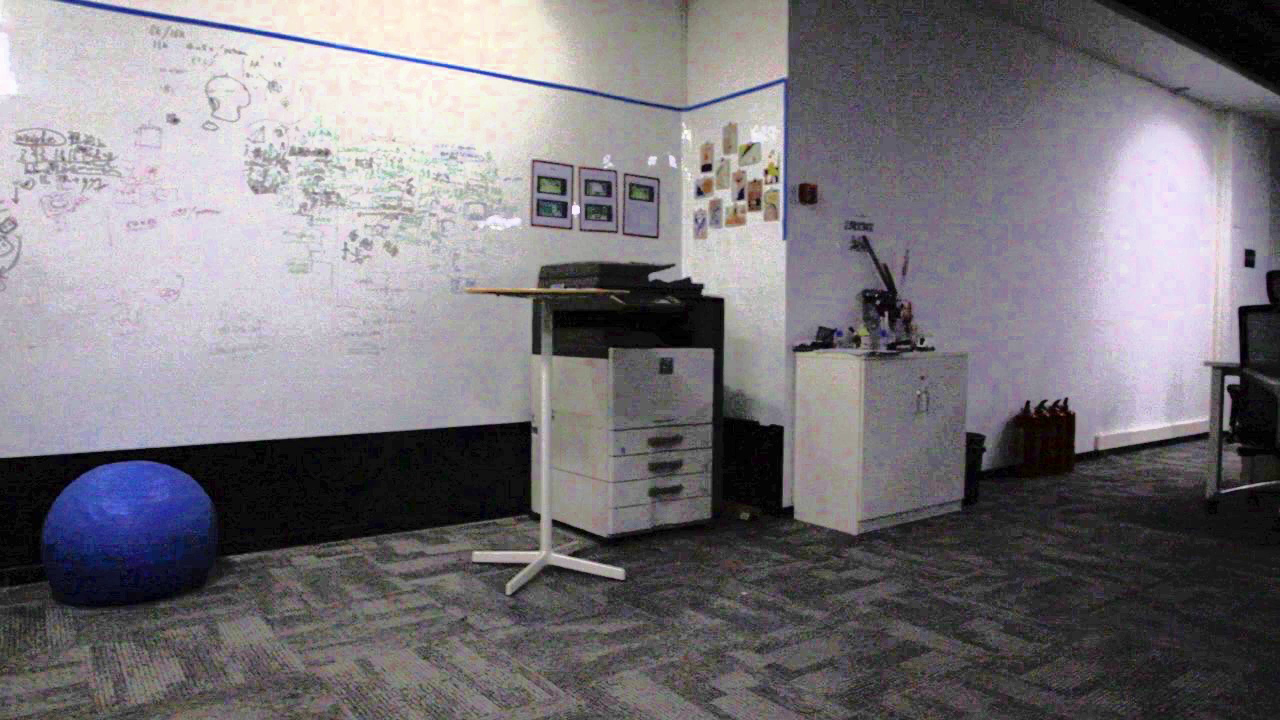}
        \caption*{SGZSL~\cite{algoSGZSL}}
        \end{minipage}
        }
        &
        {
        \begin{minipage}[b]{0.33\columnwidth}
        \centering
        \caption*{(\uppercase\expandafter{\romannumeral3})}
		\includegraphics[width=1\linewidth]{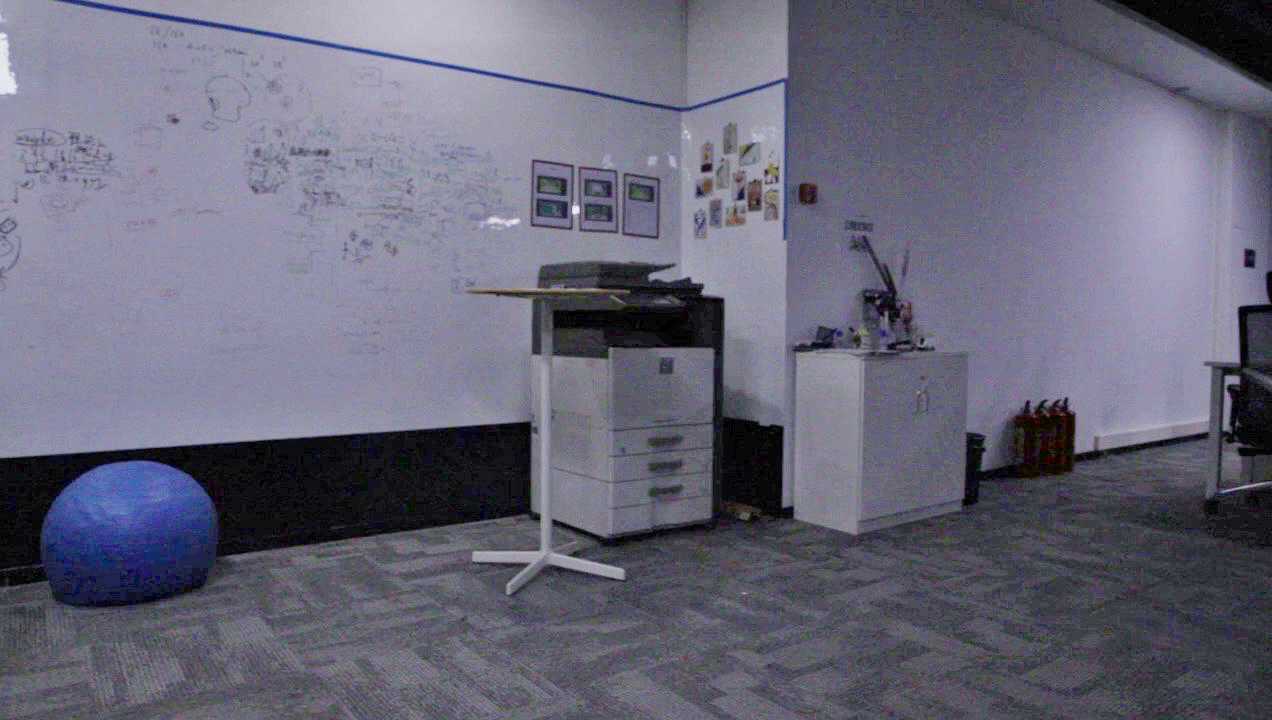}
        \caption*{StableLLVE~\cite{algoStableLLVE}}
        \end{minipage}
        }
    \end{tabular}
} \\ 
\hline
 Ground-truth &\textcolor{mygreen}{\uppercase\expandafter{\romannumeral1}} &{\uppercase\expandafter{\romannumeral2}} & {\uppercase\expandafter{\romannumeral3}} &
 Ground-truth &\textcolor{mygreen} {\uppercase\expandafter{\romannumeral1}} &{\uppercase\expandafter{\romannumeral2}} &{\uppercase\expandafter{\romannumeral3}} &
 Ground-truth &{\uppercase\expandafter{\romannumeral1}} &{\uppercase\expandafter{\romannumeral2}} & \textcolor{mygreen}{\uppercase\expandafter{\romannumeral3}} &
 Ground-truth &{\uppercase\expandafter{\romannumeral1}} &{\uppercase\expandafter{\romannumeral2}} &\textcolor{mygreen}{\uppercase\expandafter{\romannumeral3}}\\
\hdashline
Simple-VQA~\cite{vqasimple}&&{}\textcolor{mygreen}{\ding{52}}&{}&
Q-Align~\cite{vqaqalign}&{}&\textcolor{mygreen}{\ding{52}}&{}&
Simple-VQA~\cite{vqasimple}&\textcolor{mygreen}{\ding{52}}&{}&{}&
Q-Align~\cite{vqaqalign}&{}&\textcolor{mygreen}{\ding{52}}&{}\\

 Fast-VQA~\cite{vqafastvqa}&&{}&\textcolor{mygreen}{\ding{52}}&
 Light-VQA\cite{baselightvqa}&{}&{}&\textcolor{mygreen}{\ding{52}}&
 Fast-VQA~\cite{vqafastvqa}&{}&{\textcolor{mygreen}{\ding{52}}}&&
 Light-VQA\cite{baselightvqa}&{}&&{\textcolor{mygreen}{\ding{52}}}\\
 
 Max-VQA~\cite{vqamaxvqa}&{}&&{\textcolor{mygreen}{\ding{52}}}&
 \textbf{Light-VQA+}&\textcolor{mygreen}{\ding{52}}&{}&{}&
 Max-VQA~\cite{vqamaxvqa}&{}&{\textcolor{mygreen}{\ding{52}}}&&
\textbf{ Light-VQA+}&{}&&{\textcolor{mygreen}{\ding{52}}}\\
\bottomrule[1.5pt]
\end{tabular}
}
\end{table*}

Compared to the widely spreading texts and images, videos are generally more entertaining and informative. However, due to the influence of photographic devices and skills, the quality of UGC videos often varies greatly. It is frustrating that the precious and memorable moment is degraded by photographic limitations (e.g. over- and under- exposure, low frame-rate, low resolution, etc.).

To address the problems mentioned above, specific video enhancement algorithms have been proposed. There are algorithms designed for super resolution videos~\cite{shi2021learning, liu2021exploit, liu2020end, liu2018robust, yin2023online}, for frame interpolation~\cite{shi2022video, shi2021video}, for optical flow~\cite{wu2023accflow}, and for exposure~\cite{li2023fastllve}. In this paper, we focus on the quality assessment of expsure-correction, which are mainly composed of low-light videos and over-exposed videos, along with the corrected versions. Low-light videos are often captured in the low- or back-lighting environments, while over-exposed videos are often captured in direct sunlight environments. These degraded videos usually suffer from significant degradations such as low visibility and noises, thus will challenge many computer vision downstream tasks such as object detection~\cite{zou2023object}, semantic segmentation~\cite{long2015fully}, etc., which are usually resorted to the videos with good quality.

Therefore, many VEC algorithms have been developed to improve the visual quality of such damaged videos. One straightforward way is to split the video into frames, and then apply the Low-Light Image Enhancement (LLIE) and Over-Exposed Image Recovery (OEIR) algorithms to process each frame of this video. Representative traditional algorithms as such include ACE~\cite{algoACE}, AGCCPF~\cite{algoAGCCPF}, GHE~\cite{algoGHE}, IAGC~\cite{algoIAGC}, and BPHEME~\cite{algoBPHEME}. There are also some deep-learning-based algorithms that focus on images. For LLIE, there are MBLLEN~\cite{algoMBLLEN}, SGZSL~\cite{algoSGZSL}, and DCC-Net~\cite{algoDCCNET}. As for OEIR, there are DIEREC~\cite{algoDIEREC}, LMSPEC~\cite{algoLMSPEC}, LECVCM~\cite{algoLECVCM}, LCDP-Net~\cite{algolcdpnet}, ECMEIQ~\cite{algoECMEIQ}, as well as PSE-Net~\cite{algopsenet}. {While some of these algorithms are capable of correcting videos well, due to the lack of consideration in the consistency of neighboring frames in a video, some of LLIE algorithms will lead to temporal instability. To address this issue, some LLVE algorithms that take temporal consistency into account are proposed, such as MBLLVEN~\cite{algoMBLLEN}, SDSD~\cite{algoSDSD}, SMID~\cite{algoSMID}, and StableLLVE~\cite{algoStableLLVE}. In contrast, there are no existing OEVR algorithms available in the literature. One possible reason is that the temporal inconsistency when processing the videos with OEIR is not as severe as that of LLIE. Therefore, the OEVR task is still commonly addressed by OEIR algorithms. To reduce the impact of lack of OEVR algorithms when collecting our VEC-QA dataset, a commercial software named CapCut~\cite{algocapcut} is also used to recover over-exposed videos.}

All algorithms need a metric for improvement, where the quality assessment methods~\cite{kou2024subjective, wu2024towards,  zhang2023perceptual, zhang2023advancing, li2023agiqa, li2024qrefine, zhang2024reduced, algokou2023stablevqa, algozhang2023qboost} play an important role. For VQA, there are mainly two branches: subjective VQA and objective ones. Subjective VQA is to assess the video by human, which is naturally more expensive and time-consuming. Objective VQA can be divided into Full-Reference (FR) VQA~\cite{FRVQA}, Reduced-Reference (RR) VQA~\cite{RRVQA} and No-Reference (NR) VQA~\cite{NRVQA} contingent on the amount of required pristine video information. Due to the difficulty in obtaining reference videos, NR-VQA enjoys the most attention from researchers. In the early development stages of NR-VQA, researchers often evaluate video quality based on handcrafted features, such as structure, texture, and statistical features. Recently, owing to the potential in practical applications, deep-learning-based NR-VQA models with multimodal large language models have progressively dominated the VQA field. However, most existing VQA models are designed for general purposes instead of exposure correction.  Few models specifically evaluate the quality of videos correcteds by VEC algorithms, which is possibly due to the lack of corresponding datasets.

As a result, Light-VQA~\cite{baselightvqa}, specialized in LLVE, is proposed. It combines deep-learning-based features and handcrafted features to improve the accuracy of the assessment for low-light videos. To the best of our knowledge, Light-VQA is the best VQA model for assessing VEC algorithms.
One downside of Light-VQA is the usage of several traditional feature extractors such as standard deviation pooling that has the limited representative ability. Besides, it also fails to take the over-exposed videos into consideration, which is also of vital importance.

Therefore, in this paper, we elaborately build the Video Exposure Correction Quality Assessment (\textbf{VEC-QA}) dataset to facilitate the work on evaluating the performance of VEC algorithms by expanding the LLVE-QA dataset~\cite{baselightvqa}. Different from general datasets which commonly consist of original UGC videos with various degradations, VEC-QA dataset contains 254 original low-light videos and 1,806 enhanced videos from representative enhancement algorithms, and 205 original over-exposed videos and 2,253 recovered videos from representative recovery algorithms, each with a corresponding Mean Opinion Score (MOS). 

Subsequently, we propose a quality assessment model for inproper-exposed video enhancement based on Light-VQA, named \textbf{Light-VQA+}, whose capability is demonstrated in Table \ref{intro}. Since brightness and noise have the most significant impact on VEC-VQA, we specifically extract features related to brightness, noise, and brightness consistency to enhance the model's capabilities. While Light-VQA utilizes the traditional handcrafted algorithms to extract such feartures, Light-VQA+ borrows the strength of Multimodal Large Language Models (MLLM)~\cite{clipmodel}, leading to a more accurate and efficient way for extracting such features. Besides, we still need the semantic features and motion features extracted from deep neural network when evaluating the quality of a video. The features mentioned above can be categorized into two aspects: spatial and temporal ones. To fuse these features, a cross-attention module~\cite{crossattention} is deployed to combine the information from different sources. Also, it is worth noting that when a human watches a video, he/she does not focus evenly on the entire video. Specific video clips would receive more attention than others, which follows the Human Visual System (HVS)~\cite{626903}. To better imitate the HVS, a trainable attention weights is then introduced when obtaining the final quality score of a video.
Extensive experiments validate the effectiveness of our network design.

The contributions of this paper can be summarized as follows:
\begin{enumerate}
    \item After applying advanced VEC algorithms on a collection of over-exposed videos that feature diverse content and varying brightness levels, we undertake a subjective experiment to construct the OEVR-QA dataset. By adding it to the LLVE-QA dataset, we obtain a dataset specialized in VEC, named \textbf{VEC-QA}.
    \item {Benefiting from the developed dataset, we propose a Light-VQA-based quality assessment model: \textbf{Light-VQA+}, which extracts brightness and noise features through CLIP with vision-language guidances provided by delicately designed prompts. Then it combines spatial-temporal information via cross-attention, followed with a quality regression to obtain the quality score for each video clip. Finally, a set of trainable weights are employed on all video clips to obtain the final score, making the assessment align with the HVS.}
    \item The proposed Light-VQA+ surpasses Light-VQA on VEC-QA dataset as well as other public datasets. We envision that Light-VQA+ holds significant promise as a pivotal metric for the assessment as well as the development of VEC algorithms.
\end{enumerate}

\begin{table*}[ht]
\renewcommand{\arraystretch}{1.5}
    \centering
    \caption{Comprison of Existing VQA Dataset}
    \label{vqadatasets}
    \resizebox{\linewidth}{!}{
    \begin{tabular}{c c c c c c}
    \toprule[1.5pt]
    {Dataset}&{Source}&{Number}&{Length}&{Resulution}&{Specialized in VEC}\\
    \hline
    Live-Qualcomm~\cite{dataLiveQualcomm}&{mobile devices}&{208}&{15s}&{1080p}&{\ding{56}}\\
    Live-VQC~\cite{dataLiveVQC}&{43 device models}&{585}&{10s}&{Varies from 240p to 1080p}&{\ding{56}}\\
    UGC-VIDEO~\cite{dataugcvideo}&{TikTok}&{50}&{10s}&{720p}&{\ding{56}}\\
    LIVE-WC~\cite{dataLiveWC}&{Live-VQC}&{3,740}&{10s}&{Varies from 360p to 1080p}&{\ding{56}}\\
    KoNVID-1K~\cite{datakonvid}&{YFCC100m~\cite{dataYFCC100M}}&{1,200}&{8s}&{540p}&{\ding{56}}\\
    Youtube-UGC~\cite{datayoutubeugc}&{\url{https://youtube.com}}&{1,500}&{20s}&{Varies from 240p to 4k}&{\ding{56}}\\
    VDPVE~\cite{datavdpve}&{Other Datasets}&{1,211}&{8-10s}&{720p, 1080p}&{\ding{56}}\\
    \textbf{VEC-QA}&{Internet \& Other Datasets}&{4,518}&{8-10s}&{720p}&{\ding{52}}\\
    \bottomrule[1.5pt]
    \end{tabular}}
\end{table*}

\section{Related Work}

\subsection{Video Exposure Correction}

To correct the exposure of the improper exposed videos, one straightforward way is to split the video into frames, so as to take advantage of existing algorithms capable of correcting the exposure of images. ACE~\cite{algoACE} utilizes local adaptive filtering to achieve image brightness, color, and contrast adjustments with local and nonlinear features while satisfies both the gray world theory and the white patch hypothesis. AGCCPF~\cite{algoAGCCPF} enhances the brightness and contrast of images using the gamma correction and weighted probability distribution of pixels. GHE~\cite{algoGHE} applies a transformation on image histogram to redistribute the pixel intensity, resulting in a more favorable visual result. BPHEME~\cite{algoBPHEME} corrects the improper-exposed video by balancing the brightness preserving histogram with maximum entropy. IAGC~\cite{algoIAGC} employs advanced adaptive gamma correction for contrast enhancement in brightness-compromised images. 

In addition to traditional methods, deep-learning-based exposure-correction algorithms are developing rapidly. DIEREC~\cite{algoDIEREC} introduces an automatic method capable of enhancing images under varied exposure conditions with notable quality. LMSPEC~\cite{algoLMSPEC} advances a coarse-to-fine DNN approach for correcting exposure inaccuracies. LECVCM~\cite{algoLECVCM} employs a deep feature matching loss within its model, facilitating exposure-invariant feature learning for consistent image exposure. LCDP-Net~\cite{algolcdpnet} features a dual-illumination learning strategy to address exposure disparities. ECMEIQ~\cite{algoECMEIQ} presents an end-to-end model designed to correct both under- and over-exposure through a structure that comprises an image encoder, consecutive residual blocks, and an image decoder. PSE-Net~\cite{algopsenet} introduces an unsupervised enhancement framework, effective across different lighting scenarios without necessitating the well-exposed images for ground-truth comparison. Zhang et al.~\cite{algoDCCNET} propose a consistent network to improve illumination and preserve color consistency of low-light images. 

However, applying image exposure correction algorithms directly to videos sometimes lead to temporal consistency problems such as motion artifacts and brightness consistency, which will ultimately reduce the quality of videos. Therefore, in order to maintain the temporal consistency of videos, specific LLVE algorithms are proposed. MBLLVEN~\cite{algoMBLLEN} processes low-light videos via 3D convolution to extract temporal information and preserve temporal consistency. Wang et al.~\cite{algoSDSD} collect a new dataset that contains high-quality spatially-aligned video pairs in both low-light and normal-light conditions, and further design a self-supervised network to reduce noises and enhance the illumination based on the Retinex theory. Chen et al.~\cite{algoSMID} propose a siamese network and introduce a self-consistency loss to preserve color while suppressing spatial and temporal artifacts efficiently. StableLLVE~\cite{algoStableLLVE} maintains the temporal consistency after enhancement by learning and inferring motion field (\textit{i.e.}, optical flow) from the synthesized short-range video sequences.  In order to build our VEC-QA dataset, both image- and video-based exposure-correction algorithms are leveraged to increase the performance diversity of intra-frame and inter-frame exposure recovery.

\subsection{VQA Datasets}

With the purpose of facilitating the development of VQA algorithms, many VQA datasets have been proposed. Videos in LIVE-Qualcomm~\cite{dataLiveQualcomm} contain the following 6 distortion types: color, exposure, focus, artifacts, sharpness, and stabilization. LIVE-VQC~\cite{dataLiveVQC} contains 585 videos, which are captured by various cameras with different resolutions. In addition to the common distortions, the visual quality of UGC videos is influenced by compression generated when uploading to and downloading from the Internet. UGC-VIDEO~\cite{dataugcvideo} and LIVE-WC~\cite{dataLiveWC} simulate the specific distortion by utilizing several video compression algorithms. KoNViD-1k~\cite{datakonvid}, YouTube-UGC~\cite{datayoutubeugc}, and LSVQ~\cite{dataLSVQ} extensively collect in-the-wild UGC videos from the Internet, greatly expanding the scale of VQA datasets. Besides, VDPVE~\cite{datavdpve} is constructed to fill in the gaps of VQA datasets specially for video enhancement, which can further promote the refined development of VQA models. However, most of existing datasets only contain unprocessed UGC videos with various distortions. While VDPVE takes enhanced videos into account, it is still general and not targeted. LLVE-QA~\cite{baselightvqa} is a VQA dataset specialized in low-light video enhancement. However, LLVE-QA fails to take the over-exposed video recovery into consideration. Our VEC-QA dataset focus on videos with original poor exposure and their corrected versions, establishing a strong foundation for developing the specialized VQA models for exposure correction. The details of these datasets can be found in Tab. \ref{vqadatasets}.

\begin{figure*}[t]
\setlength{\abovecaptionskip}{0cm} 
\setlength{\belowcaptionskip}{0.1cm}
\centering
\begin{subfigure}[t]{0.15\textwidth}
    \includegraphics[width=1\textwidth]{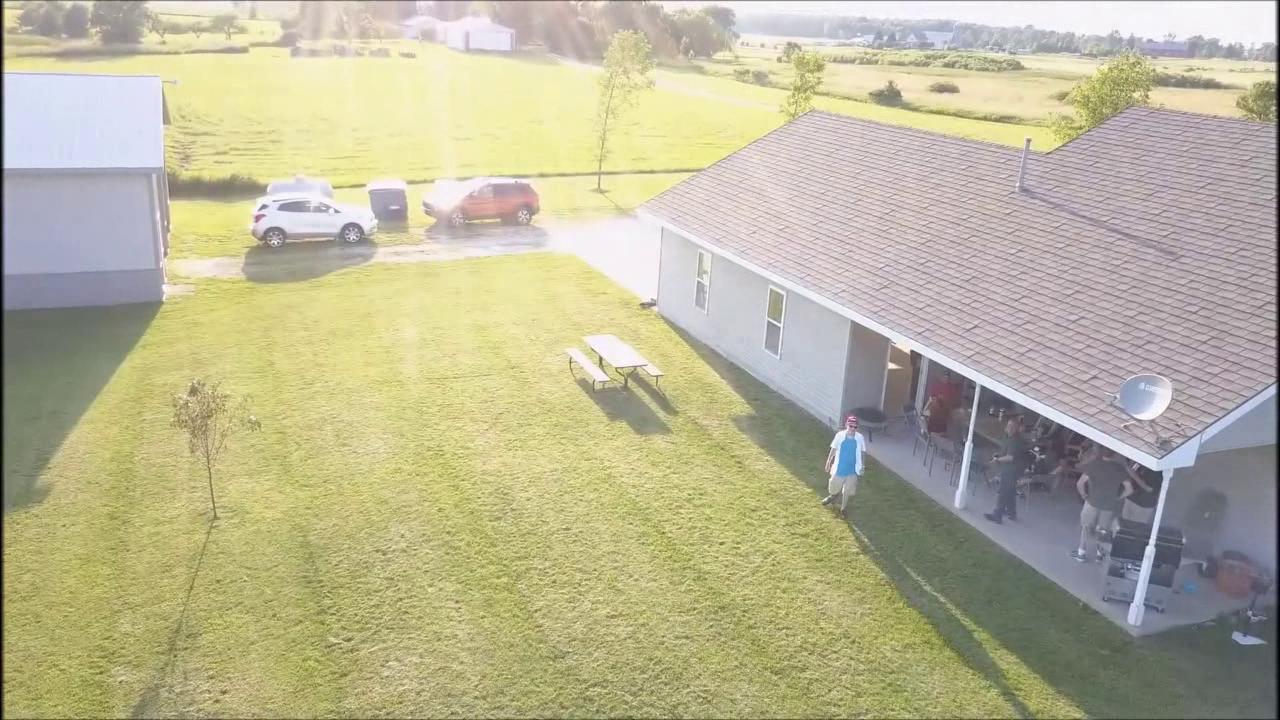}
    \subcaption*{Original}
\end{subfigure}
\begin{subfigure}[t]{0.15\textwidth}
    \includegraphics[width=1\textwidth]{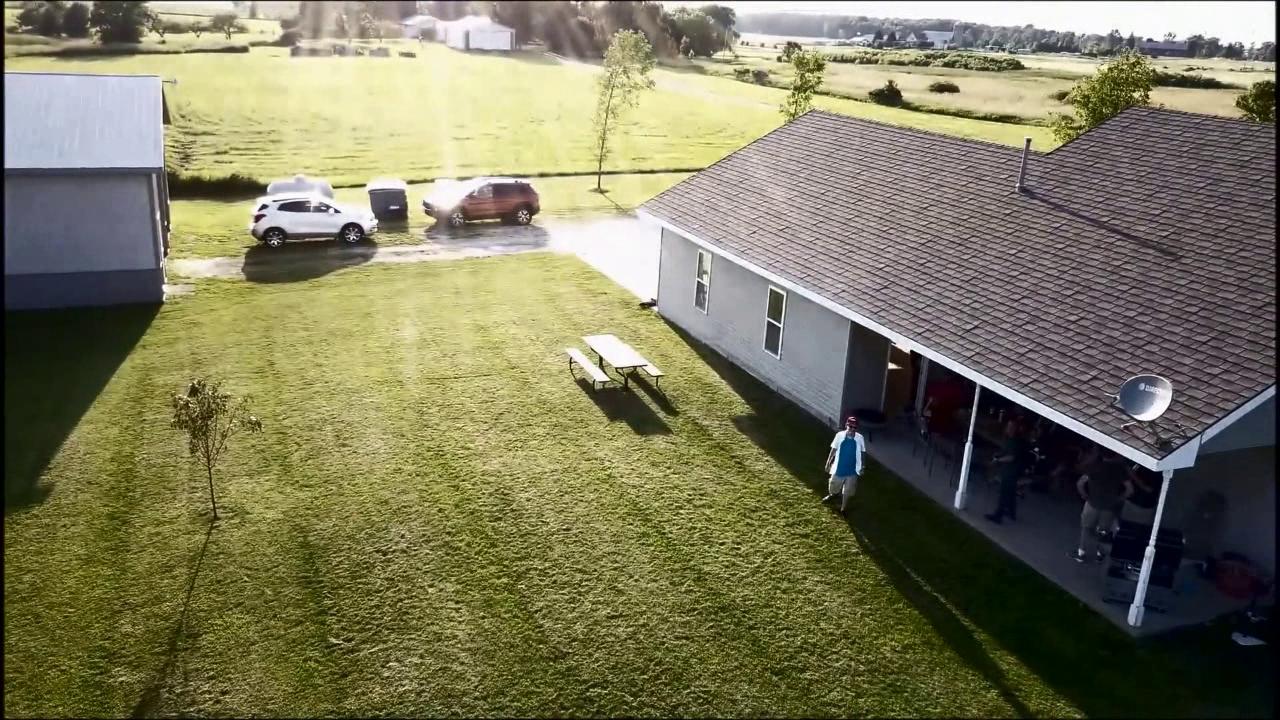}
    \subcaption*{ACE~\cite{algoACE}}
\end{subfigure}
\begin{subfigure}[t]{0.15\textwidth}
    \includegraphics[width=1\textwidth]{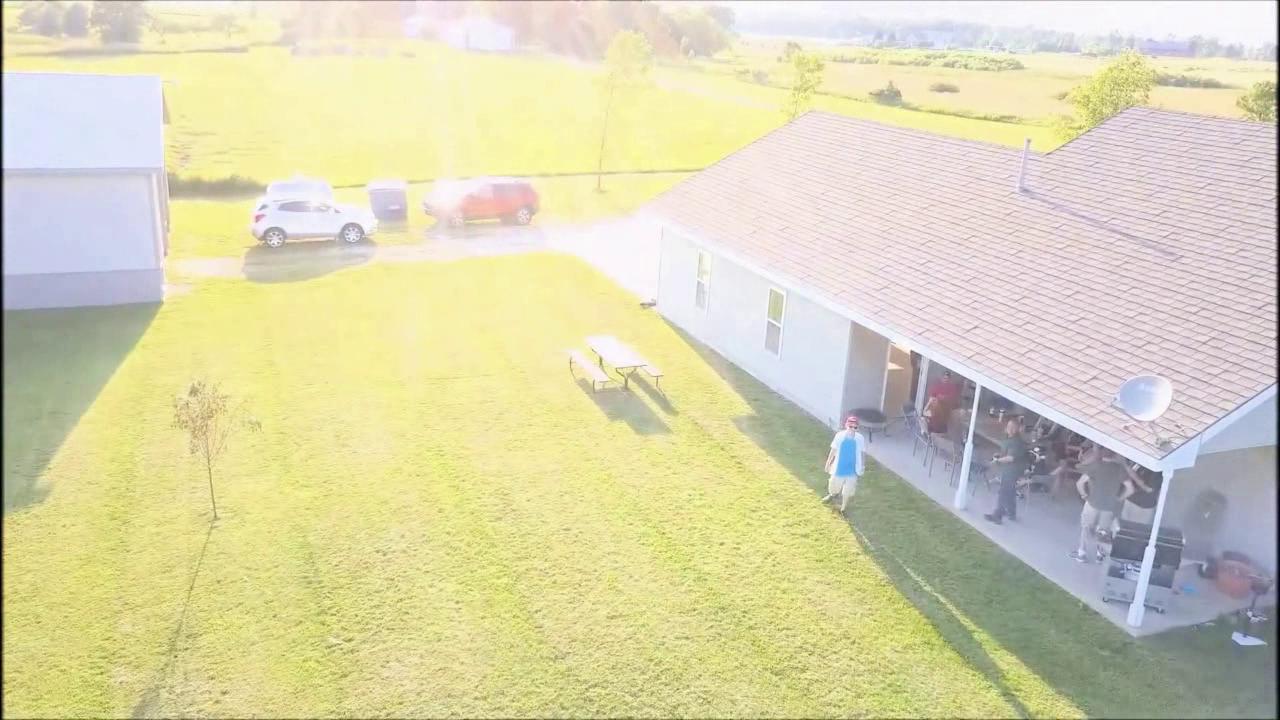}
    \subcaption*{AGCCPF~\cite{algoAGCCPF}}
\end{subfigure}
\quad
\begin{subfigure}[t]{0.15\textwidth}
    \includegraphics[width=1\textwidth]{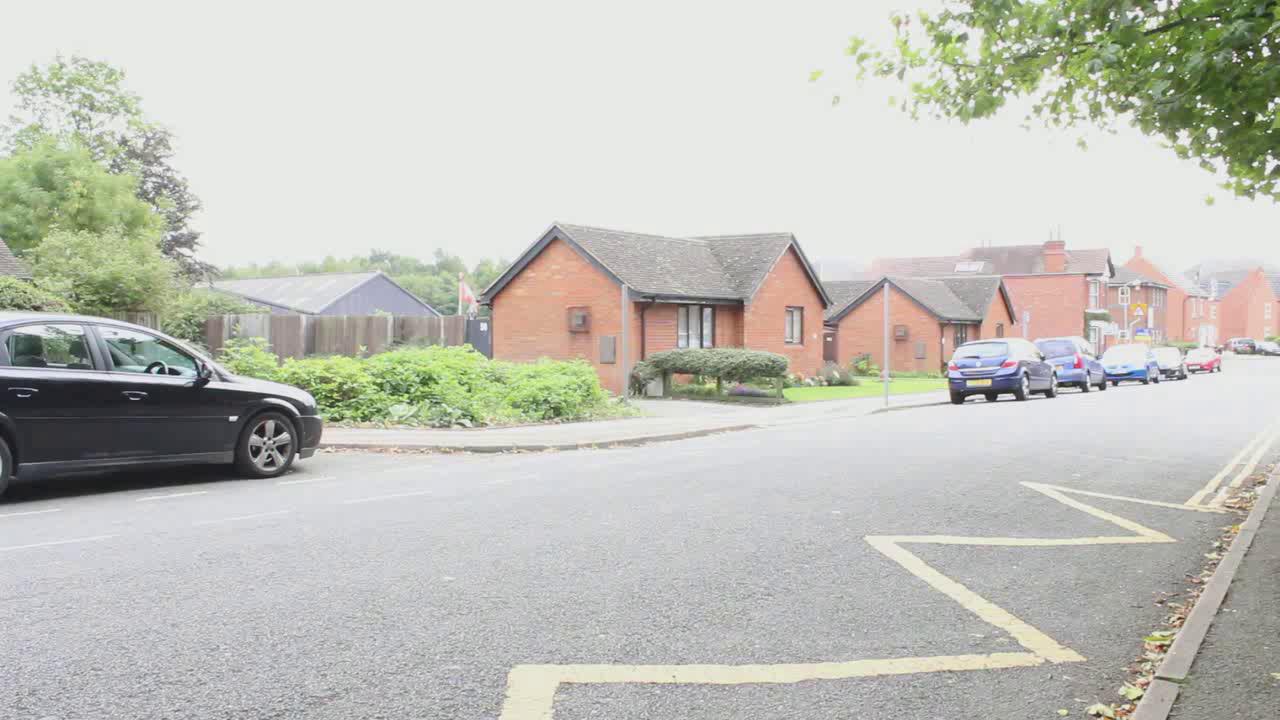}
    \subcaption*{Original}
\end{subfigure}
\begin{subfigure}[t]{0.15\textwidth}
    \includegraphics[width=1\textwidth]{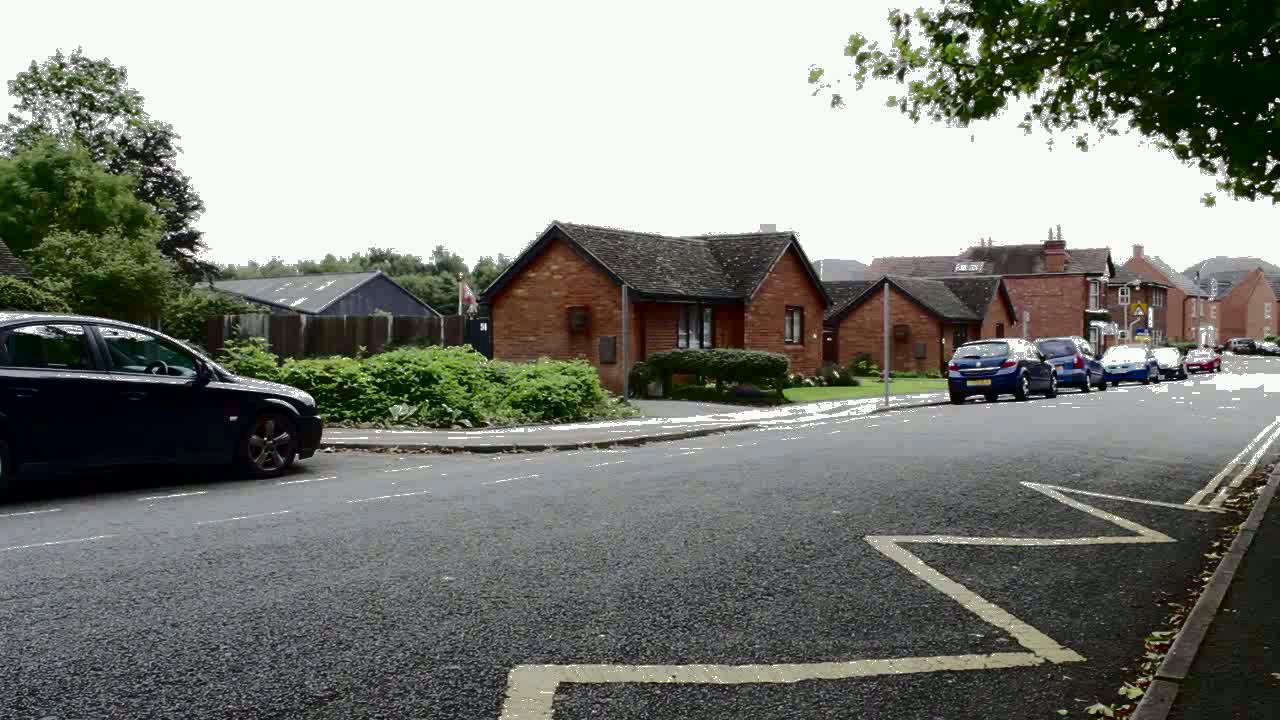}
    \subcaption*{ACE~\cite{algoACE}}
\end{subfigure}
\begin{subfigure}[t]{0.15\textwidth}
    \includegraphics[width=1\textwidth]{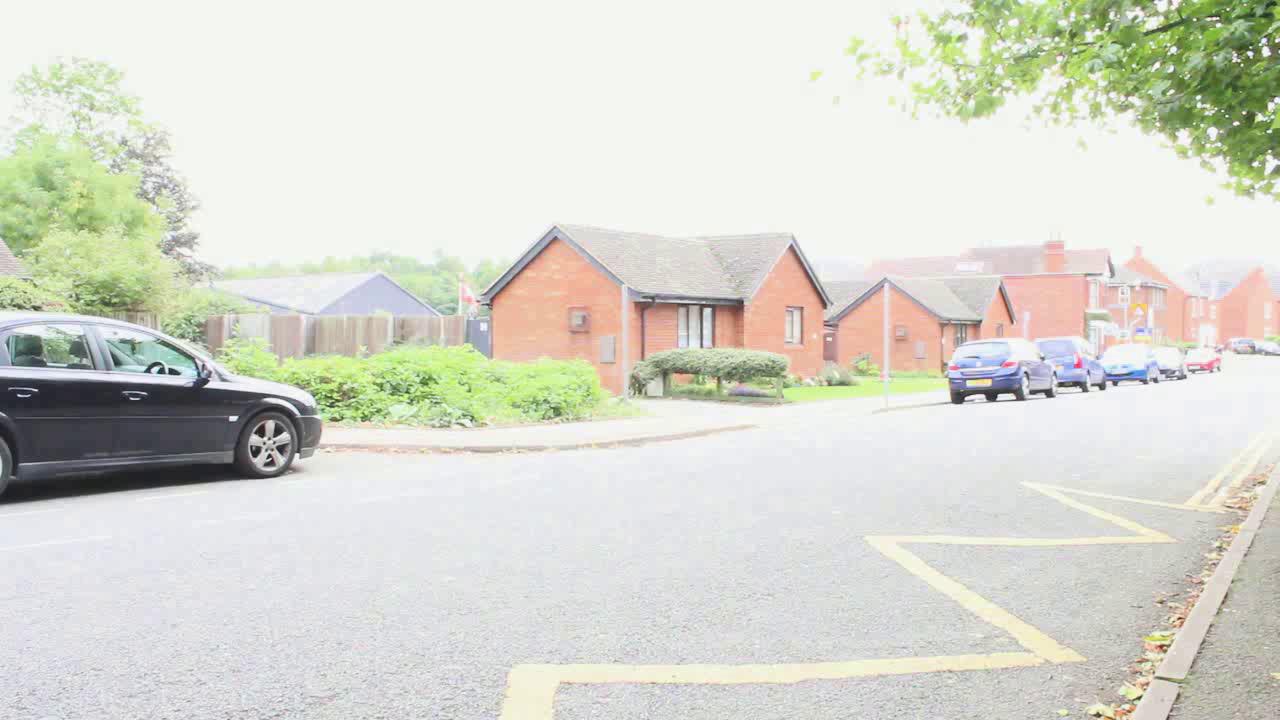}
    \subcaption*{AGCCPF~\cite{algoAGCCPF}}
\end{subfigure}
\begin{subfigure}[t]{0.15\textwidth}
    \includegraphics[width=1\textwidth]{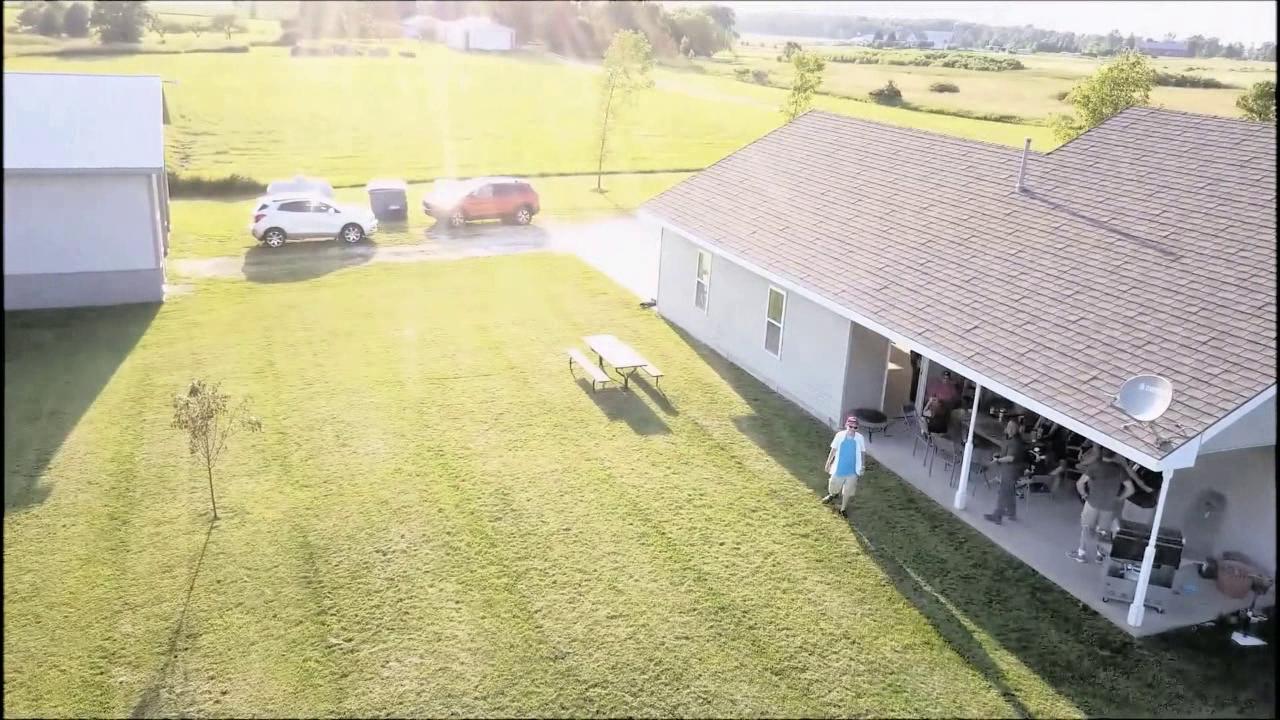}
    \subcaption*{BPHEME~\cite{algoBPHEME}}
\end{subfigure}
\begin{subfigure}[t]{0.15\textwidth}
    \includegraphics[width=1\textwidth]{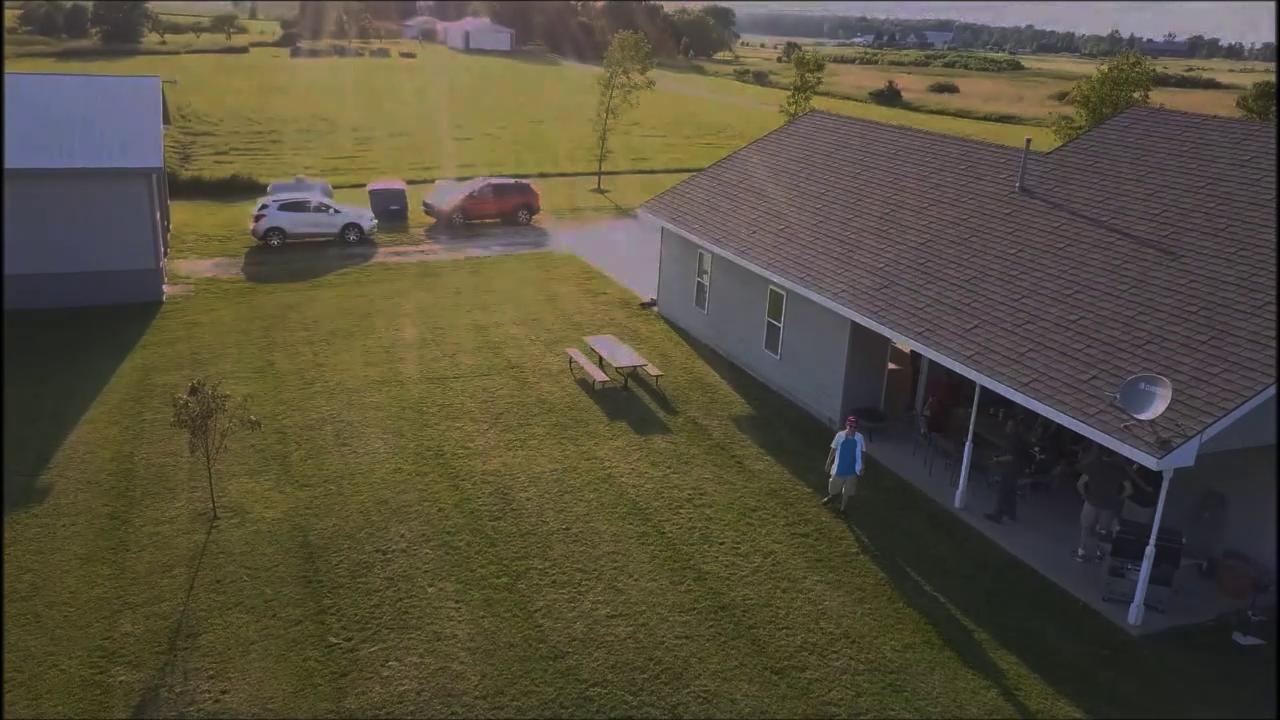}
    \subcaption*{Cap-Cut~\cite{algocapcut}}
\end{subfigure}
\begin{subfigure}[t]{0.15\textwidth}
    \includegraphics[width=1\textwidth]{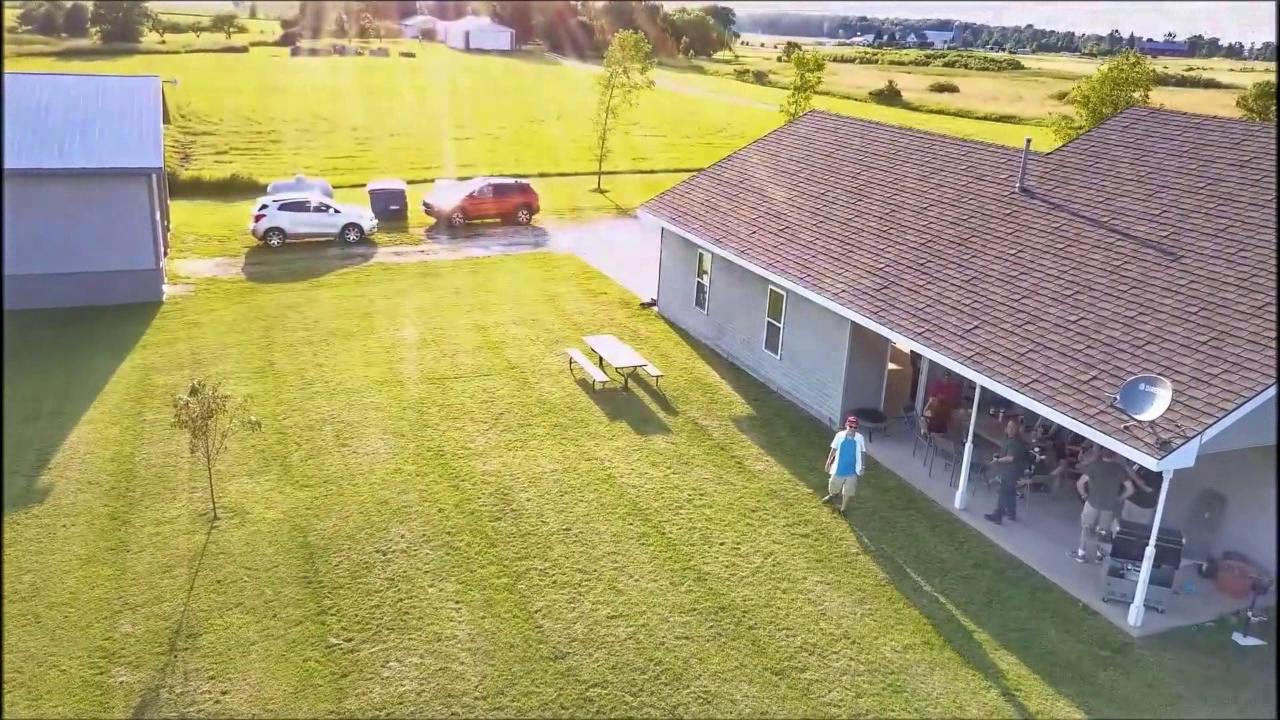}
    \subcaption*{DIEREC~\cite{algoDIEREC}}
\end{subfigure}
\quad
\begin{subfigure}[t]{0.15\textwidth}
    \includegraphics[width=1\textwidth]{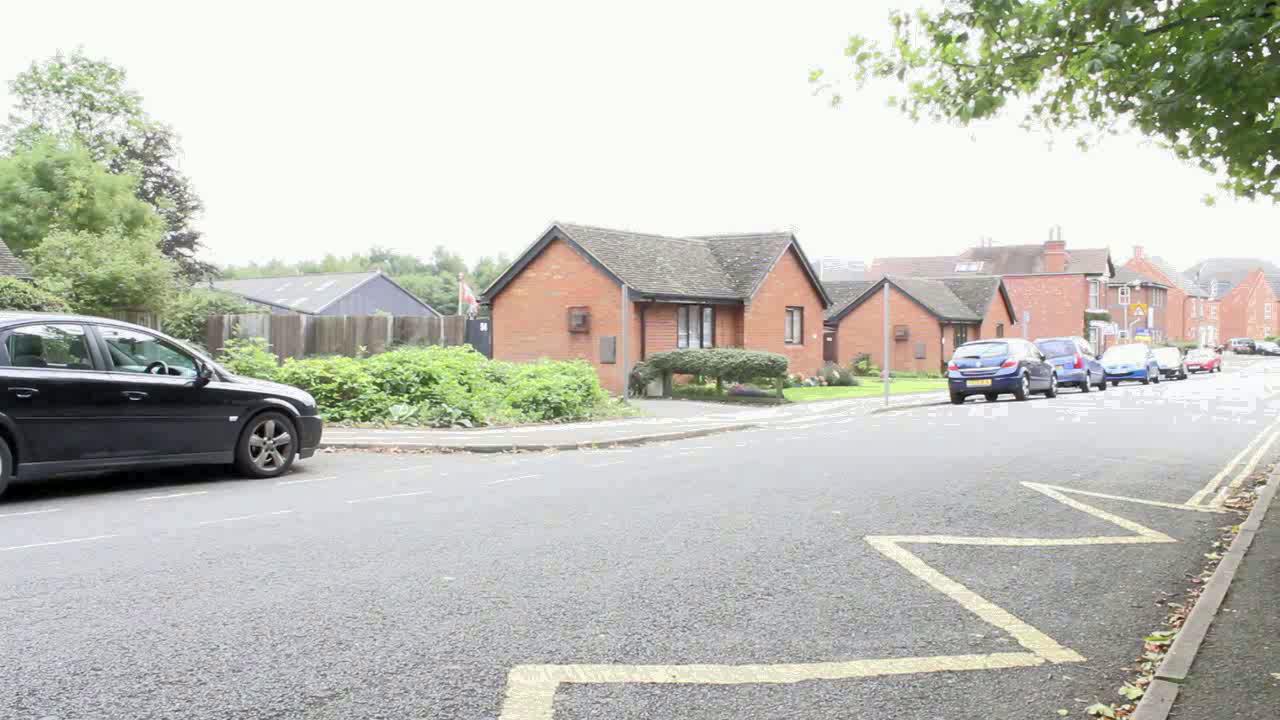}
    \subcaption*{BPHEME~\cite{algoBPHEME}}
\end{subfigure}
\begin{subfigure}[t]{0.15\textwidth}
    \includegraphics[width=1\textwidth]{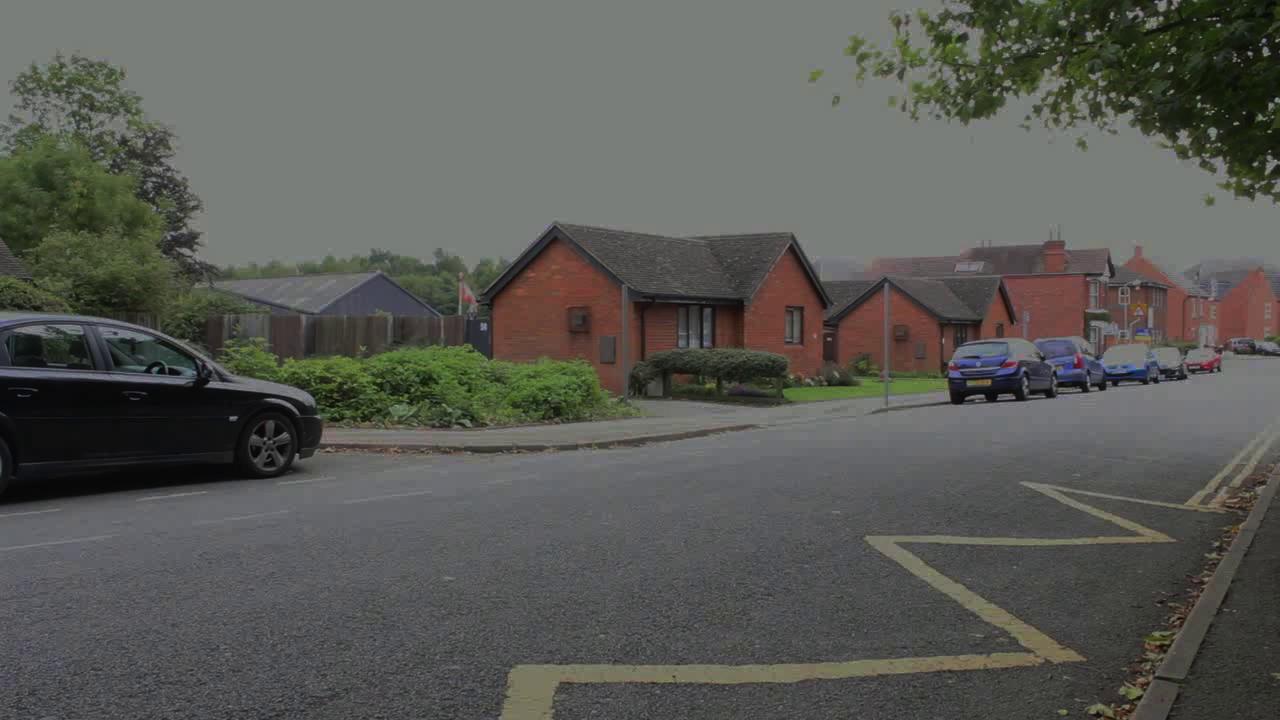}
    \subcaption*{Cap-Cut~\cite{algocapcut}}
\end{subfigure}
\begin{subfigure}[t]{0.15\textwidth}
    \includegraphics[width=1\textwidth]{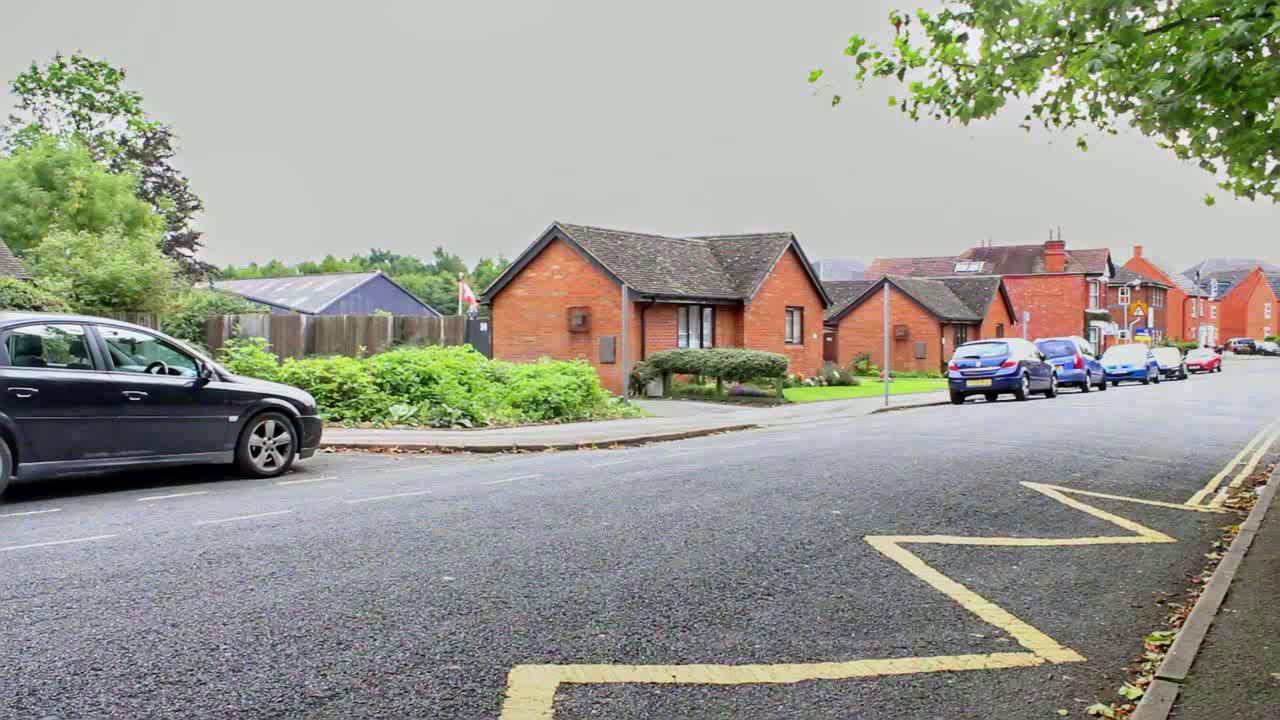}
    \subcaption*{DIEREC~\cite{algoDIEREC}}
\end{subfigure}
\quad
\begin{subfigure}[t]{0.15\textwidth}
    \includegraphics[width=1\textwidth]{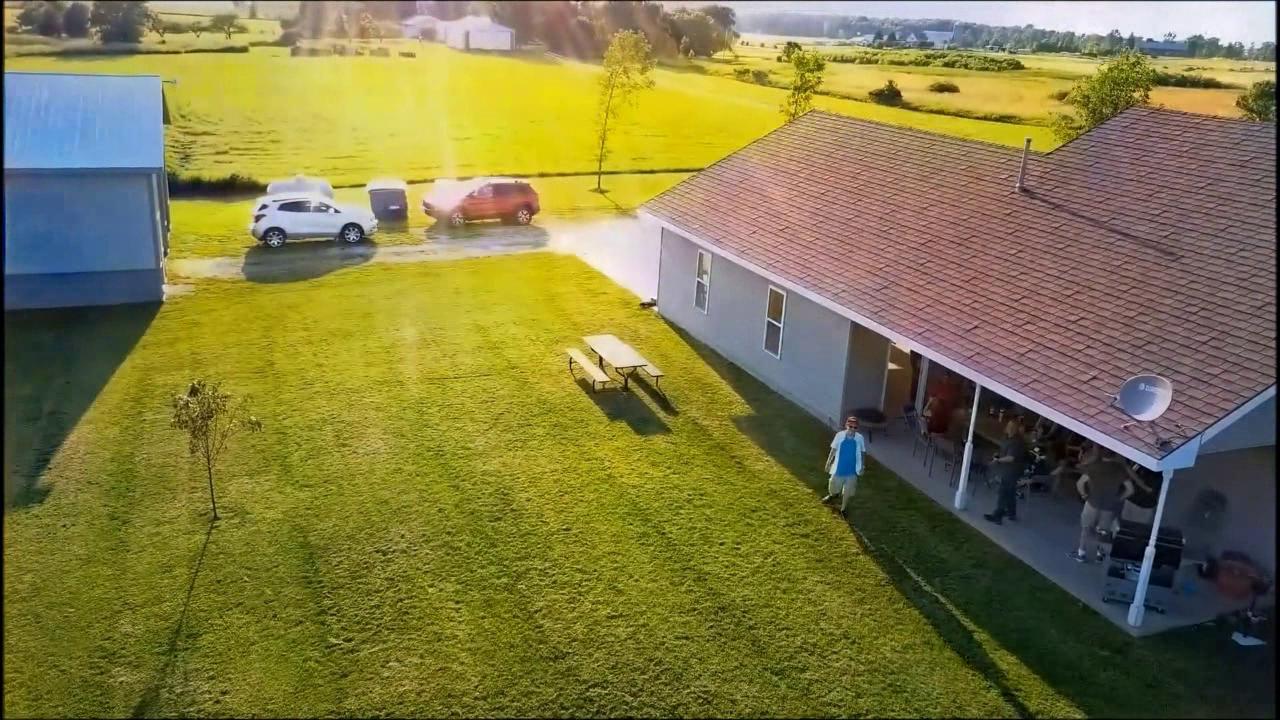}
    \subcaption*{LMSPEC~\cite{algoLMSPEC}}
\end{subfigure}
\begin{subfigure}[t]{0.15\textwidth}
    \includegraphics[width=1\textwidth]{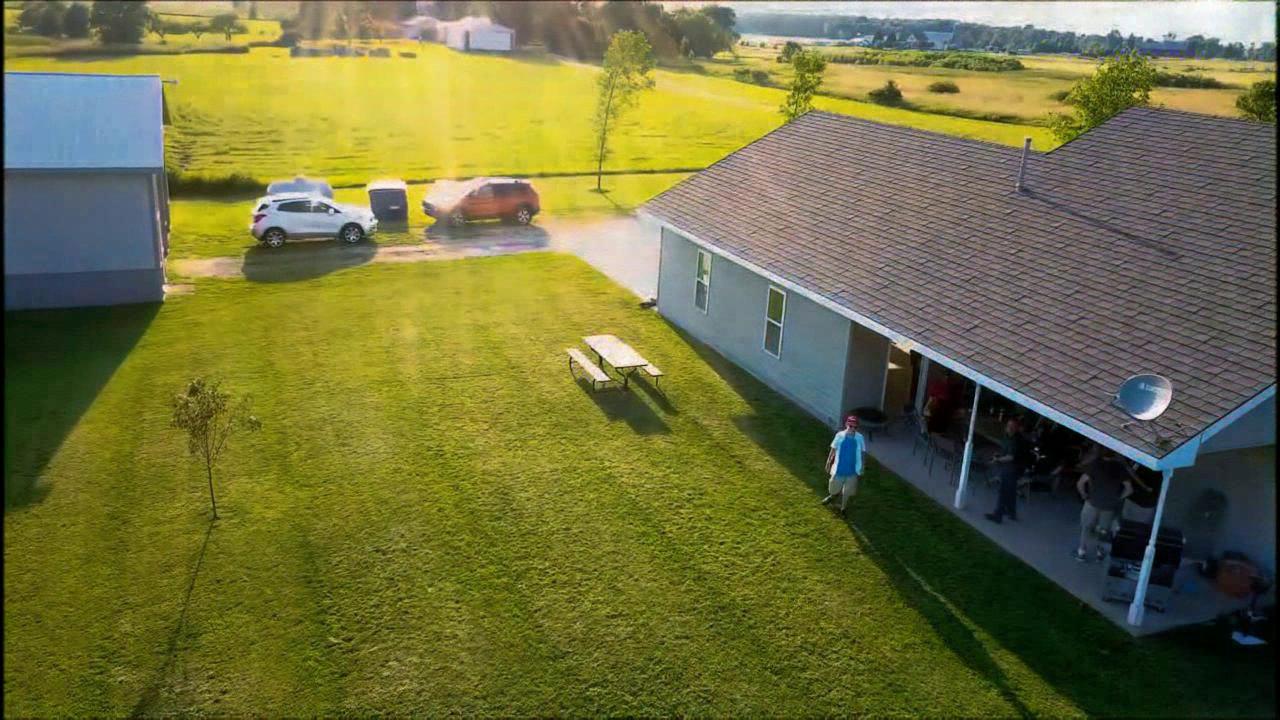}
    \subcaption*{LECVCM~\cite{algoLECVCM}}
\end{subfigure}
\begin{subfigure}[t]{0.15\textwidth}
    \includegraphics[width=1\textwidth]{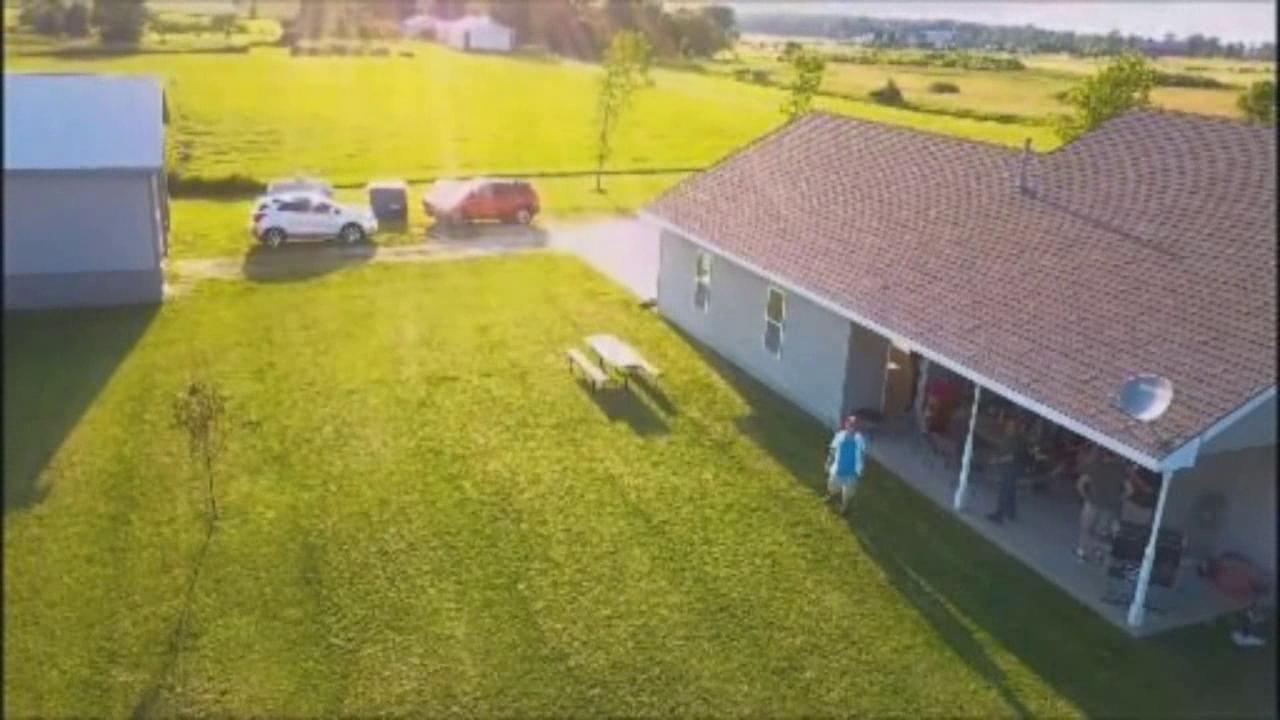}
    \subcaption*{IAGC~\cite{algoIAGC}}
\end{subfigure}
\quad
\begin{subfigure}[t]{0.15\textwidth}
    \includegraphics[width=1\textwidth]{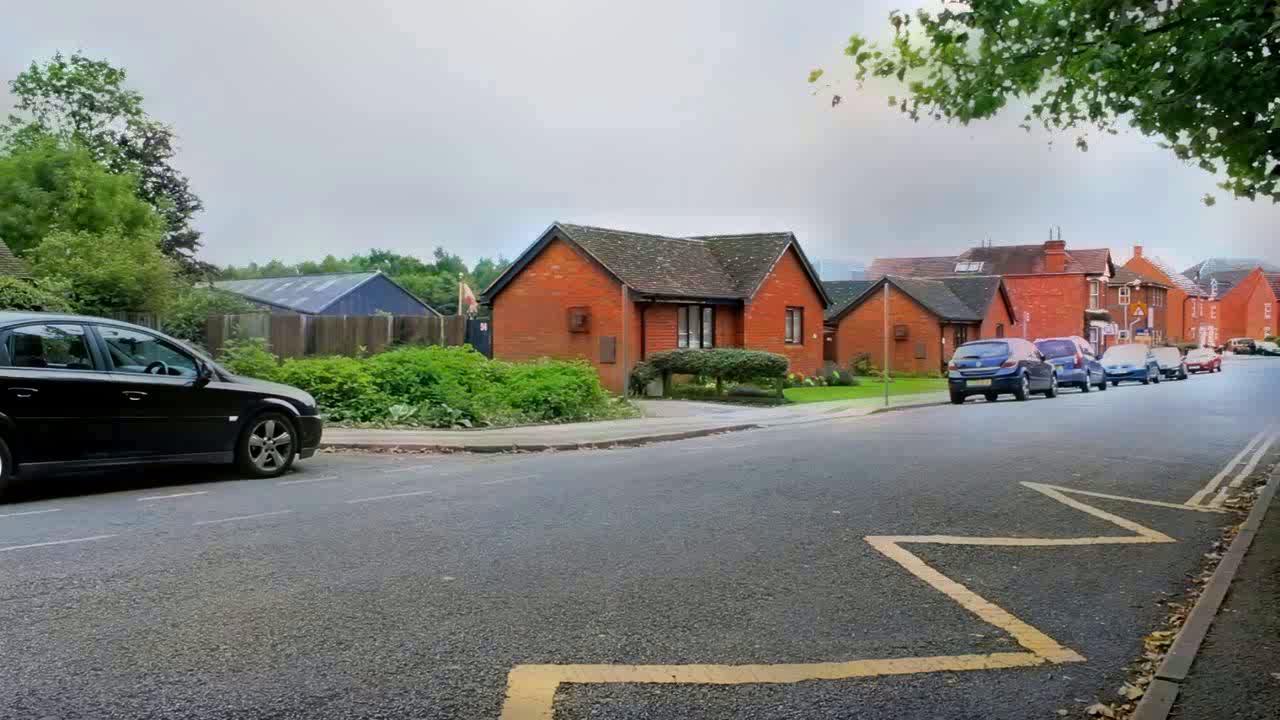}
    \subcaption*{LMSPEC~\cite{algoLMSPEC}}
\end{subfigure}
\begin{subfigure}[t]{0.15\textwidth}
    \includegraphics[width=1\textwidth]{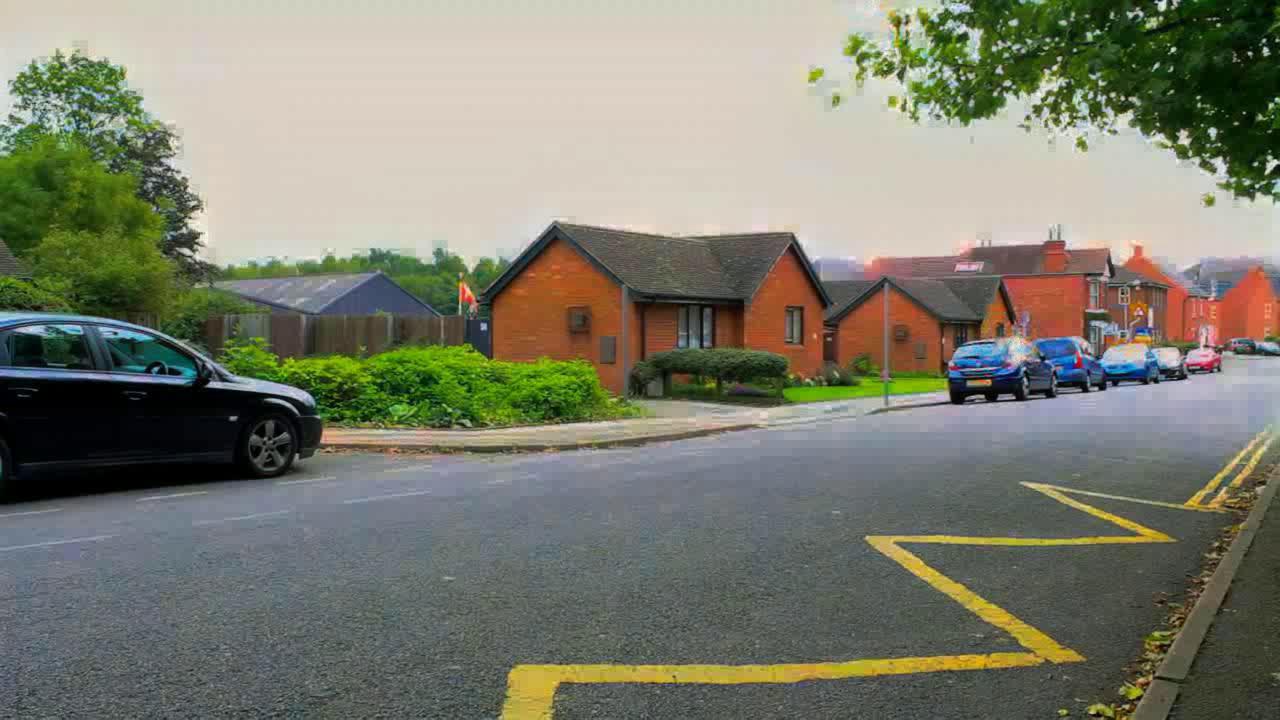}
    \subcaption*{LECVCM~\cite{algoLECVCM}}
\end{subfigure}
\begin{subfigure}[t]{0.15\textwidth}
    \includegraphics[width=1\textwidth]{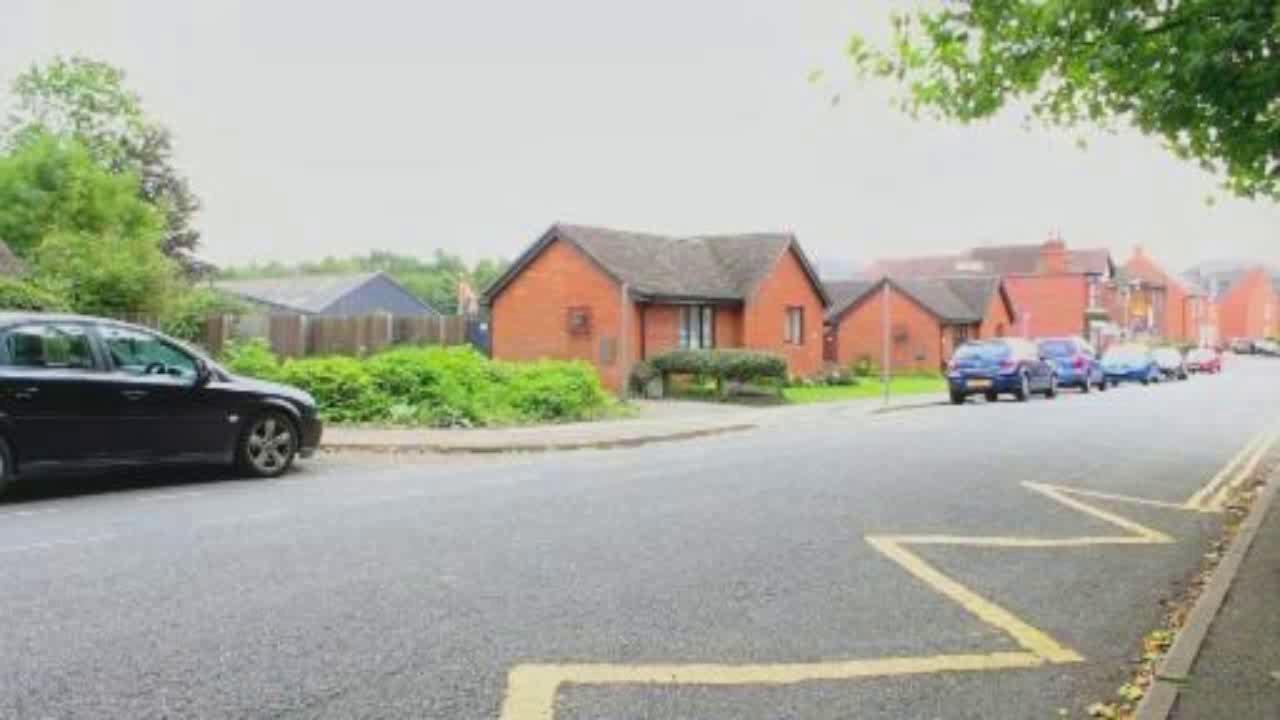}
    \subcaption*{IAGC~\cite{algoIAGC}}
\end{subfigure}
\begin{subfigure}[t]{0.15\textwidth}
    \includegraphics[width=1\textwidth]{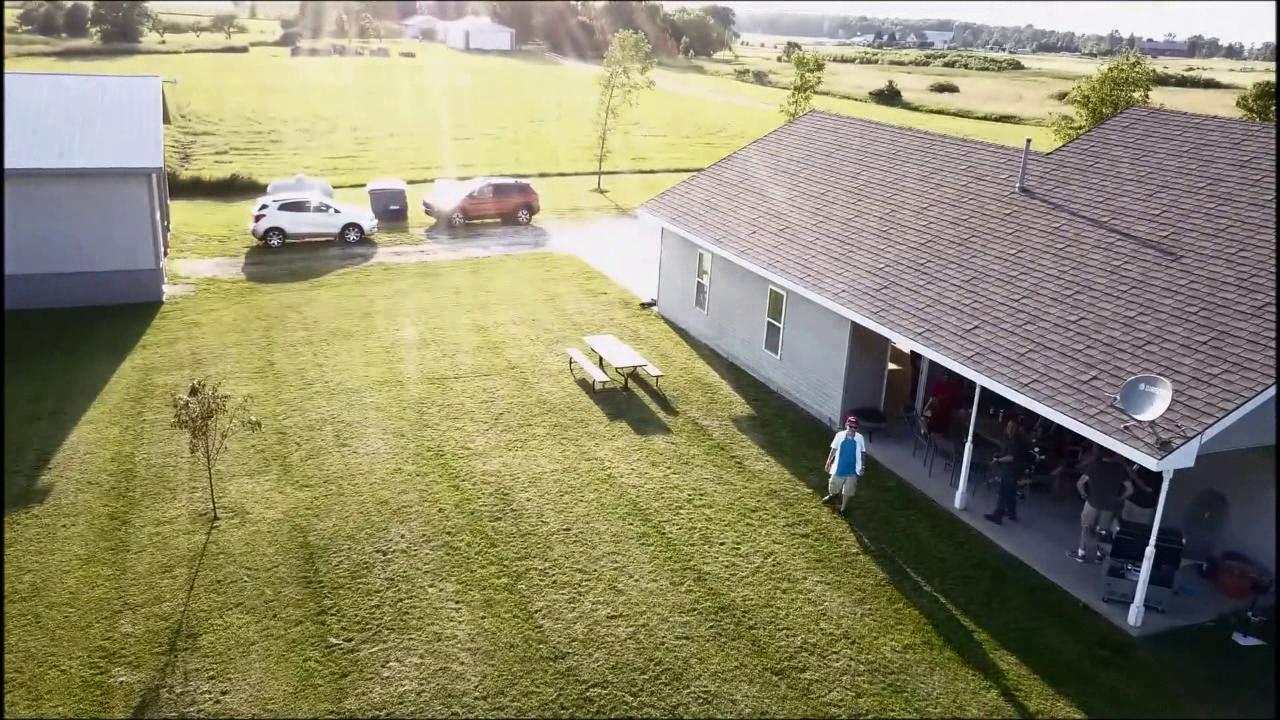}
    \subcaption*{LCDP-Net~\cite{algolcdpnet}}
\end{subfigure}
\begin{subfigure}[t]{0.15\textwidth}
    \includegraphics[width=1\textwidth]{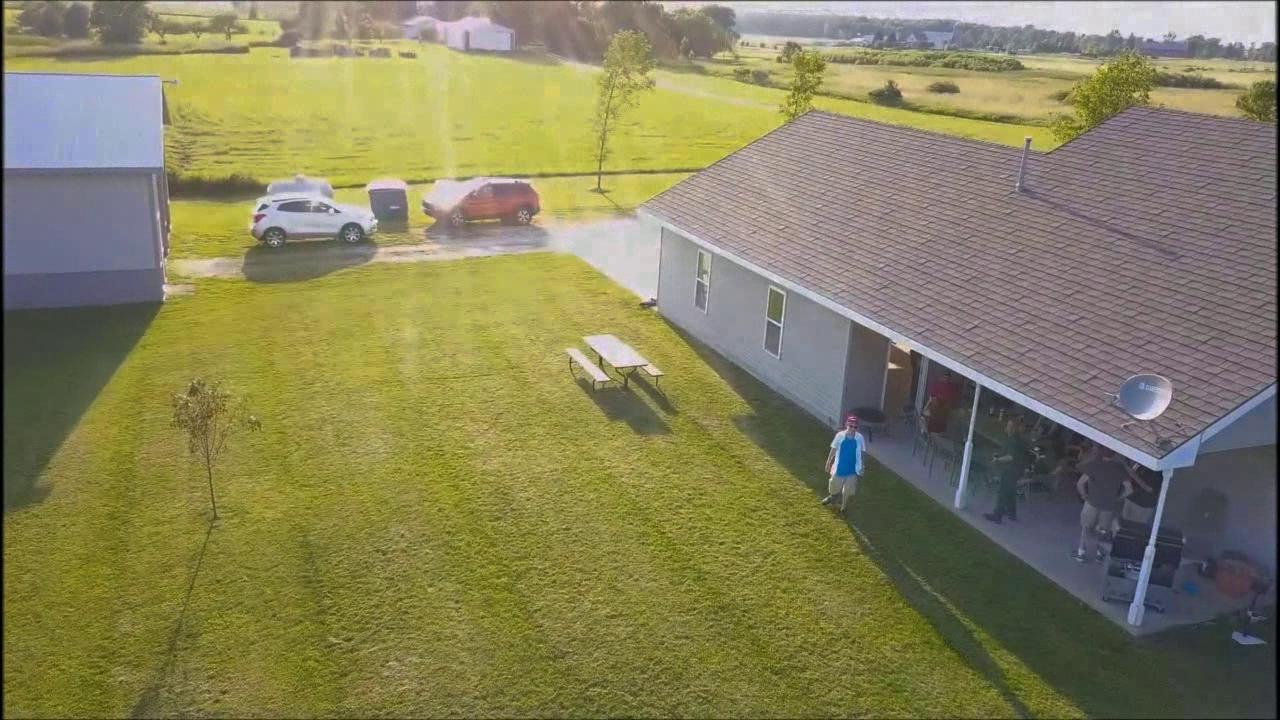}
    \subcaption*{ECMEIQ~\cite{algoECMEIQ}}
\end{subfigure}
\begin{subfigure}[t]{0.15\textwidth}
    \includegraphics[width=1\textwidth]{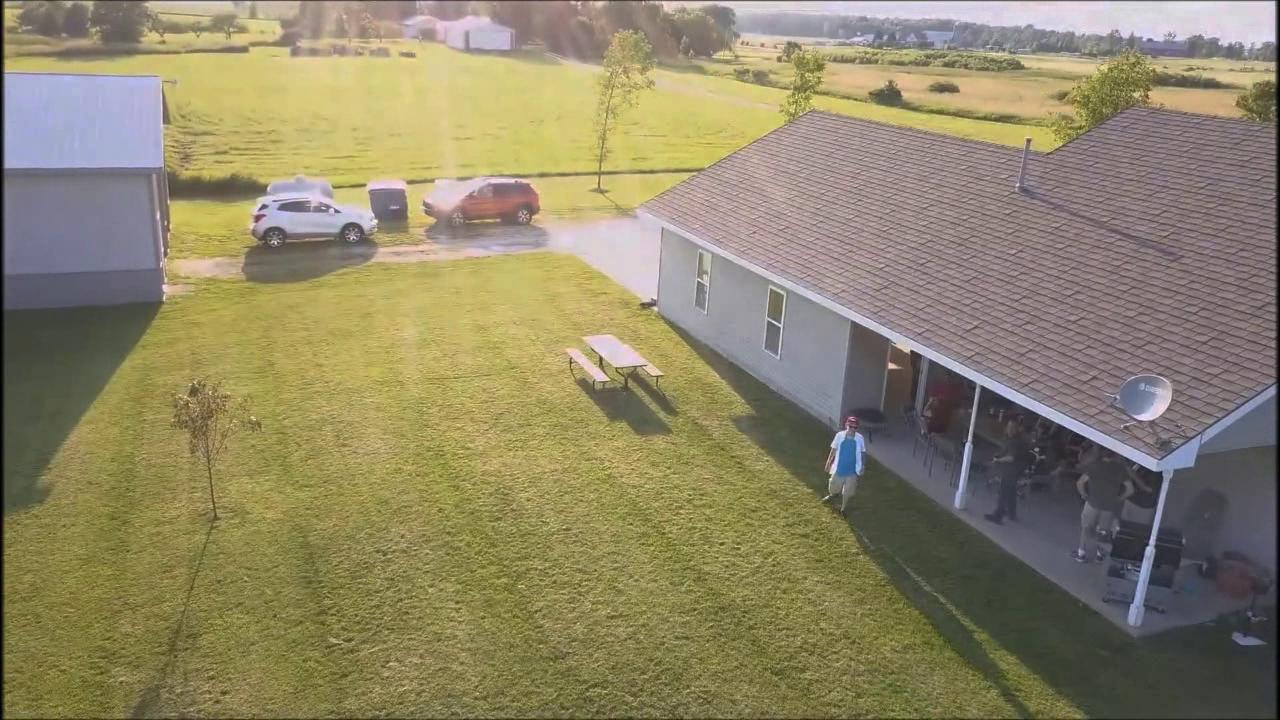}
    \subcaption*{PSE-Net~\cite{algopsenet}}
\end{subfigure}
\quad
\begin{subfigure}[t]{0.15\textwidth}
    \includegraphics[width=1\textwidth]{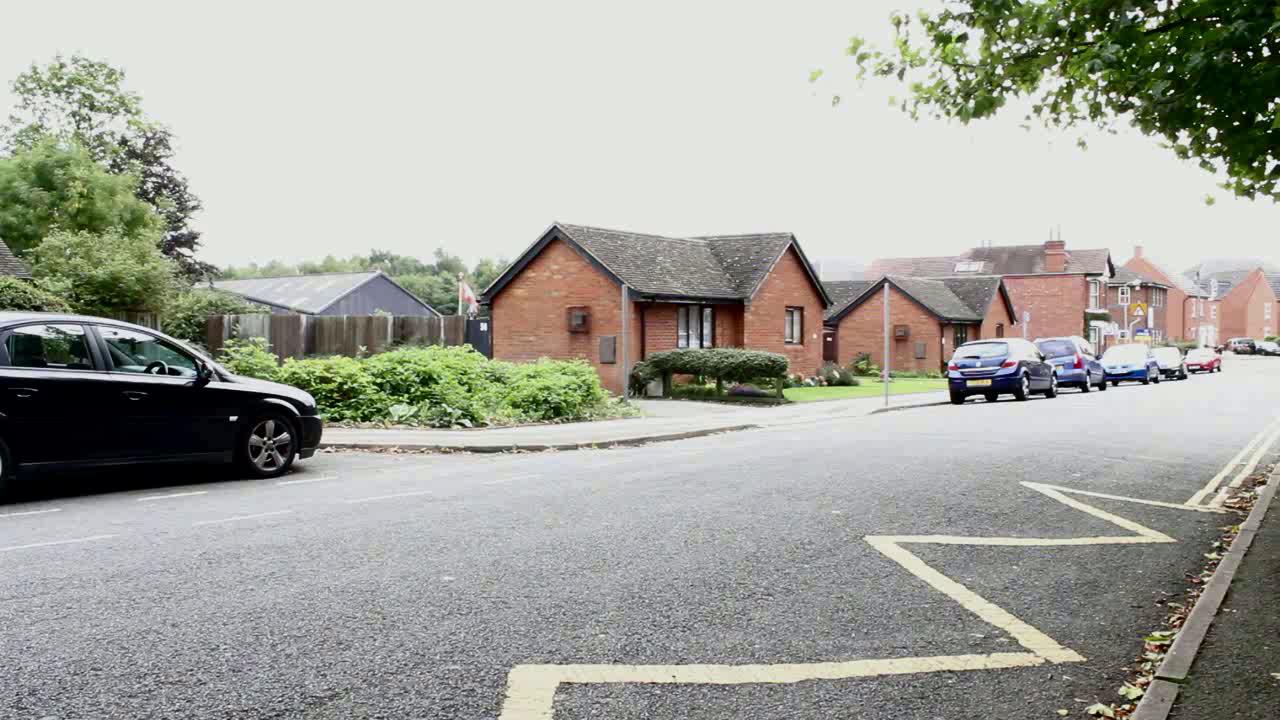}
    \subcaption*{LCDP-Net~\cite{algolcdpnet}}
\end{subfigure}
\begin{subfigure}[t]{0.15\textwidth}
    \includegraphics[width=1\textwidth]{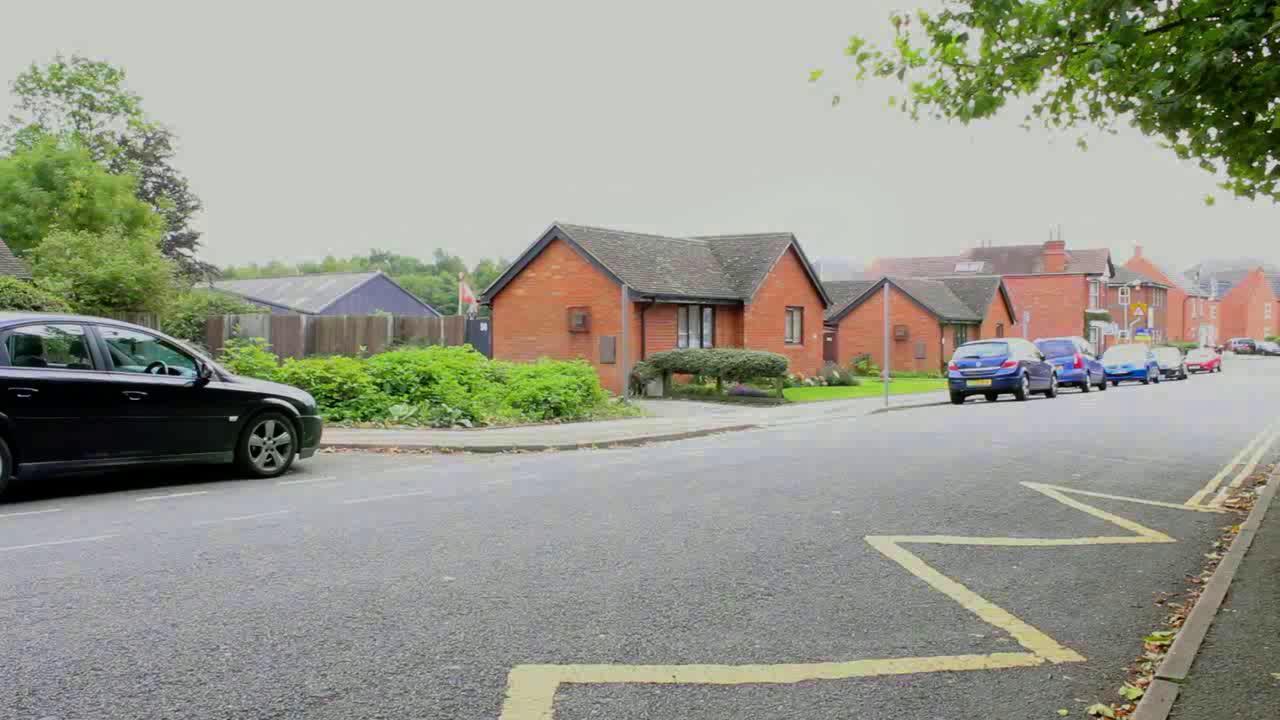}
    \subcaption*{ECMEIQ~\cite{algoECMEIQ}}
\end{subfigure}
\begin{subfigure}[t]{0.15\textwidth}
    \includegraphics[width=1\textwidth]{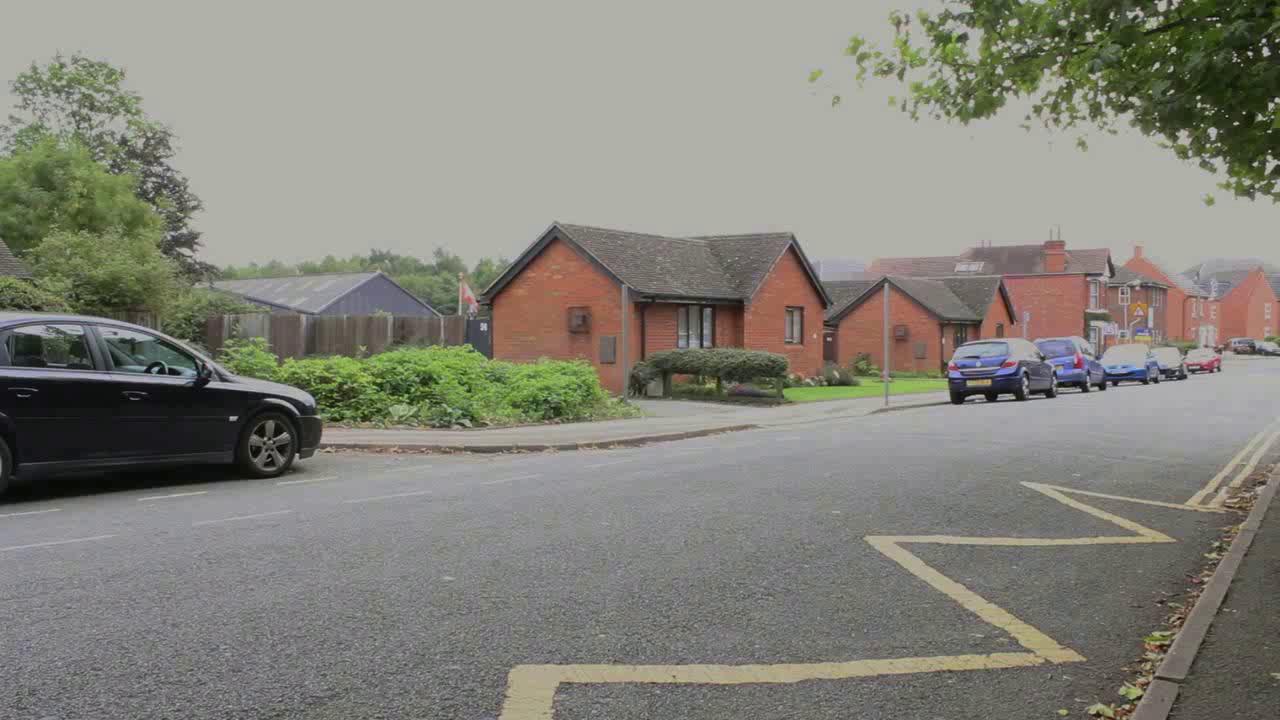}
    \subcaption*{PSE-Net~\cite{algopsenet}}
\end{subfigure}
\caption{{Representative frames of two original over-exposed videos and their corresponding recovered videos.}}
\label{Frames}
\end{figure*}

\subsection{CLIP Model}

CLIP~\cite{clipmodel} (Contrastive Language-Image Pre-Training) is a neural network trained on a variety of image-text pairs. Given an image, CLIP can predict the most relevant text snippet with the instruction from natural language without directly optimizing for the task, similarly to the zero-shot capabilities of GPT-2 and 3~\cite{modelGPT2, modelGPT3}. Its simplicity and effectiveness are demonstrated through impressive outcomes in zero-shot text-image retrieval, classification, text-to-image generation guidance, open-domain detection, segmentation, and of course, quality assessment. CLIP-IQA~\cite{vqaclipiqa}, as well as BIQA~\cite{modelBIQA}, have proven the effectness of CLIP in assessing the quality of a certain image with delicately designed prompts. It can be seen that the strength of CLIP could also be utilized in evaluating the quality of videos.

\subsection{NR-VQA Models}

The traditional and naive NR-VQA~\cite{NRVQA} models are based on handcrafted features. These handcrafted features, including spatial features, temporal features, statistical features, etc., can be extracted to learn the quality scores of videos. For example, V-BLIINDS~\cite{vqaVBLIINDS} builds a Natural Scene Statistics (NSS) module to extract spatial-temporal features and a motion module to quantify motion coherency. The core of TLVQM~\cite{vqaTLVQM} is to generate video features in two levels, where low complexity features are extracted from the full sequence first, and then high complexity features are extracted in key frames which are selected by utilizing low complexity features. VIDEVAL~\cite{vqaVIDEVAL} combines existing VQA methods together and proposes a feature selection strategy, which can choose appropriate features and then fuse them efficiently to predict the quality scores of videos.

With the rapid pace of technological advancements, VQA models based on deep learning have progressively emerged as the prevailing trend. For example, based on a pre-trained DNN model and Gated Recurrent Units (GRUs), VSFA~\cite{vqaVSFA} reflects the temporal connection between the semantic features of key frames well. BVQA~\cite{vqaBVQA} and Simple-VQA~\cite{vqasimple} further introduce motion features extracted by the pre-trained 3D CNN models. Wang et al.~\cite{vqaRFvqa} propose a DNN-based framework to measure the quality of UGC videos from three aspects: video content, technical quality, and compression level. FAST-VQA~\cite{vqafastvqa} creatively introduces a Grid Mini-patch Sampling to generate fragments, and utilizes a model with Swin-Transformer~\cite{modelswintransformer} as the backbone to extract features efficiently from these fragments. RAPIQUE~\cite{vqaRAPIQUE} leverages quality-aware statistical features and semantics-aware convolutional features, which first attempts to combine handcrafted and deep-learning-based features. 

Lately, Multimodal Large Language Models (MLLM) have gained researchers' attention for their incredible performance in multimodal learning. MaxVQA~\cite{vqamaxvqa} combines CLIP, DOVER~\cite{vqadover}, and Fast-VQA to extract features from the input videos, making the process of VQA explainable. Q-Align~\cite{vqaqalign} borrows the strength of LLaVA~\cite{llavamodel1, llavamodel2, llavamodel3} as well as LoRA~\cite{modellora} to evaluate the quality of all image and video related contents. With well-designed prompts and powerful inference abilities, Q-Align outperforms many deep-learning-based methods in many downstream tasks.

While prior VQA models are designed for general UGC videos without exception, our model focuses on VEC quality assessment exclusively. We also utilize the 
multimodal large laguage model CLIP to enhance the assessment accuracy, stability and interpretability of our model.

\section{DATASET PREPARATION}

\subsection{Video Collection}

\begin{algorithm*}
\captionsetup{justification=centering} 
\caption{Video Quality Assessment Process}
\label{rules}
\begin{algorithmic}
\State \textbf{Grading Scale:} \\
The scale is centesimal with video quality divided into five categories:
\textit{Very Poor (0-20)}, \textit{Poor (20-40)}, \textit{Average (40-60)}, \textit{Good (60-80)}, \textit{Excellent (80-100)}.
\State \textbf{Key Grading Factors:} \\
Primary factors are \textit{Video Brightness} and \textit{Stability of Brightness}. Secondary factor is \textit{Presence of Visual Noise}.
\State \textbf{Assessment Process:}\\
1. Determine initial category from the five based on whether overall brightness affects viewing.\\
2. Adjust category based on brightness stability:\\
    \begin{itemize}[label={}]
    \item Minor flickering results in a downgrading by 2-3 categories from the initial grade.
    \item Severe flickering results in categorizing as \textit{Very Poor} (0-20) directly.
    \end{itemize}\\
3. Adjustments based on visual noise:\\
    \begin{itemize}[label={}]
    \item\textit{Severe distortion:} Directly categorized as \textit{Very Poor} (0-20).
    \item\textit{Significant noise distortion:} Downgrade by 1-2 categories from the initial grade.
    \item\textit{Minor noise distortion:} Score in the lower half range of the initial category.
    \item\textit{No distortion:} Score in the upper half range of the initial category.
    \end{itemize}
\end{algorithmic}
\end{algorithm*}

LLVE-QA~\cite{baselightvqa} has already constructed a well-equipped dataset to assess LLVE. However, in the field of over-exposure recovery, as opposite to low-light enhancement, there is no such dataset functioning in assessing the quality of recovered over-exposed videos. To fill in the gap, we additionally gather 205 such videos from diverse sources including KoNVID-1K~\cite{datakonvid}, VDPVE~\cite{datavdpve}, YouTube-UGC~\cite{datayoutubeugc}, LIVE-VQC~\cite{dataLiveVQC} dataset, and UGC-video websites including \url{https://youtube.com} and \url{https://vimeo.com}. These selected videos feature a wide range of content and brightness levels. To recover the exposure of these videos, we apply 10 different over exposure recovery algorithms on the collected videos: ACE~\cite{algoACE}, AGCCPF~\cite{algoAGCCPF}, BPHEME~\cite{algoBPHEME}, DIEREC~\cite{algoDIEREC}, LMSPEC~\cite{algoLMSPEC}, LECVCM~\cite{algoLECVCM}, IAGC~\cite{algoIAGC}, LCDP-Net~\cite{algolcdpnet}, ECMEIQ~\cite{algoECMEIQ}, PSE-Net~\cite{algopsenet}, and the commercial software CapCut~\cite{algocapcut}. This process yields 2,253 enhanced videos. Along with the original 205 over-exposed videos, we form the Over-Exposure Video Recovery Quality Assessment (OEVR-QA) dataset, which is constructed symmetrically to LLVE-QA~\cite{baselightvqa}. By integrating OEVR-QA with LLVE-QA, we form the comprehensive Video Exposure Correction Quality Assessment (VEC-QA) dataset, which is of significance for developing well-performed VQA models for exposure correction. To the best of our knowledge, VEC-QA is the first dataset in this kind that aims explicitly at assessing the performance of VEC algorithms. Representative frames from two original videos and their corresponding enhanced versions are illustrated in Fig. \ref{Frames}. 

\begin{figure}[t]
\centering
    \includegraphics[width=\linewidth]{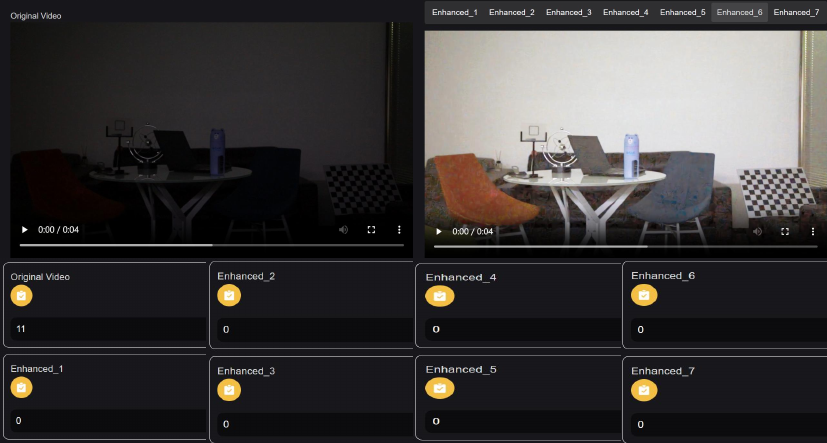}
\caption{The scoring interface during the subjective experiment.}
\label{inter}
\end{figure}
\subsection{Subjective Experiment}

\begin{figure*}[t]
\centering
\includegraphics[width=\textwidth]{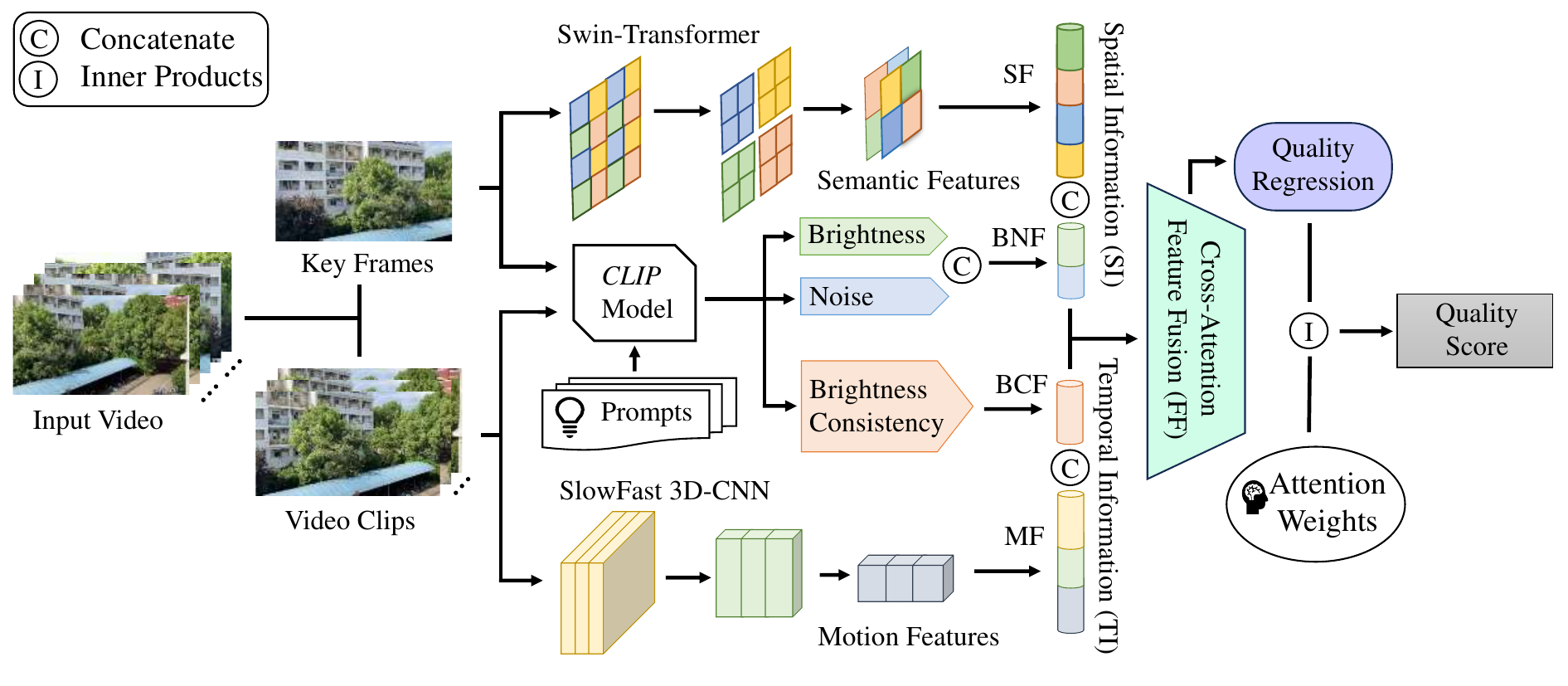}
\caption{Framework of Light-VQA+. The model contains the spatial and temporal information extraction module via CLIP~\cite{clipmodel}, the feature fusion module via cross-attention, and the quality regression module with HVS~\cite{626903}. Concretely, Spatial Information contains semantic, brightness, and noise features, while Temporal Information contains motion and brightness consistency features. [Key: SF: Semantic Features; BNF: Brightness \& Noise Features; BCF: Brightness Consistency Features; MF: Motion Features]}
\label{Network}
\end{figure*}
A high-quality dataset is a prerequisite for a well-performed model.
To collect accurate annotations of the VEC-QA, we invite 21 experienced data labeling evaluators for a subjective experiment. Participants are tasked with rating video quality on a scale from 0 to 100, where a higher score indicates better quality. To ensure evaluators focus on perceptual quality rather than content, we guild the evaluators with a custom scoring process, details of which are shown in Algorithm \ref{rules}. Participants are first required to evaluate the original video and then assess its corrected counterparts placed aside in a random order. Using this random order rather than a fixed one can more accurately reflect the perceptual differences attributed to exposure correction in the subjective experiment. To achieve this, we have developed a customized scoring interface shown in Fig. \ref{inter}.

After conducting the subjective experiment, we gathered a total of 51,660 scores, calculated as $21 \times 205 \times 12$. To mitigate subjective discrepancies between the LLVE-QA dataset and our newly collected OEVR-QA dataset, we applied a linear transformation to both datasets as follows:
\begin{equation}
MOS_{i}^{'} = 100\times \frac{MOS_{i}-MOS_{min}}{MOS_{max}-MOS_{min}},
\end{equation}
where $MOS_{i}^{'}$, $MOS_i$, $MOS_{max}$, and $MOS_{min}$ represent the transformed, original, the maximum, and the minimum Mean Opinion Score (MOS) respectively. This transformation aims to symmetrize the scores of under/over-exposed videos relative to their brightness levels and reduce the subjective errors.

\begin{figure*}[ht]
\centering
\includegraphics[width=\textwidth]{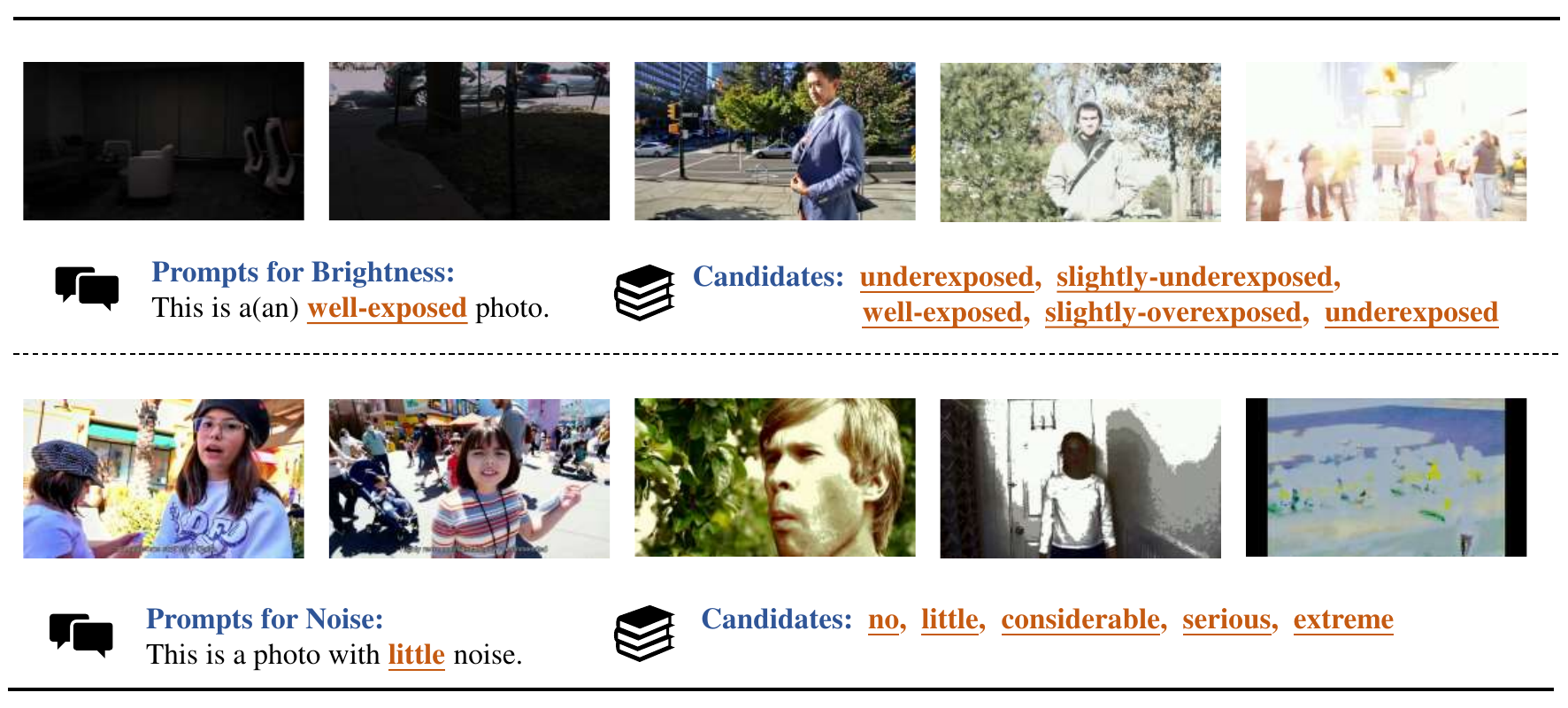}
\caption{The prompts utilized for extracting the brightness \& noise information in Light-VQA+.}
\label{Prompts}
\end{figure*}

\section{Proposed Method}

Benefiting from the constructed VEC-QA dataset as well as the success of Light-VQA~\cite{baselightvqa}, we further update the model into Light-VQA+, a multi-dimensional quality assessment model specialized in assessing exposure-corrected videos with vision-language guidance. This enhanced model explores the ability of multimodal large language models, whose structure is depicted in Fig. \ref{Network}. 
{Among the open MLLMs with the ability of image-text interaction, CLIP and LLaVA~\cite{llavamodel1, llavamodel2, llavamodel3} are the top performers to date. LLaVA employs the image-text pairing capabilities of CLIP, enhanced by integration with the strengths of LLaMA~\cite{touvron2023llama} in language processing. However, its capacity for image processing remains identical to that of CLIP. Given that our model does not require natural language generation, we choose to directly employ CLIP instead of LLaVA to conserve computational resources.}
At the beginning, we divide the input video into 8 clips. Then, on each video clip, with the purpose of assessing them from a more comprehensive perspective, we extract both the spatial and temporal information for evaluation. To be specific, the Spatial Information (SI) is composed of deep-learning-based semantic features via Swin-Transformer~\cite{modelswintransformer} and CLIP~\cite{clipmodel}-captured brightness and noise features, while the Temporal Information (TI) consists of deep-learning-based motion features via SlowFast Model~\cite{modelslowfast} and CLIP-captured brightness consistency features. These extracted features need a fusion procedure to create a quality-aware representation since they are generated from different perspectives. To this end, a cross-attention module~\cite{crossattention} is applied to integrate them, followed by a two Fully-Connected (FC) layers that is designed for regressing these fused features into a unified video quality score for each clip. Last but not least, after obtaining a quality score for each video clip, {we assign these video clips a set of trainable weights} that are autonomously learned to be consistent with HVS~\cite{626903}. {Based on HVS weights, the weighted average of different video clips is calculated to predict the final assessment score.}

\subsection{Spatial Information}

Spatial Information (SI) mainly focuses on a certain video from the intra-frame perspective. Since the adjacent frames of a video contain plenty of redundant contents, spatial information shows the extreme sensitivity to the video resolution and is not quite relevant to the video frame rate. Therefore, in order to reduce the computational complexity, from each video clip, we select one key frame to calculate the spatial information. In Light-VQA+, we design two branches to simultaneously extract features. Concretely, one is for deep-learning-based features, which contain rich semantic information. The features extracted by it are denoted as SF (Semantic Features). The other one is for CLIP-captured features, which contain specifically designed brightness and noise features, denoted as BNF (Brightness and Noise Features).

Swin-Transformer has achieved more excellent performance than traditional CNNs. For deep-learning-based features, we utilize the semantic information extracted from the last two stages of the pre-trained Swin-Transformer:
\begin{equation}
\begin{split}
    \alpha_j &=  GAP(VF_i^j), j\in \{ 1,2\},
    \\
    SF_i &=  \alpha_1 \oplus \alpha_2, i \in \{ 1, ..., k \}.
\end{split}
\end{equation}

Here for the $i$-th sampled key frame of a video, $SF_i$ represents the extracted semantic features, $\oplus$ denotes the concatenation operation, $GAP(\cdot)$ indicates the Global Average Pooling operation, $VF_i^j$ refers to the feature maps from the $i$-th key frame produced by the $j$-th last stage of the Swin-Transformer, and $\alpha_j$ stands for the features from post-average pooling.

To provide our model with high-quality vision-language guidance, based on the designed a set of prompts, we utilize the pre-trained CLIP model to extract brightness and noise features. The CLIP model takes an image and a series of text-prompts, and then outputs the matching probability of the image to each prompt. Since the image size  of standard input for CLIP ($224\times 224$) is incompatible with our video size ($1,280\times 720$), and simply resizing the frames on such large scale could cause severe information loss, it is essential to properly pre-process the size of input images. To this end, we first resize the image slightly to $1,120\times 672$, and then slice one input frame into 15 sub-images to $(224\times 5) \times (224\times 3)$ that fit the input size for CLIP.

Equally important is the design of prompts as vision-language guidance for our model, which is composed of two branches: brightness and noise. The training process of CLIP involves millions of text-image pairs with various textual prompts. To avoid the inaccuracy caused by using only single prompt, we design a series of prompts with progressive descriptions related to the frame brightness and noise when applying CLIP. To this end, we design 5 parallel sets of prompts as illustrated in Fig. \ref{Prompts}. To be specific, the usage of these prompts is as follows:
\begin{itemize}
\item \textbf{Brightness:} This is a(an) \textless sys\_hint\textgreater\space photo.
\item \textbf{Noise:} This is a photo with \textless sys\_hint\textgreater\space noise.
\end{itemize}
With the assistance of image and text encoder in CLIP, the probabilities of the image that match the prompts are obtained via Softmax. In total, we obtain $10$ probabilities from $5\times 2$ prompts and utilize them as the brightness and noise features. 
By concatenating the brightness and noise features extracted from the 15 ($5\times 3$) sub-images, we obtain a \(150\)-dimensional \((15 \times 5 \times 2)\) feature vector:
\begin{equation}
    BN_i = BF_i \oplus NF_i,  
\end{equation}
where $BN_i$ denotes the combined brightness and noise information of a key frame, with $BF_i$ and $NF_i$ specifically representing the brightness and noise features.

\begin{figure}[t]
\centering
    \includegraphics[width=\linewidth]{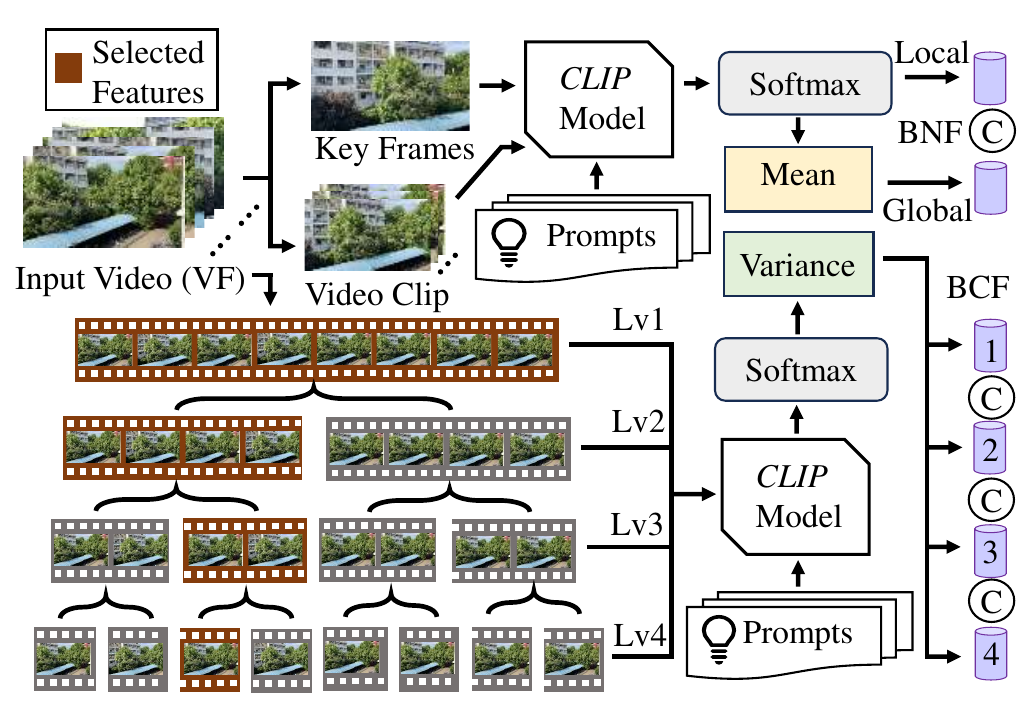}
\caption{The structure of extracting the BNF and BCF from the third video clip. The ``sub-videos'' $V^{p}_{l,k}$ squared in red stand for the features that cover the third video clip. [Key: VF: Video Frames; BNF: Brightness \& Noise Features; BCF: Brightness Consistency Features]}
\label{BCNF}
\end{figure}

Furthermore, in order to retain more information, we also extract the brightness and noise features on all frames from a video clip and then average them to obtain two overall features, one for brightness and the other for noise. By integrating the features from key frame with those from entire video clip, our model enables considering the overall details of the video. The procedure of extracting the brightness and noise features are demonstrated in Fig. \ref{BCNF}.

The last step is to concatenate the local and global features together:
\begin{equation}
    BNF_i = BN_i \oplus \overline{BN_i},
\end{equation}
where $BNF_i$ symbolizes the brightness and noise features for a video clip, while $\overline{BN_i}$ represents the averaged brightness and noise features across the entire video clip. The features from key frame and from entire video clip focus on local and global video attributes respectively. By combining them, we enhance the representation of frame brightness and noise. The final spatial information is formulated as:
\begin{equation}
    SI_i = SF_i \oplus BNF_i,
\end{equation}
where $SI_i$ indicates the ultimate spatial information of the $i$-th video clip.

\subsection{Temporal Information}

Temporal Information (TI) mainly evaluates a video from the perspective of inter-frame. Different from SI, TI is extremely susceptible to video frame-rate variations but not sensitive to resolution. Note that each video in our VEC-QA dataset can be uniformly split into 8 clips. Therefore, in order to preserve adequate temporal information while reducing computational complexity, we extract TI on all the video clips (denoted as $V_{1-8}$) separately. Concretely, similar to the extraction of SI, we design two branches to obtain deep-learning-based and CLIP-captured features respectively. One is for Motion Features (MF), and the other is for Brightness Consistency Features (BCF), {serving as a significant factor of temporal consistency.}

For deep-learning-based features, we employ a pre-trained Slow-Fast network~\cite{modelslowfast} to extract the MF from each video clip:
\begin{equation}
    MF_i = \phi (V_i),
\end{equation}
where $V_i$ represents the $i$-th video clip, $\phi (\cdot)$ symbolizes the extraction operation for motion feature, and $MF_i$ denotes the motion features extracted from the $i$-th video clip.

To extract the BCF, we still choose to harness the capabilities of CLIP. Since BCF only concerns the brightness information, the same brightness-related prompts shown in Fig.~\ref{Prompts} are employed again as the vision-language guidance:
\begin{itemize}
\item \textbf{Brightness:} This is a(an) \textless sys\_hint\textgreater\space photo.
\end{itemize}

To be specific, just as the BNF, we firstly slice a frame into 15 sub-images with the size of $224\times 224$. 
Then we use the brightness prompts to obtain a 75-dimensional ($15\times 5$) brightness feature for each frame.
After extracting the features from all frames in a video clip, we calculate the variance on each dimension, resulting in a feature with 75 dimensions. {We denote this process as $BCF(\cdot)$.}

However, utilizing the brightness variance within video clip to depict brightness consistency for entire video may not be sufficient. For example, the brightness change is not severe within a video clip but prominent between video clips. In such scenario, the aforementioned method may perform poorly.

{To address this limitation, we implement a novel 4-level view strategy that focuses more comprehensively on the entire video rather than isolated video clips. We suppose the video contains $8n$ frames and employ a systematic approach to break down and reassemble the video into various ``sub-videos'' across different levels. For each level $l$, the video is divided into $2^{l-1}$ parts, with each part generating $2^{4-l}$ ``sub-videos''. Each ``sub-video" captures frames with a stride of $2^{4-L}$, starting from the frame index $k$:}
\begin{equation}
    V^{p}_{l, k} = \{ VF_{k + 2^{4-l}\cdot(t+pn)} \},
\end{equation}
where $l$ is the level index (1 to 4), $k \in \{1, 2, ..., 2^{4-l}\}$, $t \in \{0, 1, ..., n-1\}$, and $p \in \{0, 1, ..., 2^{l-1}-1\}$. $VF_{x}$ denotes the $x$-th frame of the entire video.
\begin{itemize}
    \item Level 1 (Lv1): The video is not divided, and 8 ``sub-videos'' are created by sampling every 8 frames.
    \item Level 2 (Lv2): The video is split into 2 parts, with each part generating 4 ``sub-videos'' by sampling every 4 frames.
    \item Level 3 (Lv3): The video is divided into 4 quarters, with each quarter generating 2 ``sub-videos'' by sampling every 2 frames.
    \item Level 4 (Lv4): The video is divided into 8 segments, with each segment generating a ``sub-video'' by sampling every frame. Note that in this level, the \textbf{``sub-videos''} are equal to  \textbf{video clips}.
\end{itemize}
In each video segment, we calculate the brightness consistency of ``sub-videos'' in it and average the variances of them to obtain the features that describe the overall video's brightness consistency at different levels of granularity:
\begin{equation}
    BCF_{l,p}=Mean(BCF(V^{p}_{l, k})),
\end{equation}
{where $BCF_{l,p}$ denotes the feature of the $p$-th video segment from Level $l$, and $Mean(\cdot)$ denotes the mean of values corresponding to all instances of $k$. The symbol $BCF(\cdot)$ refers to the method for extracting a 75-dimensional feature via CLIP that quantifies the consistency of brightness in a video.}

Now we have already generated $15$ features ($1+2+4+8$), each focusing on different segments of the entire video. The final BCF for a certain video clip is composed of features from 4 levels. For every level, since there is only one feature that contains the information of current video clip, we only need to select this particular feature that fully covers the video clip. For example, when evaluating the 3-rd video clip, we choose the only feature in Lv1, the first feature in Lv2 since it covers the 3-rd video, the second feature in Lv3 for the same reason, and the third feature in Lv4 that is extracted from the 3-rd video, as shown in Fig. \ref{BCNF}. This multi-level method enhances the extraction of brightness consistency information, providing a more detailed and comprehensive analysis. The final brightness consistency of the given video clip is extracted by concatenating the chosen 4 features:
\begin{equation}
    BCF^i =  BCF_1\oplus BCF_2\oplus BCF_3\oplus BCF_4,
\end{equation}
where the final brightness consistency feature is denoted as $BCF^i$, and the chosen Lv$j$ brightness consistency feature is represented by $BCL_j$. Fig. \ref{BCNF} is an illustration for the aforementioned process. With this novel process, our model can evaluate the video from dynamic perspectives at different levels, and eventually provide a more thorough assessment. The final temporal information is formulated as:
\begin{equation}
    TI_i = MF_i\oplus BCF_i,
\end{equation}
where $TI_i$ indicates the temporal information of the $i$-th video clip.

\subsection{{Feature Fusion Module}}

After acquiring both SI and TI, it becomes important to fuse them to achieve a more comprehensive representation of features, given that they originate from different perspectives. The cross-attention module~\cite{crossattention}, known for its effectiveness in integrating features from disparate sources, plays an important role in many tasks, such as in Stable Diffusion~\cite{modelstablediffusion}. Consequently, in this paper, we employ the cross-attention module as our fusion module to incorporate the SI with TI effectively.
We first normalize the dimensions of the SI and TI by two linear layers, denoted as $\mathcal{A}$ and $\mathcal{B}$, respectively. Following this, we apply a multi-head cross-attention module, denoted as $\mathcal{C}$. Specifically, the fusion process is conducted in two steps to underscore the equal importance of both SI and TI. In the first step, we treat the normalized SI as both keys and values, with the normalized TI serving as the query. In the second step, we reverse these roles: the normalized TI is used as keys and values, while the normalized SI functions as the query. This symmetrical approach allows us to derive two distinct features, emphasizing the balanced integration of both types of information. Subsequently, we concatenate these two features and apply an additional linear layer, denoted as $\mathcal{F}$, to produce the final fused feature.
The overall feature fusion module are represented as follows:
\begin{equation}
    SI_i^{'} = \mathcal{A}(SI_i), \quad TI_i^{'} = \mathcal{B}(TI_i),
\end{equation}
\begin{equation}
    FF_i = \mathcal{F}(\mathcal{C}(SI_i^{'}, TI_i^{'}) \oplus \mathcal{C}(TI_i^{'}, SI_i^{'})),
\end{equation}
where $\mathcal{A}$, $\mathcal{B}$, and $\mathcal{F}$ represent different linear layers, $\mathcal{C}$ represents the multi-head cross-attention module, and $SI_i^{'}$, $TI_i^{'}$, and $FF_i$ represent the normalized SI, TI, and the fused feature for the $i$-th video clip, respectively.
 
\subsection{{Quality Regression}}

\begin{figure}[t]
\centering
\includegraphics[width=\linewidth]{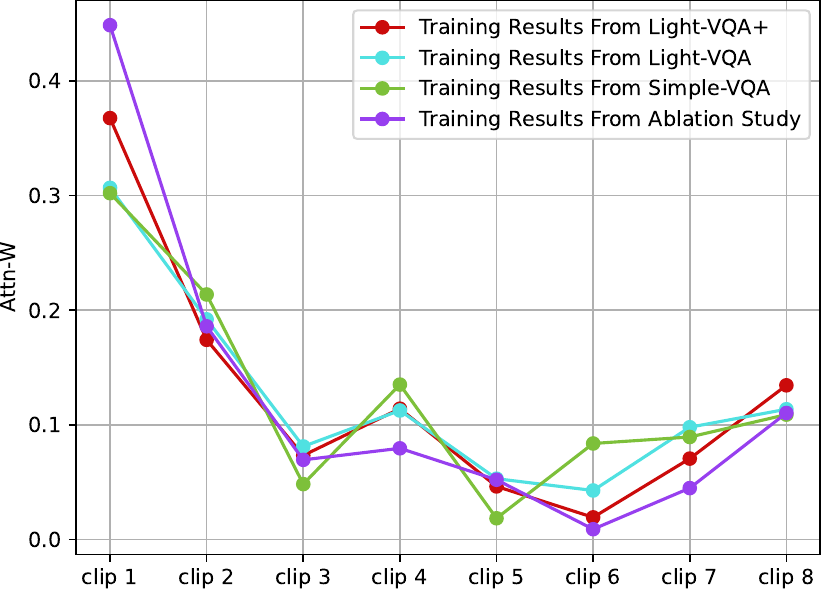}
\caption{{The HVS weights under different models.}}
\label{weights}
\end{figure}

Subsequently, we employ two Fully Connected (FC) layers to regress the fused feature ($FF_i$) into the video quality score:
\begin{equation}
    Q_i = FC(FF_i),
\end{equation}
where $Q_i$ denotes the quality score for the $i$-th video clip. Upon calculating the quality scores for each of the video clips, we introduce eight trainable parameters into our model to perform a weighted average calculation of these scores. These parameters allow the model to learn the relative importance of each video clip in determining the overall video quality, thus allocating more weight to those clips that have a greater impact. 

The final score is derived using the following weighted average:
\begin{equation}
    Q = \frac{\sum\limits_{i=1}^{k} w_i \times Q_i}{\sum\limits_{i=1}^{k} w_i},
\end{equation}
where $Q$ represents the overall quality score of the video, and $w_i$ denotes the weight assigned to the $i$-th video clip's quality score. 

To evaluate the effectiveness of this method, we conducted a series of experiments. Initially, identical {HVS weights} are assigned to all eight video clips. This method was then applied across various models, and we tracked the convergence of their parameters. The results demonstrate that, irrespective of the model used, the final {HVS weights} tend to be similar, as depicted in Figure \ref{weights}. These results are consistent with the HVS that human perception usually pays more attention to the beginning of a video, confirming the effectiveness of our method.

\begin{figure}[t]
	\centering
	
	
	\begin{subfigure}{0.9\linewidth}
		\centering
		\includegraphics[width=\linewidth]{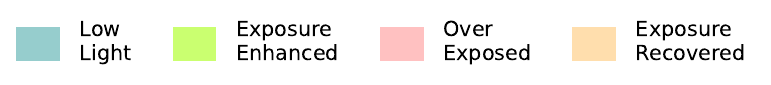}
	\end{subfigure}
	\begin{subfigure}{0.45\linewidth}
		\centering
		\includegraphics[width=\linewidth]{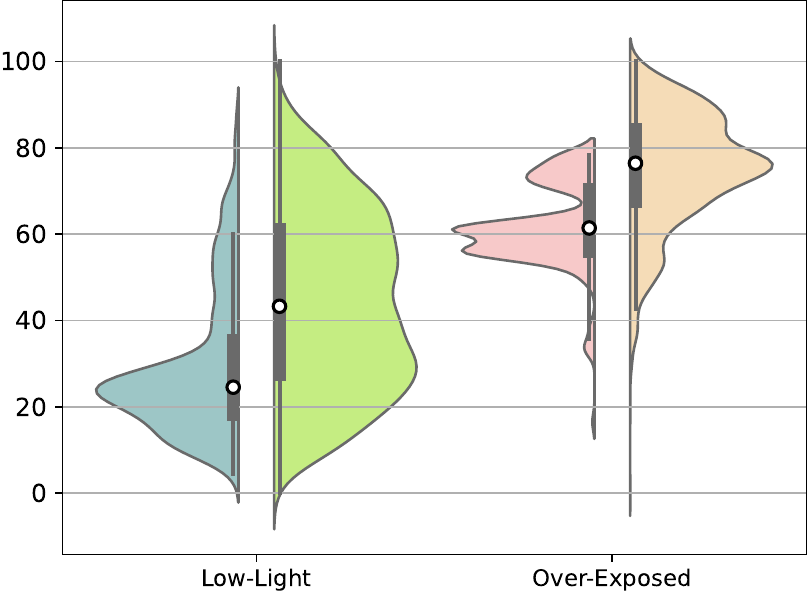}
		\caption{MOS}
	\end{subfigure}%
	\quad
	\begin{subfigure}{0.45\linewidth}
		\centering
		\includegraphics[width=\linewidth]{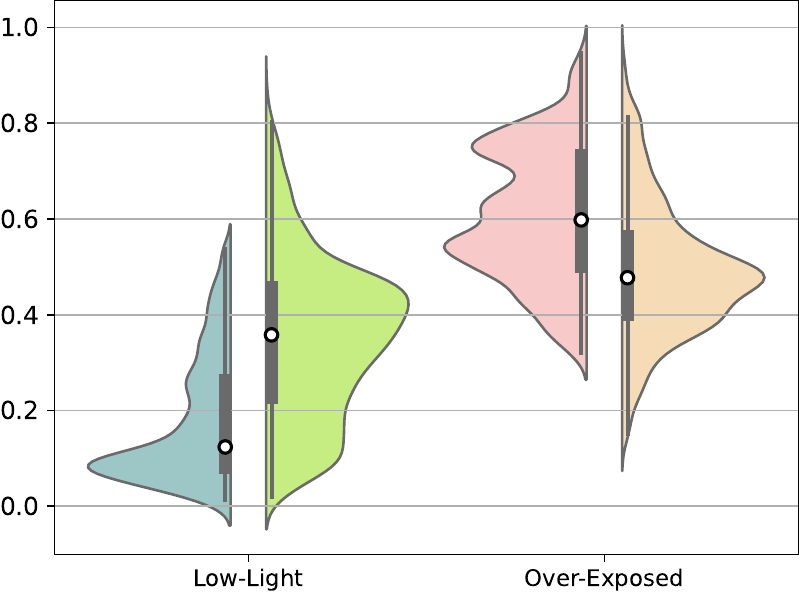}
		\caption{Brightness}
	\end{subfigure}
	\begin{subfigure}{0.45\linewidth}
		\centering
		\includegraphics[width=\linewidth]{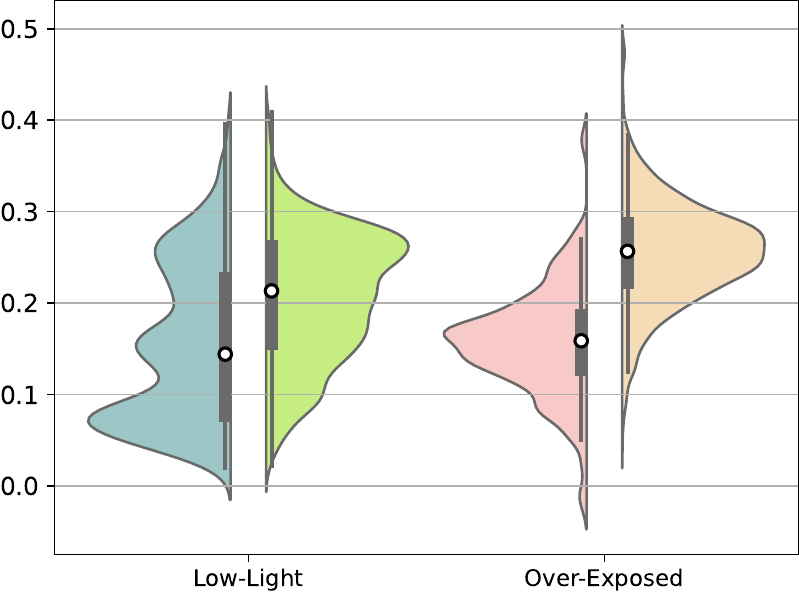}
		\caption{Contrast}
	\end{subfigure}%
	\quad
	\begin{subfigure}{0.45\linewidth}
		\centering
		\includegraphics[width=\linewidth]{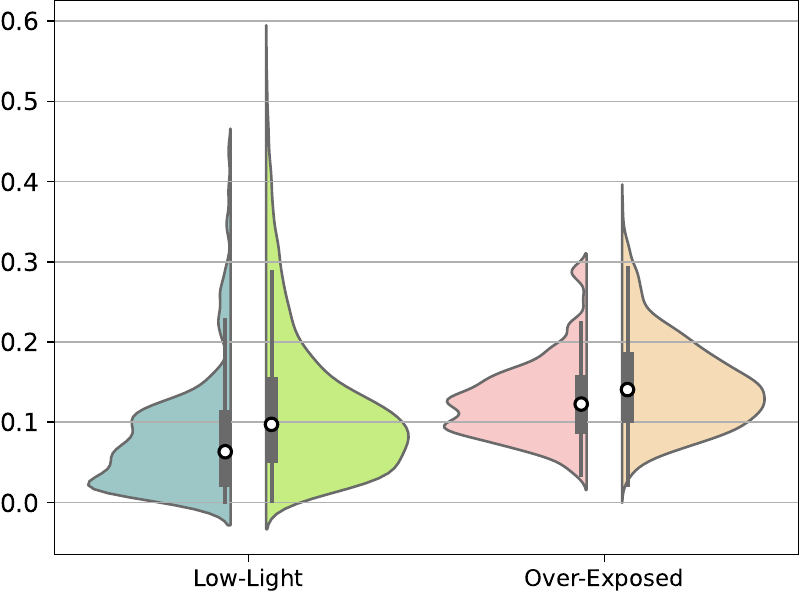}
		\caption{Colorfulness}
	\end{subfigure}
	\caption{Distributions of MOS, brightness, contrast, and colorfulness over the original under/over-exposed videos as well as corresponding exposure corrected videos in our VEC-QA dataset.}
	\label{Indexes}
\end{figure}

\subsection{{Loss Function}}

Our training loss function comprises two components: the Mean Absolute Error (MAE) loss ($L_{mae}$) and the rank loss ($L_{rk}$)~\cite{rankloss}. The MAE loss, a common metric in various deep learning applications, is defined as:
\begin{equation}
    L_{mae} = \frac{1}{N} \sum\limits_{m=1}^N |Q_m - \hat{Q}_m|,
\end{equation}
where $Q_m$, $\hat{Q}_m$ represent the ground-truth and the predicted MOS of the $m$-th video within a batch, and $N$ denotes the batch size.

The rank loss is particularly useful for learning the relative quality among videos, aligning with our goal to evaluate the performance of different VEC algorithms. It is calculated as follows:
\begin{equation}
    L_{rk} = \frac{1}{N^2} \sum\limits_{m=1}^N \sum\limits_{n=1}^N L_{m,n},
\end{equation}

\begin{equation}
    L_{m,n} = \max(0, |\hat{Q}_m - \hat{Q}_n| - e_{m,n}),
\end{equation}

\begin{equation}
    e_{m,n} =
    \begin{cases}
    Q_m - Q_n, & \text{if } \hat{Q}_m \geq \hat{Q}_n,\\
    Q_n - Q_m, & \text{if } \hat{Q}_m < \hat{Q}_n,
    \end{cases}
\end{equation}
where $m$ and $n$ represent different videos within the same training batch.

The overall training loss is then defined by combining these losses:
\begin{equation}
    L = L_{mae} + \beta \cdot L_{rk},
\end{equation}
where $\beta$ is a hyper-parameter used to balance the MAE and rank losses.

\section{Experiments}

\subsection{Data Analysis}

In order to measure the perceptual differences between original and corrected videos, we calculate 4 video attributes: MOS, brightness, contrast, and colorfulness, which are normalized and shown in Fig. \ref{Indexes} in the form of violin plots. The MOS of the over-exposed videos are significantly higher in general. This is quite reasonable since the over-exposed videos often have better brightness, resulting in more favorable visual experience than low-light ones. Colorfulness is not significantly changed before and after exposure correction. In comparison, the contrast and brightness change greatly, which is in line with visual perception. {Since there is a large amount of redundant information between adjacent frames, we only select a subset of all video frames for processing, \textit{i.e.}, choosing one frame from every four continuous frames.} The concrete calculation process is listed as follows:

\textit{(1) Brightness:} Given a video frame, we convert it to grayscale and compute the average of pixel values. Then the brightness result of a video is obtained by averaging the brightness of all selected frames.

\textit{(2) Contrast:} For a video frame, its contrast is obtained simply by computing standard deviation of pixel grayscale intensities. Then we average the contrast results of all selected frames to get the contrast of a video.

\textit{(3) Colorfulness:} We utilize Hasler and Suesstrunk’s metric~\cite{colorfulness} to calculate this attribute. Specifically, given a video frame in RGB format, we compute $rg = R - G$ and $yb = \frac{1}{2}(R + G) - B$ first, and the colorfulness is calculated by $\sqrt{\sigma_{rg}^{2} + \sigma_{yb}^{2}} + \frac{3}{10}\sqrt{\mu_{rg}^{2} + \mu_{yb}^{2}}$, where $\sigma^{2}$ and $\mu$ represent the variance and mean values respectively. Finally, we average the colorfulness values of all selected frames to obtain the colorfulness of a video.

\subsection{Performance Comparisons}

\begin{table*}[ht]
    \renewcommand{\arraystretch}{1.5}
    \caption{\textbf{Experimental performance on our constructed VEC dataset along with its subset.} Our proposed Light-VQA+
    achieves the best performance. ``HC'', ``DL'' and ``MLLM''  denote Hand-Crafted-based, Deep-Learning-based and Large-Language-Models-based features respectively. 
    The handcrafted models are inferior to deep-learning-based models, and deep-learning-based models are inferior to MLLM-involved models. [Key: \textcolor{red}{\textbf{Best}}; \textcolor{blue}{Second Best}]}
    \label{Performance}
    \centering
    \resizebox{\linewidth}{!}{
    \rowcolors{4}{white}{verylightgray}
    \begin{tabular}{|c|ccc|ccc|ccc|ccc|}
        \hline 
        \multirow{2}{*}{VQA Model}  &   \multirow{2}{*}{HC} & \multirow{2}{*}{DL} & \multirow{2}{*}{MLLM} & \multicolumn{3}{c|}{LLVE-QA} & \multicolumn{3}{c|}{OEVR-QA} & \multicolumn{3}{c|}{VEC-QA}\\ \cline{5-13}
        & & & & SRCC$\uparrow$ & PLCC$\uparrow$ & RMSE$\downarrow$ & SRCC$\uparrow$ & PLCC$\uparrow$ &RMSE$\downarrow$ & SRCC$\uparrow$ & PLCC$\uparrow$ &RMSE$\downarrow$\\ \hline
        V-BLIINDS~\cite{vqaVBLIINDS}  &  \ding{52}&	& &	0.6414 &0.6591 &18.5885 & 0.4553 & 0.4971 & 12.5661 & 0.7413 & 0.7655 & 15.5921 \\ 
        VIDEVAL~\cite{vqaVIDEVAL}  &  \ding{52}&	& &0.7624 &0.7658 &15.2591 & 0.4499& 0.4529&12.0544& 0.7865 & 0.8323 &13.6037 \\ 
        Simple-VQA~\cite{vqasimple}  &  &\ding{52}	& &	0.8955 &0.8983 &9.3784 & 0.5771 &0.5894 & 10.2310& 0.8608 & 0.9090 & 9.9034 \\ 
        FAST-VQA~\cite{vqafastvqa}&  &	\ding{52}& &	0.9130&0.9138 &9.0714 & 0.5440 &0.5608&10.5926 & 0.8574& 0.9126& 9.9307 \\ \hline
        MAX-VQA~\cite{vqamaxvqa} &&\ding{52}&\ding{52}&	0.9056 &0.9084 & 13.9633 & 0.3856& 0.4565&17.8459& 0.8219 & 0.9007 & 16.1984 \\
        Q-Align~\cite{vqaqalign}& & & \ding{52}&0.9105 &0.9107&	8.8343& 0.5257 & 0.5162& 11.3640 & 0.8500 & 0.9069&10.2925 \\
        Light-VQA~\cite{baselightvqa}  &\ding{52}&\ding{52}& &	\textcolor{blue}{0.9215} &	\textcolor{blue}{0.9239} &	\textcolor{blue}{8.1662}& \textcolor{blue}{0.5991} & \textcolor{blue}{0.6358} & \textcolor{blue}{9.7752} & \textcolor{blue}{0.8712} & \textcolor{blue}{0.9223} & \textcolor{blue}{9.1832} \\
         \hline
        \textbf{Light-VQA+} & & \ding{52}& \ding{52}&	\textbf{\textcolor{red}{0.9404}} &\textbf{\textcolor{red}{0.9393}} &	\textbf{\textcolor{red}{7.3710}} & \textbf{\textcolor{red}{0.7407}} & \textbf{\textcolor{red}{0.7661}} & \textbf{\textcolor{red}{8.3181}}&\textbf{\textcolor{red}{0.9121}}&\textbf{\textcolor{red}{0.9449}}&\textbf{\textcolor{red}{7.7744}} \\
        \hline
    \end{tabular}}
\end{table*}

To validate the effectiveness of Light-VQA+ on the constructed VEC-QA dataset, we choose Light-VQA~\cite{baselightvqa} as our baseline model, and compare their performance with 6 SOTA VQA models: V-BLIINDS~\cite{vqaVBLIINDS}, VIDEVAL~\cite{vqaVIDEVAL}, Simple-VQA~\cite{vqasimple}, FAST-VQA~\cite{vqafastvqa}, MaxVQA~\cite{vqamaxvqa}, and Q-Align~\cite{vqaqalign}, among which MaxVQA applies CLIP~\cite{clipmodel}, and Q-Align applies LLaVA~\cite{llavamodel1, llavamodel2, llavamodel3} as well as LoRA~\cite{modellora}. We utilize the same training strategy to train all models on the VEC-QA dataset and ensure their convergence. To be specific, our training process is to first train the models on the training data, then select the model that performs best on the validation data. At last, the selected model is tested on the test data. For Q-Align, we apply its pretrained one-align model and finetune it on our VEC-QA dataset via LoRA. The numbers of videos in training dataset, validation dataset, and test dataset are 3,162, 451, and 905 respectively. The test dataset is composed of 411 videos from LLVE-QA and 494 videos from OEVR-QA. The overall experimental results on VEC-QA test dataset and its  LLVE-QA and OEVR-QA subsets are shown in Tab.~\ref{Performance}.

\begin{figure*}[ht]
\centering
\begin{subfigure}[]{0.24\textwidth}
    \begin{minipage}[]{1\textwidth}
        \centering
        \includegraphics[width=1\textwidth]{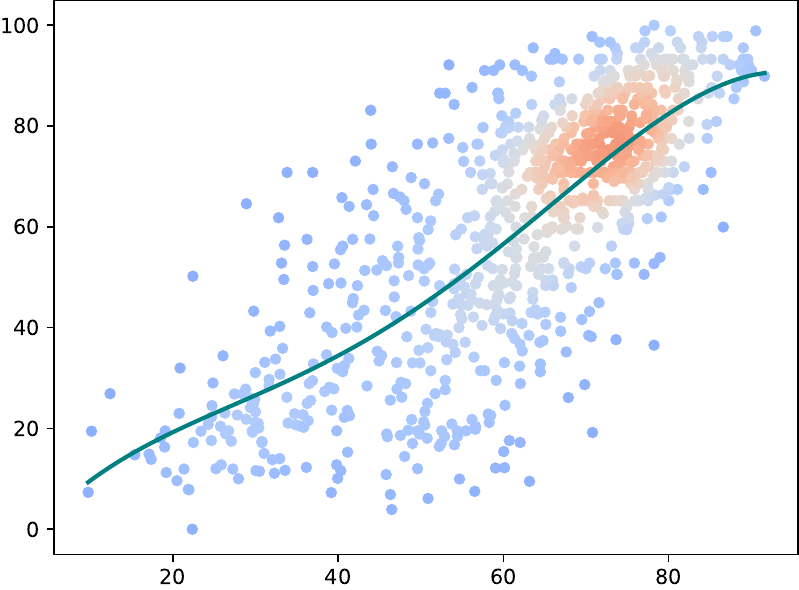}
    \end{minipage}
    \vspace{2pt}
    \begin{minipage}[]{1\textwidth}
        \hspace{2pt}
        \includegraphics[width= 0.47\textwidth]{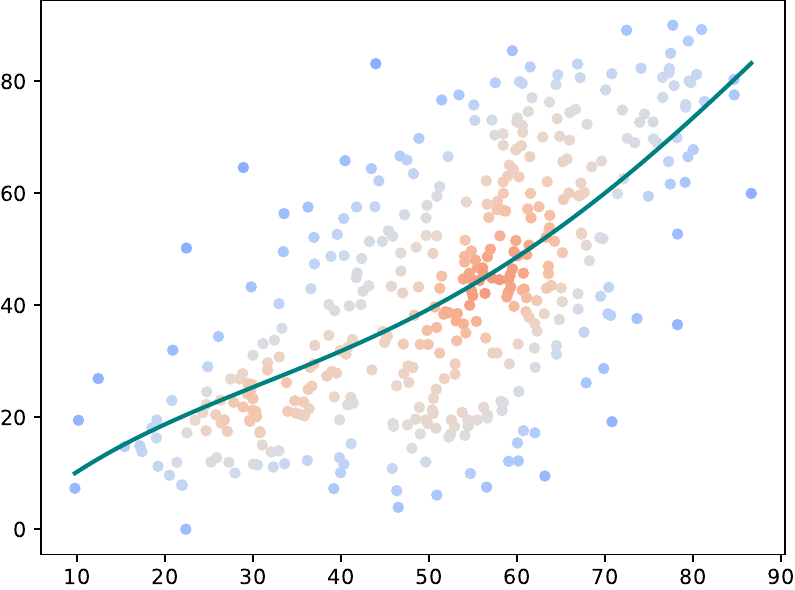}
        \includegraphics[width= 0.47\textwidth]{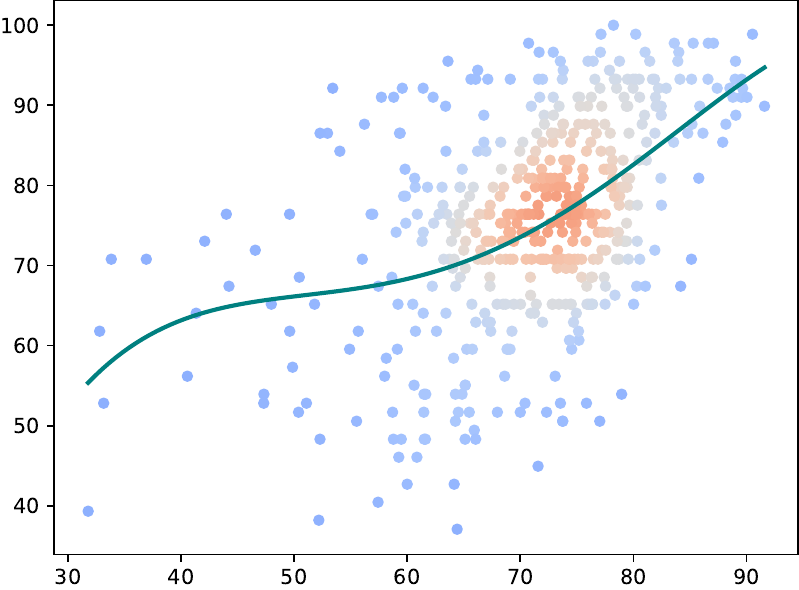} 
    \end{minipage}
    \subcaption{V-BLIINDS~\cite{vqaVBLIINDS}}
\end{subfigure}
\begin{subfigure}[]{0.24\textwidth}
    \begin{minipage}[]{1\textwidth}
        \centering
        \includegraphics[width=1\textwidth]{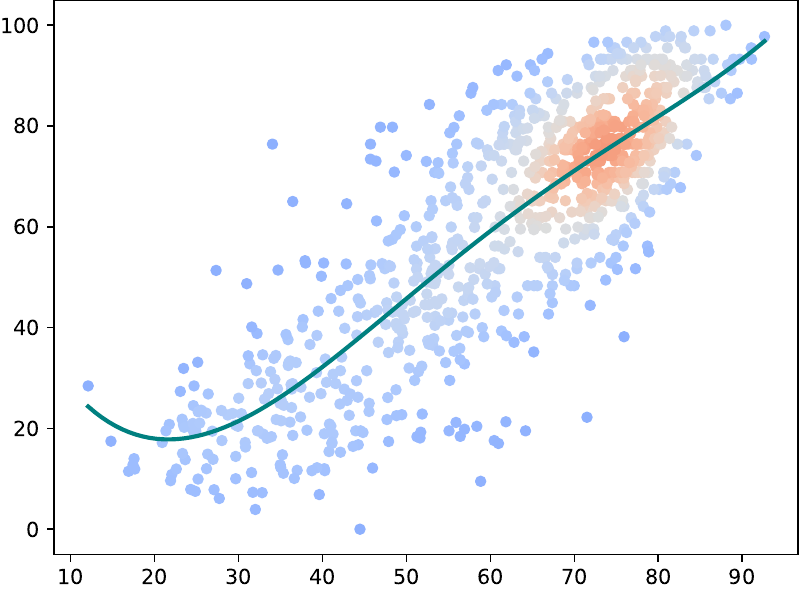}
    \end{minipage}
    \vspace{2pt}
    \begin{minipage}[]{1\textwidth}
        \hspace{2pt}
        \includegraphics[width= 0.47\textwidth]{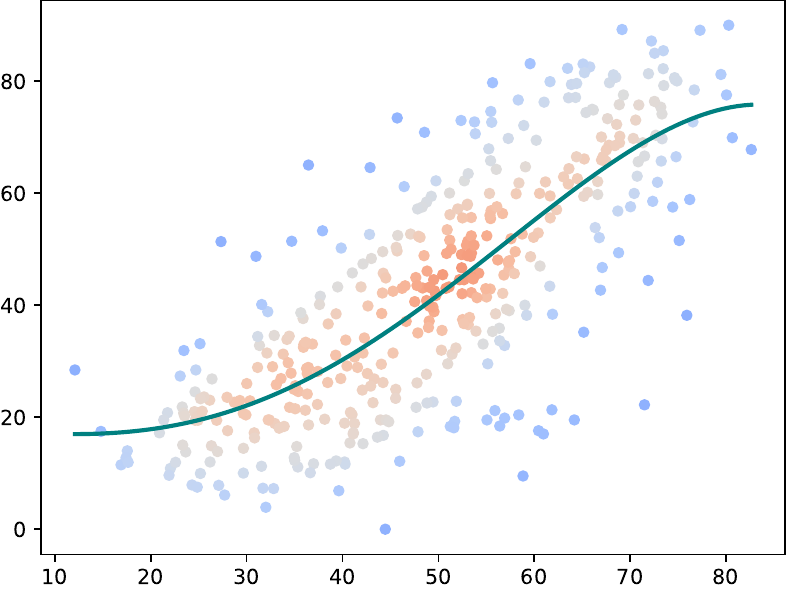}
        \includegraphics[width= 0.47\textwidth]{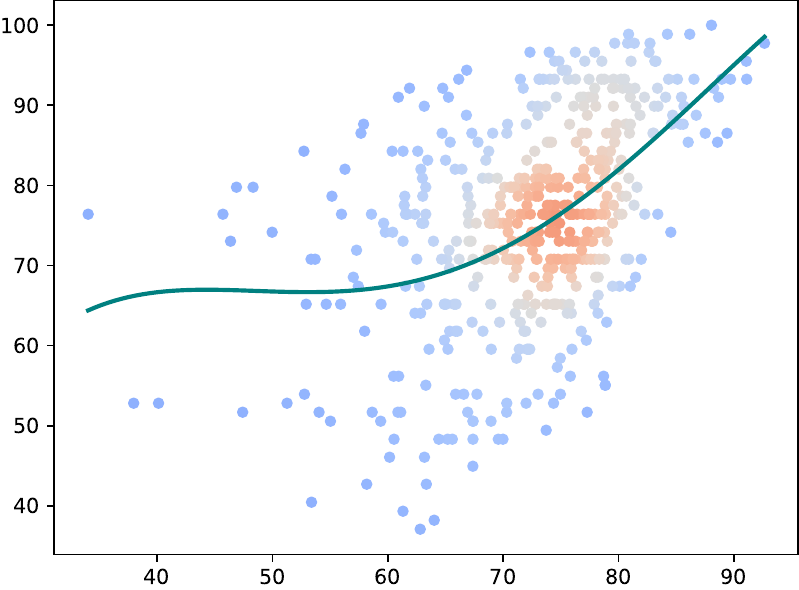} 
    \end{minipage}
    \subcaption{VIDEVAL~\cite{vqaVIDEVAL}}
\end{subfigure}
\begin{subfigure}[]{0.24\textwidth}
    \begin{minipage}[]{1\textwidth}
        \centering
        \includegraphics[width=1\textwidth]{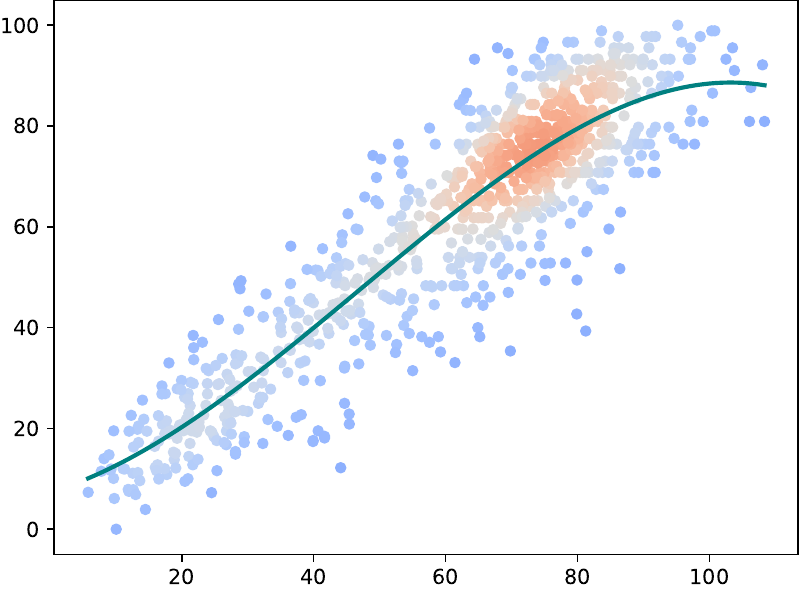}
    \end{minipage}
    \vspace{2pt}
    \begin{minipage}[]{1\textwidth}
        \hspace{2pt}
        \includegraphics[width= 0.47\textwidth]{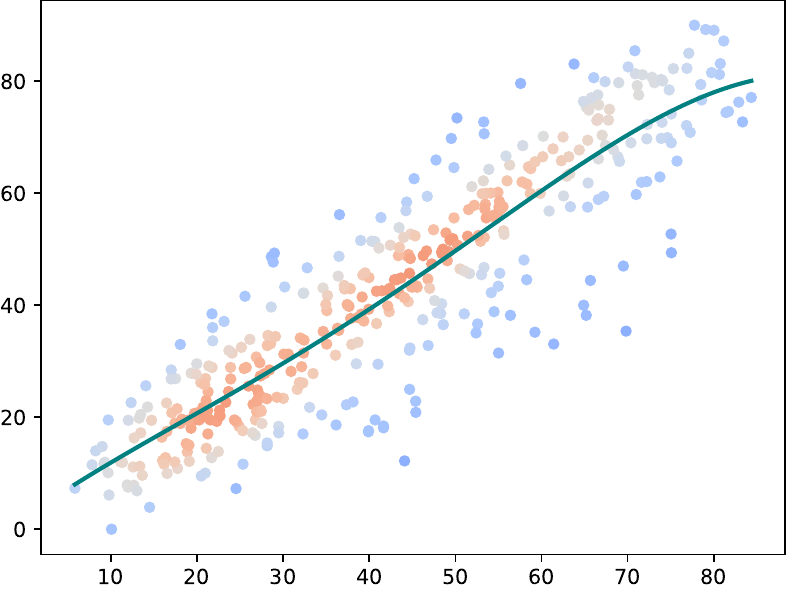}
        \includegraphics[width= 0.47\textwidth]{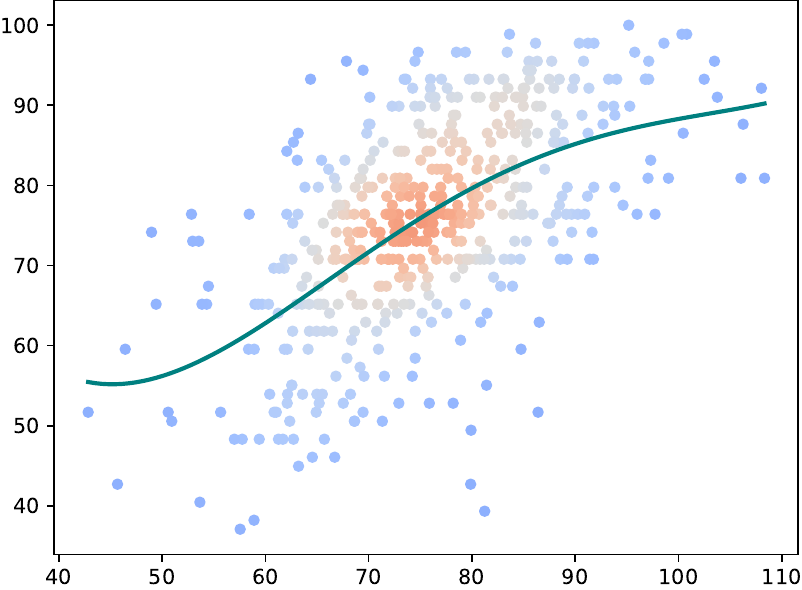} 
    \end{minipage}
    \subcaption{Simple-VQA~\cite{vqasimple}}
\end{subfigure}
\begin{subfigure}[]{0.24\textwidth}
    \begin{minipage}[]{1\textwidth}
        \centering
        \includegraphics[width=1\textwidth]{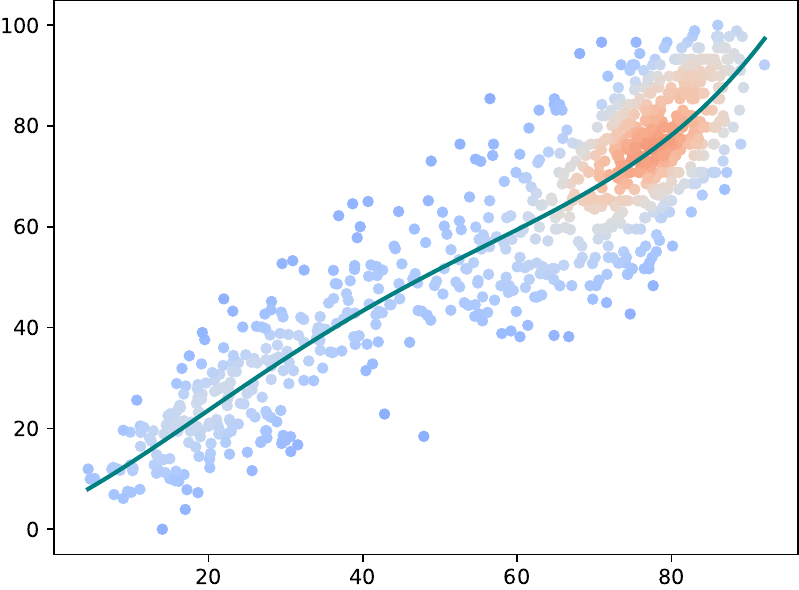}
    \end{minipage}
    \vspace{2pt}
    \begin{minipage}[]{1\textwidth}
        \hspace{2pt}
        \includegraphics[width= 0.47\textwidth]{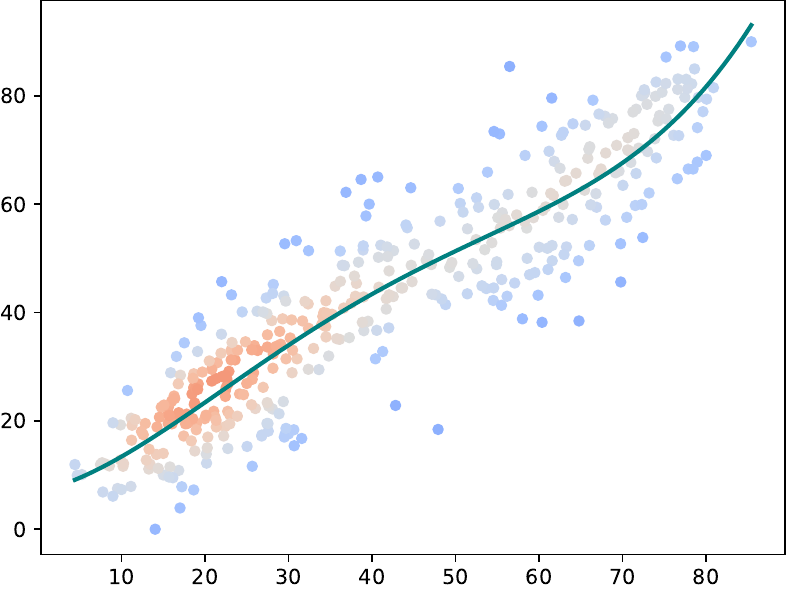}
        \includegraphics[width= 0.47\textwidth]{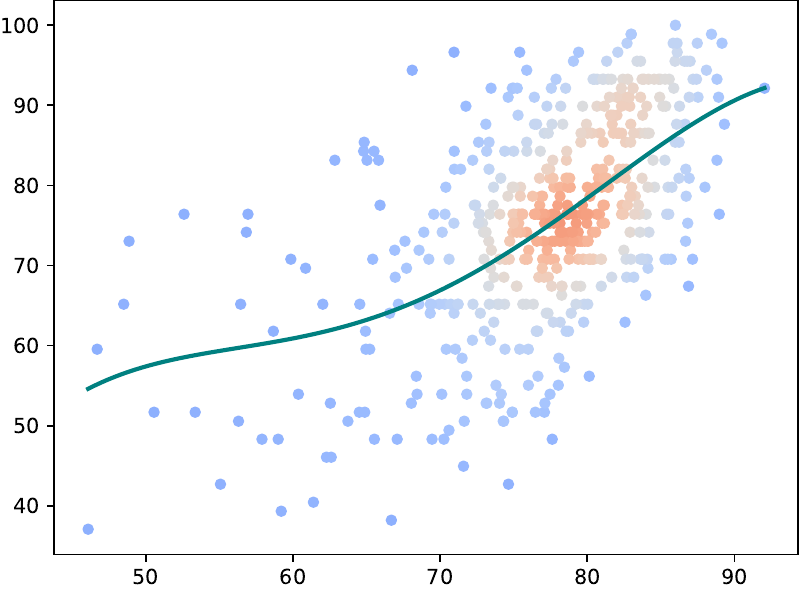} 
    \end{minipage}
    \subcaption{Fast-VQA~\cite{vqafastvqa}}
\end{subfigure}
\vspace{5pt}
\begin{subfigure}[]{0.24\textwidth}
    \begin{minipage}[]{1\textwidth}
        \centering
        \includegraphics[width=1\textwidth]{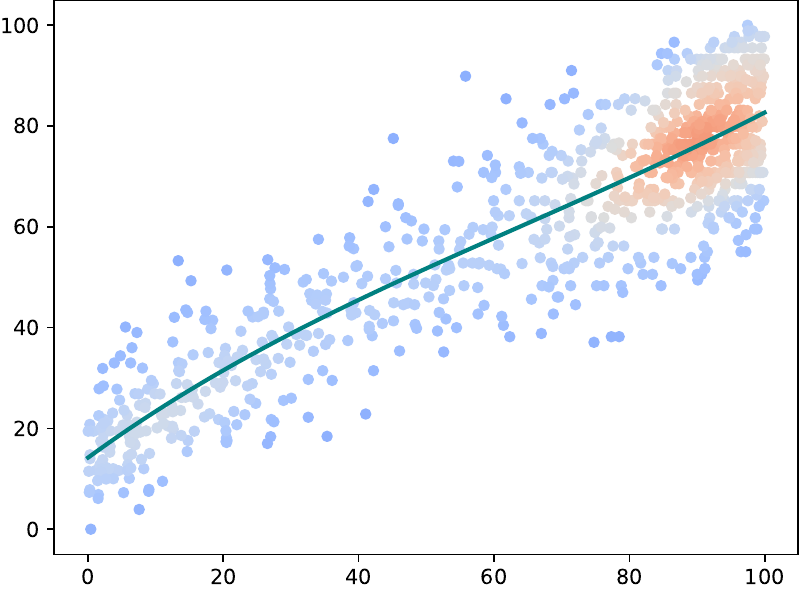}
    \end{minipage}
    \vspace{2pt}
    \begin{minipage}[]{1\textwidth}
        \hspace{2pt}
        \includegraphics[width= 0.47\textwidth]{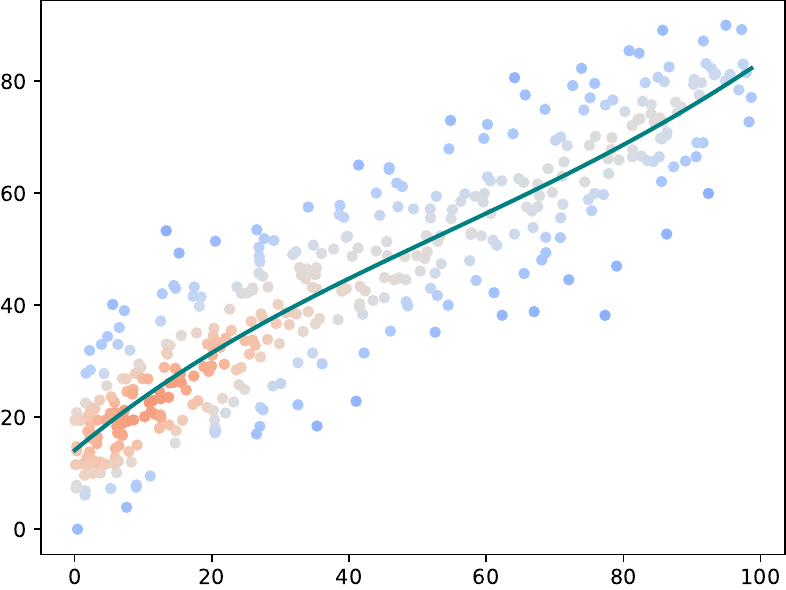}
        \includegraphics[width= 0.47\textwidth]{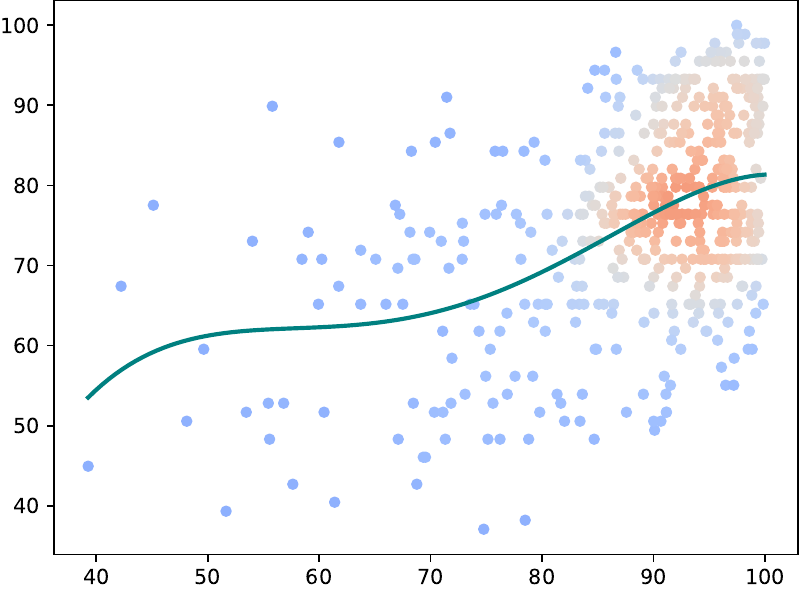} 
    \end{minipage}
    \subcaption{Max-VQA~\cite{vqamaxvqa}}
\end{subfigure}
\begin{subfigure}[]{0.24\textwidth}
    \begin{minipage}[]{1\textwidth}
        \centering
        \includegraphics[width=1\textwidth]{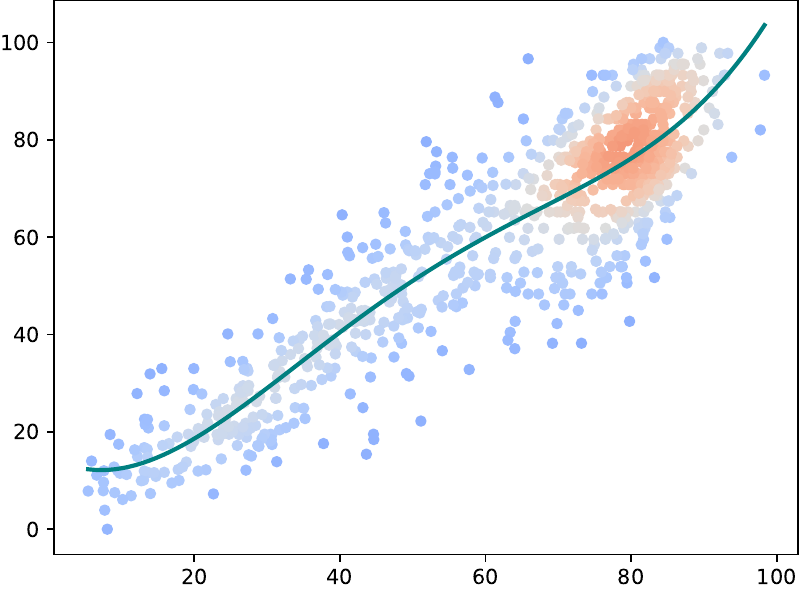}
    \end{minipage}
    \vspace{2pt}
    \begin{minipage}[]{1\textwidth}
        \hspace{2pt}
        \includegraphics[width= 0.47\textwidth]{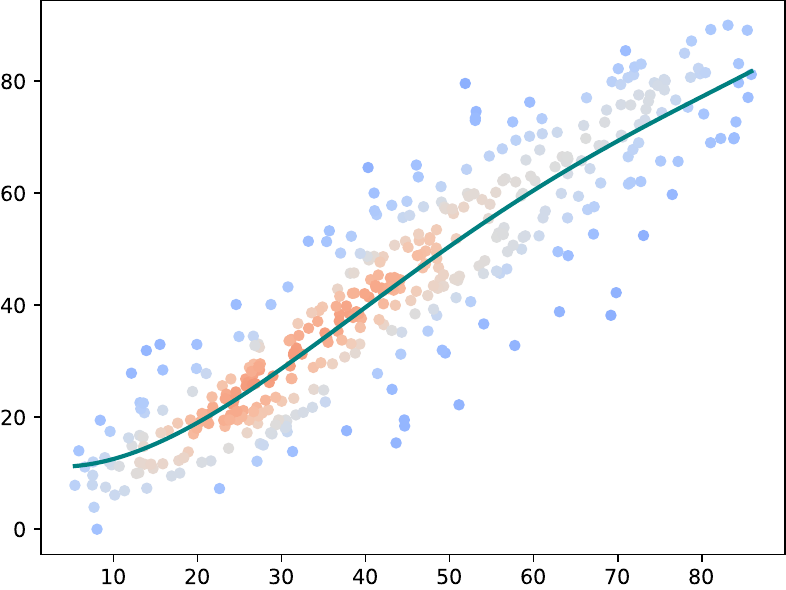}
        \includegraphics[width= 0.47\textwidth]{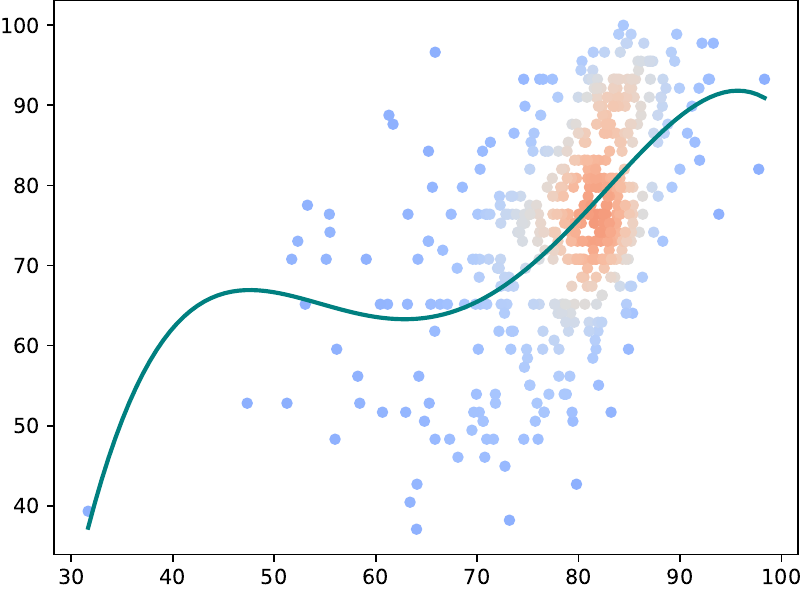} 
    \end{minipage}
    \subcaption{Q-Align~\cite{vqaqalign}}
\end{subfigure}
\begin{subfigure}[]{0.24\textwidth}
    \begin{minipage}[]{1\textwidth}
        \centering
        \includegraphics[width=1\textwidth]{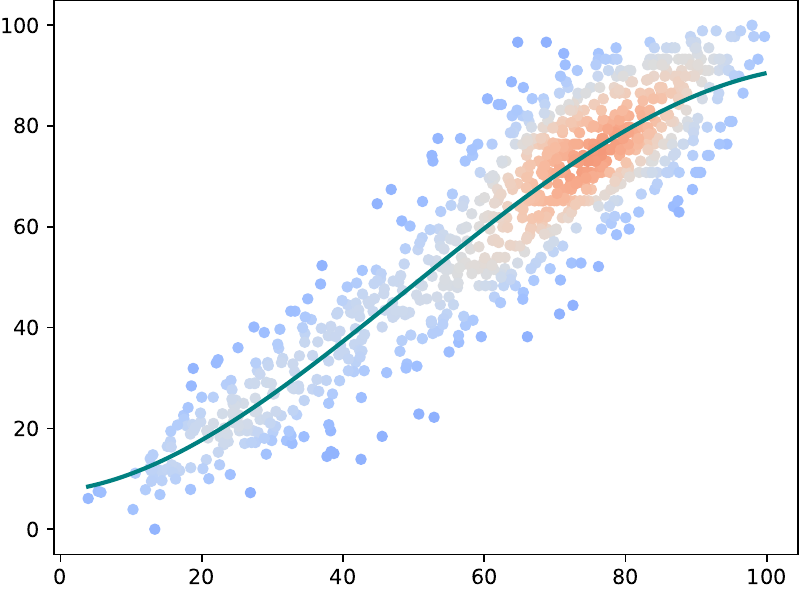}
    \end{minipage}
    \vspace{2pt}
    \begin{minipage}[]{1\textwidth}
        \hspace{2pt}
        \includegraphics[width= 0.47\textwidth]{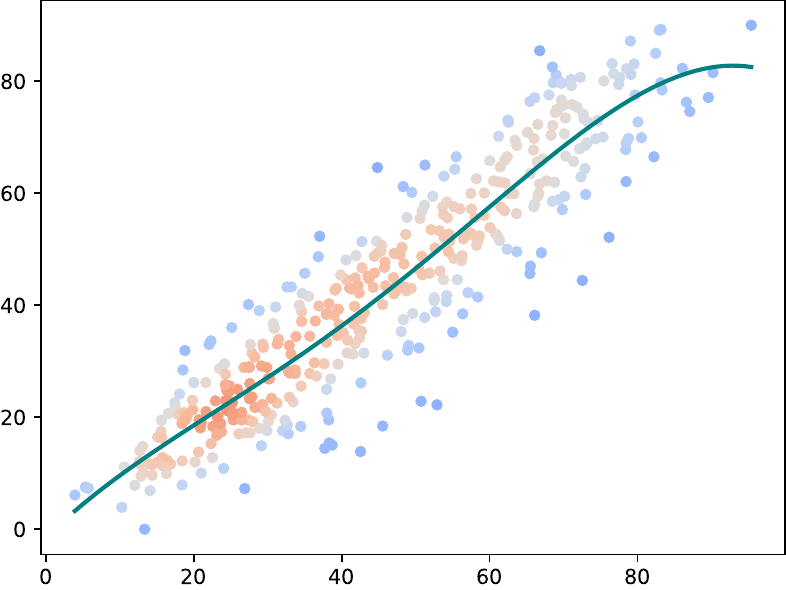}
        \includegraphics[width= 0.47\textwidth]{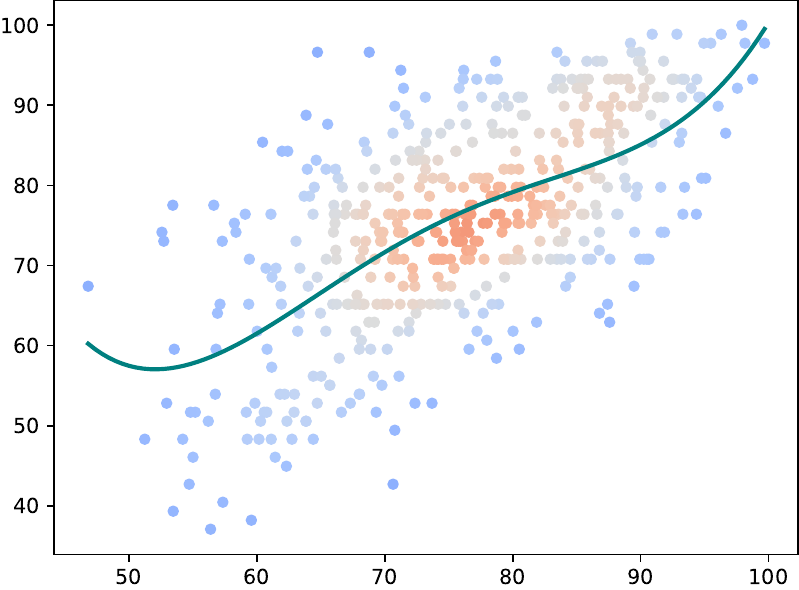} 
    \end{minipage}
    \subcaption{Light-VQA~\cite{baselightvqa}}
\end{subfigure}
\begin{subfigure}[]{0.24\textwidth}
    \begin{minipage}[]{1\textwidth}
        \centering
        \includegraphics[width=1\textwidth]{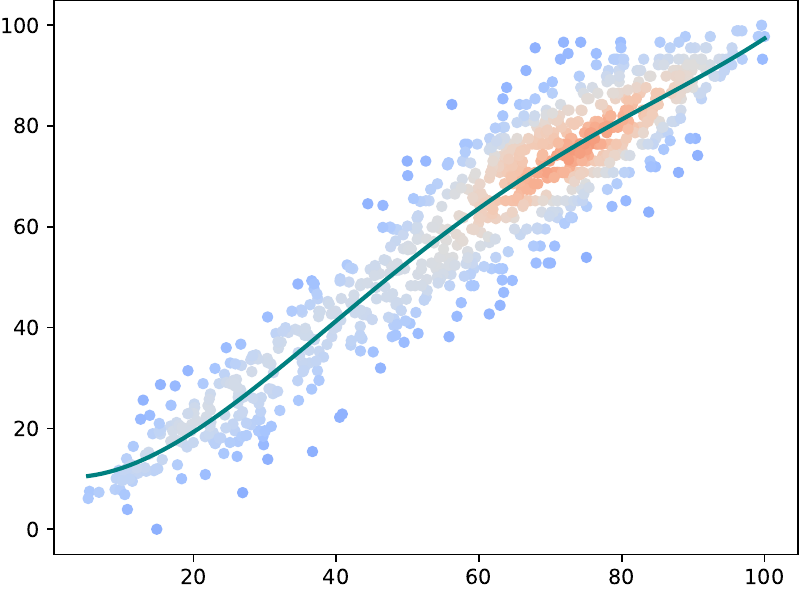}
    \end{minipage}
    \vspace{2pt}
    \begin{minipage}[]{1\textwidth}
        \hspace{2pt}
        \includegraphics[width= 0.47\textwidth]{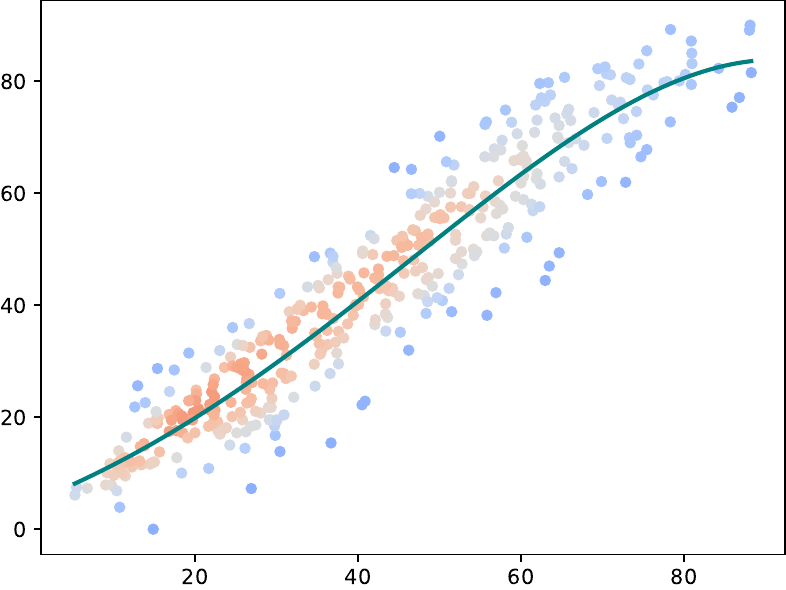}
        \includegraphics[width= 0.47\textwidth]{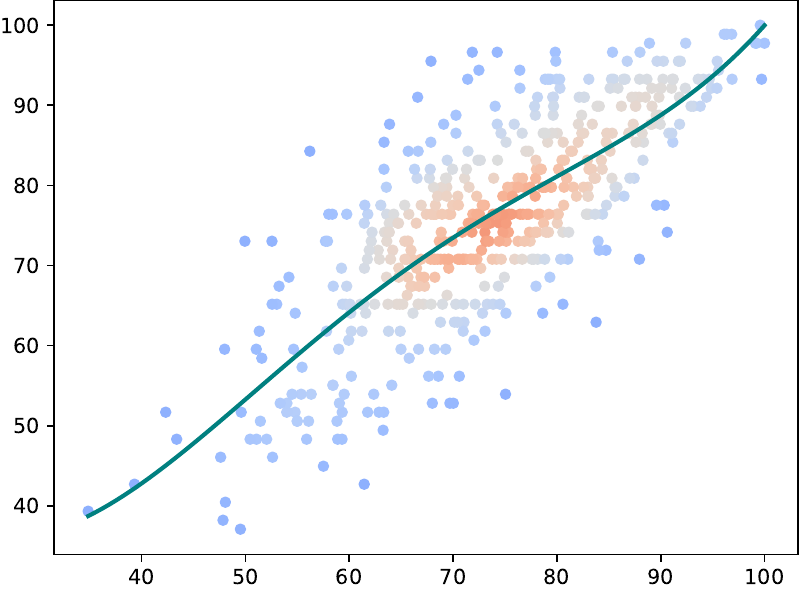} 
    \end{minipage}
    \subcaption{\textbf{Light-VQA+}}
\end{subfigure}
\caption{The scatter plots of the predicted scores v.s. the MOS. The curves are obtained by a four-order polynomial nonlinear fitting. For a model, there are 3 scatter plots, standing for the testing results on VEC-QA, LLVE-QA, and OEVR-QA respectively. It is evident that the predicted scores of our proposed VQA bear the closest resemblance to the MOS.}
\label{Scatter}
\end{figure*}

Fig. \ref{Scatter} shows the scatter plots of the predicted MOS v.s. the ground-truth MOS on VEC-QA dataset for $7$ VQA models. The shown curves are obtained by a four-order polynomial nonlinear fitting. According to Tab. \ref{Performance}, Light-VQA+ achieves the best performance in all $7$ models and leads the second best method (\textit{i.e.}, Light-VQA) by a large margin, which demonstrates its effectiveness for the perceptual quality assessment of video exposure correction.

\subsection{{Cross-Dataset Validation}}

\begin{table*}[t]
    \renewcommand{\arraystretch}{1.5}
    \caption{\textbf{Cross-dataset validation on different VQA datasets.} Our proposed Light-VQA+ achieves the best performance. ``HC'', ``DL'' and ``MLLM''  denote Hand-Crafted-based, Deep-Learning-based and Large-Language-Models-based features respectively. The handcrafted models are inferior to deep-learning-based models, and deep-learning-based models are inferior to MLLM-involved models. [Key: \textcolor{red}{\textbf{Best}}; \textcolor{blue}{Second Best}]}
    \label{Crossdataset}
    \centering
    \resizebox{\linewidth}{!}{
    \rowcolors{4}{white}{verylightgray}
    \begin{tabular}{|c|ccc|ccc|ccc|ccc|}
        \hline 
        \multirow{2}{*}{VQA Model}  &   \multirow{2}{*}{HC} & \multirow{2}{*}{DL} & \multirow{2}{*}{MLLM} & \multicolumn{3}{c|}{Mixed-A} & \multicolumn{3}{c|}{Mixed-B} & \multicolumn{3}{c|}{Mixed-C}\\ \cline{5-13}
        & & & & SRCC$\uparrow$ & PLCC$\uparrow$ & RMSE$\downarrow$ & SRCC$\uparrow$ & PLCC$\uparrow$ &RMSE$\downarrow$ & SRCC$\uparrow$ & PLCC$\uparrow$ &RMSE$\downarrow$\\ \hline
        V-BLIINDS~\cite{vqaVBLIINDS}  &  \ding{52}&	& &0.5493&0.5617&15.9107&0.6534&0.6695&15.1749&0.5922&0.5916&19.4193\\  
        VIDEVAL~\cite{vqaVIDEVAL}  &  \ding{52}&	& &0.6556&0.6539&14.3835&0.6667&0.6817&15.0229&0.6665&0.6671&16.8866\\
        Simple-VQA~\cite{vqasimple}  &  &\ding{52}	& &0.7116&0.7094&12.5252&0.6845&\textcolor{blue}{0.7176}&13.8639&0.7477&0.7678&12.2789\\
        FAST-VQA~\cite{vqafastvqa}&  &	\ding{52}& &\textcolor{blue}{0.7396}&0.6859&14.0843&\textcolor{blue}{0.6919}&{0.6988}&15.4489&0.7675&0.7545&13.4308\\ \hline
        MAX-VQA~\cite{vqamaxvqa} &&\ding{52}&\ding{52}&0.6824&0.6525&22.0832&0.5466&0.5895&16.8935&0.6075&0.6161&16.2767\\
        Q-Align~\cite{vqaqalign}& & & \ding{52}&0.7391&0.6797&15.5960&0.6516&0.6639&17.1240&\textcolor{blue}{0.8000}&0.7882&14.9778\\
        Light-VQA~\cite{baselightvqa}  &\ding{52}&\ding{52}& &	{0.7284} &	\textcolor{blue}{0.7286} &	\textcolor{blue}{12.1709}& {0.6358} & {0.6836} & \textcolor{blue}{14.5292} & {0.7872} & \textcolor{blue}{0.8091} & \textcolor{blue}{11.2627} \\
         \hline
        \textbf{Light-VQA+} & & \ding{52}& \ding{52}&	\textbf{\textcolor{red}{0.7435}} &\textbf{\textcolor{red}{0.7419}} &	\textbf{\textcolor{red}{11.9154}} & \textbf{\textcolor{red}{0.7328}} & \textbf{\textcolor{red}{0.7601}} & \textbf{\textcolor{red}{12.9351}}&\textbf{\textcolor{red}{0.8016}}&\textbf{\textcolor{red}{0.8181}}&\textbf{\textcolor{red}{11.0225}} \\
        \hline
    \end{tabular}}
\end{table*}

To evaluate the cross-dataset performance of the model, we conduct experiments using mixed subsets of videos with improper exposure from three sources: VDPVE~\cite{datavdpve}, KoNViD\_1k~\cite{datakonvid}, and Live-VQC~\cite{dataLiveVQC}. {After selecting videos from these datasets, we combine them and divided them into three subsets, designated as Mixed-A, Mixed-B, and Mixed-C from over-exposed, low-light, and merged perspectives, each containing 600 videos.} We then directly apply models pre-trained on the VEC-QA dataset to test these new subsets, facilitating an efficient assessment process. The overall experimental results are presented in Tab. \ref{Crossdataset}.

It is important to note that the VEC-QA dataset is specifically designed to include videos with improper exposure as well as their corrected counterparts, making it suitable for training exposure-related quality assessment models. However, the test subsets are not originally intended for this task. This mismatch in dataset design has led to less effective performance of the quality-aware representations learned from the VEC-QA dataset, resulting in a performance decline across all tested methods. Despite this general downturn, our proposed Light-VQA+ method still manages to outperform the other seven VQA methods, showcasing its robust generalization ability in assessing the quality of exposure-corrected videos.

\subsection{Ablation Studies}
\begin{table*}[t]
\renewcommand{\arraystretch}{1.5}
\centering
\caption{\textbf{Experimental performance of ablation studies on our constructed VEC dataset along with its subset.} [Key: \textcolor{red}{\textbf{Best}}; \textcolor{blue}{Second Best}; HVS-W: HVS weights]}
\label{ablation}
\resizebox{\linewidth}{!}{
\rowcolors{4}{white}{verylightgray}
\begin{tabular}{|c|ccc|ccc|ccc|ccc|}
\hline
\multirow{2}{*}{Model}  & \multirow{2}{*}{BN \& BC} & \multirow{2}{*}{Fusion} & \multirow{2}{*}{HVS-W} &  \multicolumn{3}{c|}{LLVE-QA}&  \multicolumn{3}{c|}{OEVR-QA} &  \multicolumn{3}{c|}{VEC-QA} \\ \cline{5-13}
& & & &SRCC$\uparrow$ & PLCC$\uparrow$ & RMSE$\downarrow$ &SRCC$\uparrow$ & PLCC$\uparrow$ & RMSE$\downarrow$ &SRCC$\uparrow$ & PLCC$\uparrow$ & RMSE$\downarrow$ \\ \hline
1  & None & MLP &\ding{56} &0.9125&0.9159&8.5649&0.5591 & 0.5947& 10.1813&0.8614&0.9138&9.6487\\
2 & Handcraft & MLP & \ding{56}&0.9215&0.9239&8.1662&0.5991&0.6358&9.7752&0.8712&0.9223&9.1832\\
3 & MLLM (Local) & MLP & \ding{56}&0.9231&0.9261&8.0513&0.7118&0.7260&8.7087&0.9010&0.9343&8.4690\\
4 & MLLM (Global) & MLP & \ding{56}&0.9231&0.9260&8.0588&0.7098&0.7316&8.6338&0.9021&0.7378&0.9350\\
5 & MLLM (Both) & MLP &\ding{52}&\textcolor{blue}{0.9317}&\textcolor{blue}{0.9324}&\textcolor{blue}{7.7131}&0.7061&0.7264&8.7039&0.9010&0.9367&8.3187\\
6 & MLLM (Both) & Cross-Attn & \ding{56} &0.9290&0.9310&7.7910&\textcolor{blue}{0.7215}&\textcolor{blue}{0.7457}&\textcolor{blue}{8.4386}&\textcolor{blue}{0.9056}&\textcolor{blue}{0.9384}&\textcolor{blue}{8.2094}\\
\textbf{Light-VQA+} & MLLM (Both) & Cross-Attn &\ding{52} &\textbf{\textcolor{red}{ 0.9428}} &\textbf{\textcolor{red}{0.9450}} &	\textbf{\textcolor{red}{6.9803}} & \textbf{\textcolor{red}{0.7462}} & \textbf{\textcolor{red}{0.7772}} & \textbf{\textcolor{red}{7.9690}}&\textbf{\textcolor{red}{0.9177}}&\textbf{\textcolor{red}{0.9480}}&\textbf{\textcolor{red}{7.5597}} \\
\hline      
\end{tabular}
}
\end{table*}

In this subsection, we conduct a series of ablation studies to evaluate the individual contributions of different modules within Light-VQA+. The results of these ablation studies are presented in Tab. \ref{ablation}. \textit{Model 1} removes the BNF and BCF extracted via CLIP, the cross-attention module~\cite{crossattention}, and {the HVS weights}. Based on \textit{Model 1}, \textit{Model 2} utilizes the handcrafted methods to extract the BNF and BCF. \textit{Model 3} utilizes the local features extracted by CLIP, which are the local BNF in SI and the Lv4 BCF in TI. Opposing to \textit{Model 3}, \textit{Model 4} exploits the global features (\textit{i.e.}, the global BNF and the BCF from Lv1 to Lv3). \textit{Model 5} removes the cross-attention module and applies a MLP~\cite{modelmlp} instead, while \textit{Model 6} uses a simple average to replace the parameters that align with the HVS. The last one, Light-VQA+, is the complete model we propose, in which we fuse all the spatial and temporal information, and obtain the best results. Based on the results of the 7 models, we can analyze the contribution of each modules in Light-VQA+.
\subsubsection{Feature Extraction Module}

For Light-VQA+, a significant advancement lies in the method that extracts the BNF and BCF. Notably, multimodal large language models demonstrate the superior performance compared to traditional methods in assessing video quality. \textit{Model 1} and \textit{Model 2} are specifically designed to evaluate the efficacy of CLIP in capturing features. To ensure that the test results remain unaffected by the feature-fusion and regression modules, we also disabled the cross-attention module and weight parameters in both \textit{Model 1} and \textit{Model 2}. The results clearly indicate that CLIP significantly outperforms the traditional methods in extracting such features from frames. {Furthermore, the results of \textit{Model 3} and \textit{Model 4} validate the effectiveness of integrating both local and global perspectives.}

\subsubsection{{Feature Fusion \& Regression Modules}}

Extracting features is as crucial as utilizing them effectively. To this end, verifying the efficacy of the cross-attention module is essential. 

To this end, \textit{Model 5} is crafted to test the effectiveness of the cross-attention module by excluding it from the final model. Additionally, the importance of HVS weights should also be verified. Thus, \textit{Model 6} excludes these {HVS weights} during the training process. By comparing the outcomes from \textit{Model 5} and \textit{Model 6} with our full model (\textit{i.e.}, {Light-VQA+}), it becomes apparent that both the cross-attention and HVS weights play significant roles in enhancing the predictive performance of Light-VQA+.

\subsection{{Benefits to Exposure Correction Algorithms}}
\begin{table}[t]
\setlength\tabcolsep{1pt}
\renewcommand{\arraystretch}{1.3}
\centering
\caption{\textbf{The detailed qualitative comparisons of experiments.} The videos we test are the original low-light and over-exposed videos from VEC-QA corrected by the pre-trained and fine-tuned FEC-Net models. {The scores presented are the average of the corrected 100 videos assessed by Light-VQA+.} The figures below are the example frames from the corrected videos. [Key: OE: Over Exposed; LL: Low Light; Score: OE/LL; \textcolor{red}{\textbf{Best}}; \textcolor{blue}{Second Best}]}
\label{refinement}
\begin{tabular}{c|ccc}
\hline
{Video}&{Original}&{Pre-trained}&{Fine-tuned}\\ 
\hline
Score&68.1/46.2&\textcolor{blue}{69.6/59.8}&\textbf{\textcolor{red}{74.2/64.0}}\\
\hline
\multirow{3}{*}{OE}&
\multirow{3}{*}{
        \begin{minipage}[b]{0.27\linewidth}
        \centering
        \vspace*{0.8mm} 
		\includegraphics[width=1\linewidth]{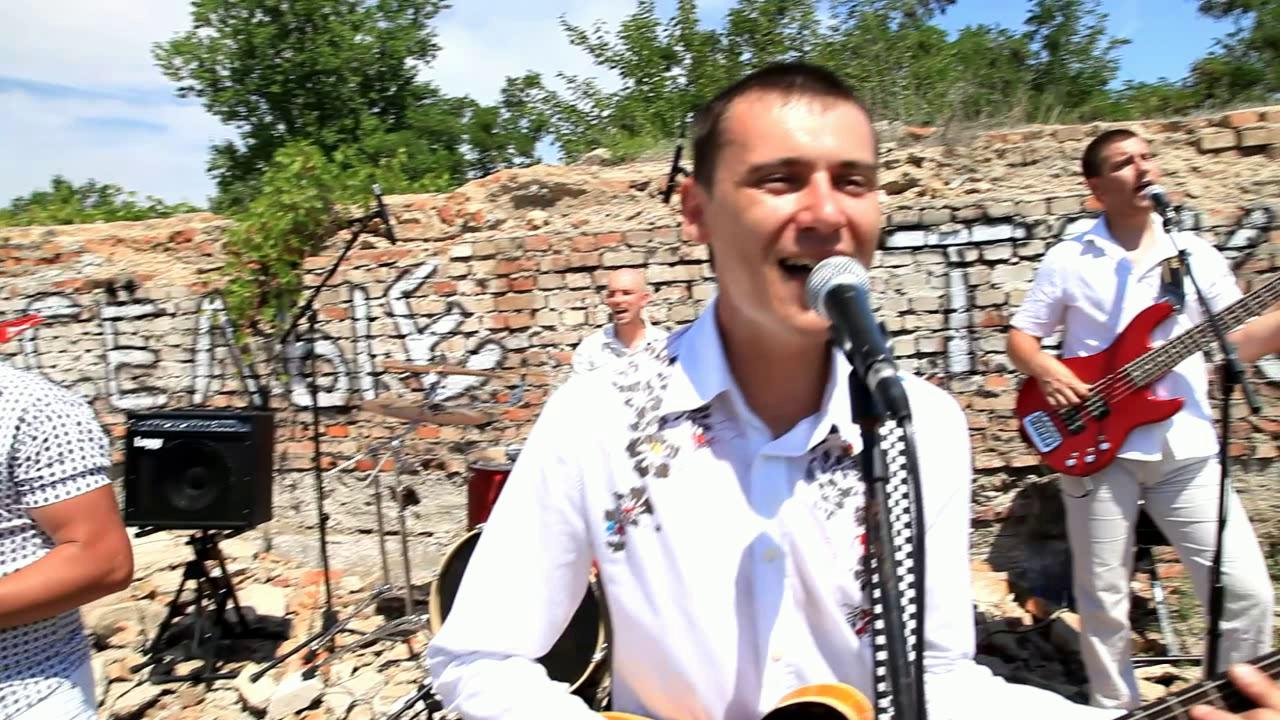}
        \end{minipage}
        }&
\multirow{3}{*}{
        \begin{minipage}[b]{0.27\linewidth}
        \centering
        \vspace*{0.8mm} 
		\includegraphics[width=1\linewidth]{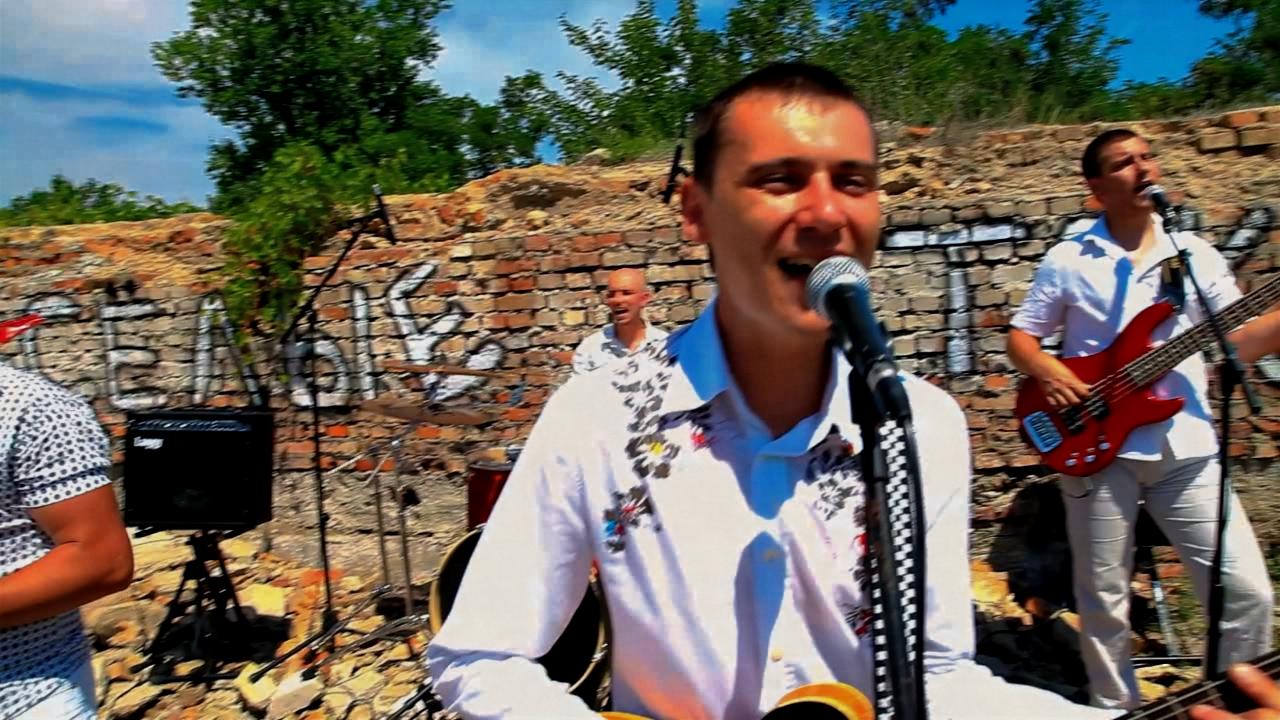}
        \end{minipage}
        }&
\multirow{3}{*}{
        \begin{minipage}[b]{0.27\linewidth}
        \centering
        \vspace*{0.8mm} 
		\includegraphics[width=1\linewidth]{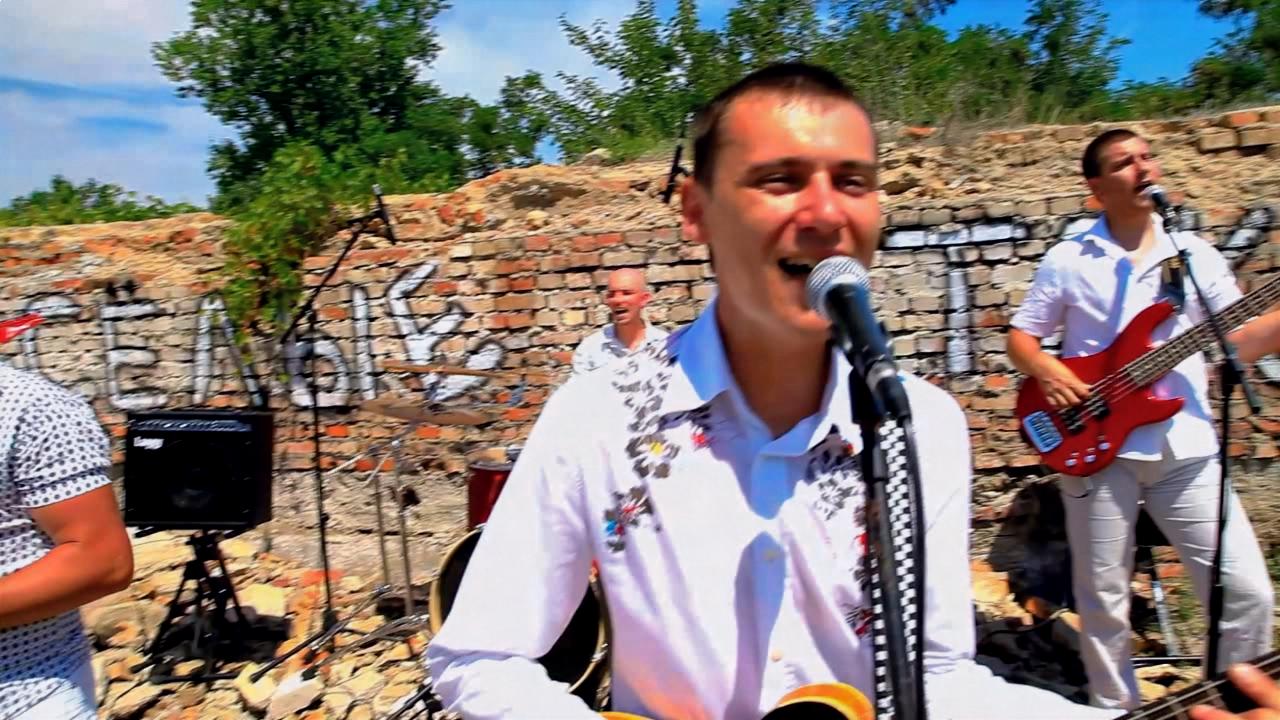}
        \end{minipage}
        }
\\
\\
\\
\hline
\multirow{3}{*}{LL}&
\multirow{3}{*}{
        \begin{minipage}[b]{0.27\linewidth}
        \centering
        \vspace*{0.8mm} 
		\includegraphics[width=1\linewidth]{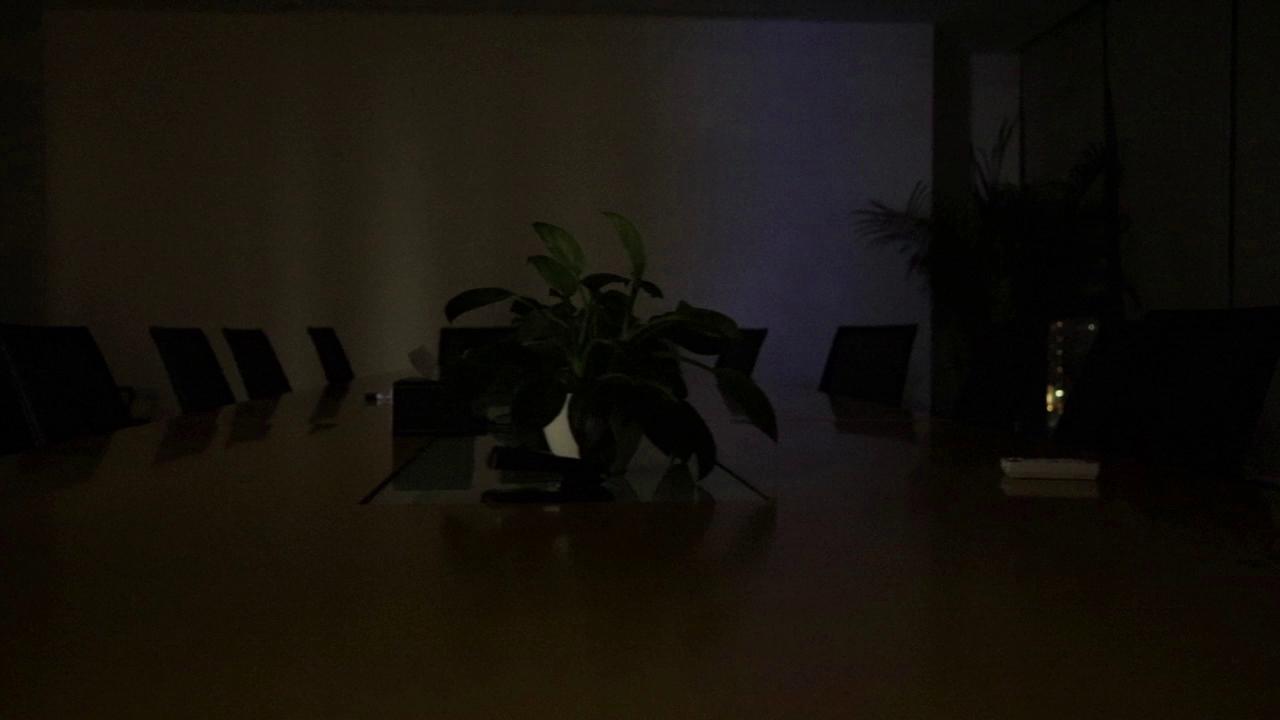}
        \end{minipage}
        }&
\multirow{3}{*}{
        \begin{minipage}[b]{0.27\linewidth}
        \centering
        \vspace*{0.8mm} 
		\includegraphics[width=1\linewidth]{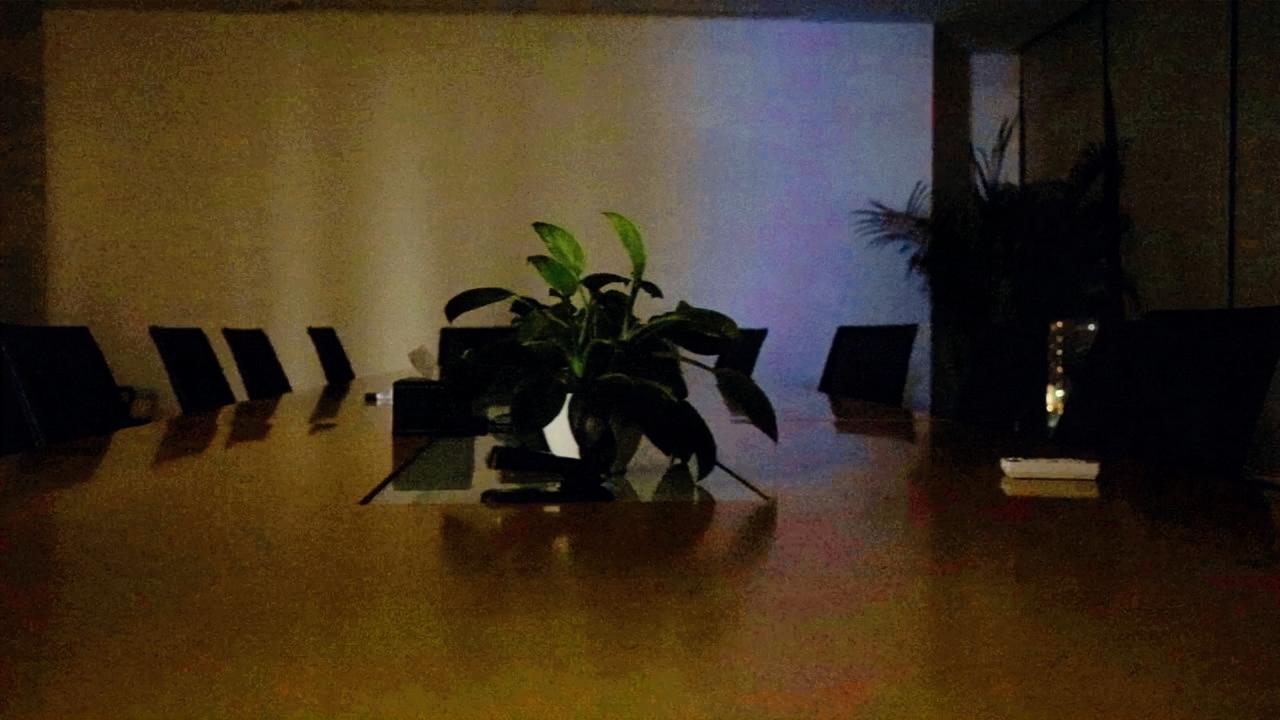}
        \end{minipage}
        }&
\multirow{3}{*}{
        \begin{minipage}[b]{0.27\linewidth}
        \centering
        \vspace*{0.8mm} 
		\includegraphics[width=1\linewidth]{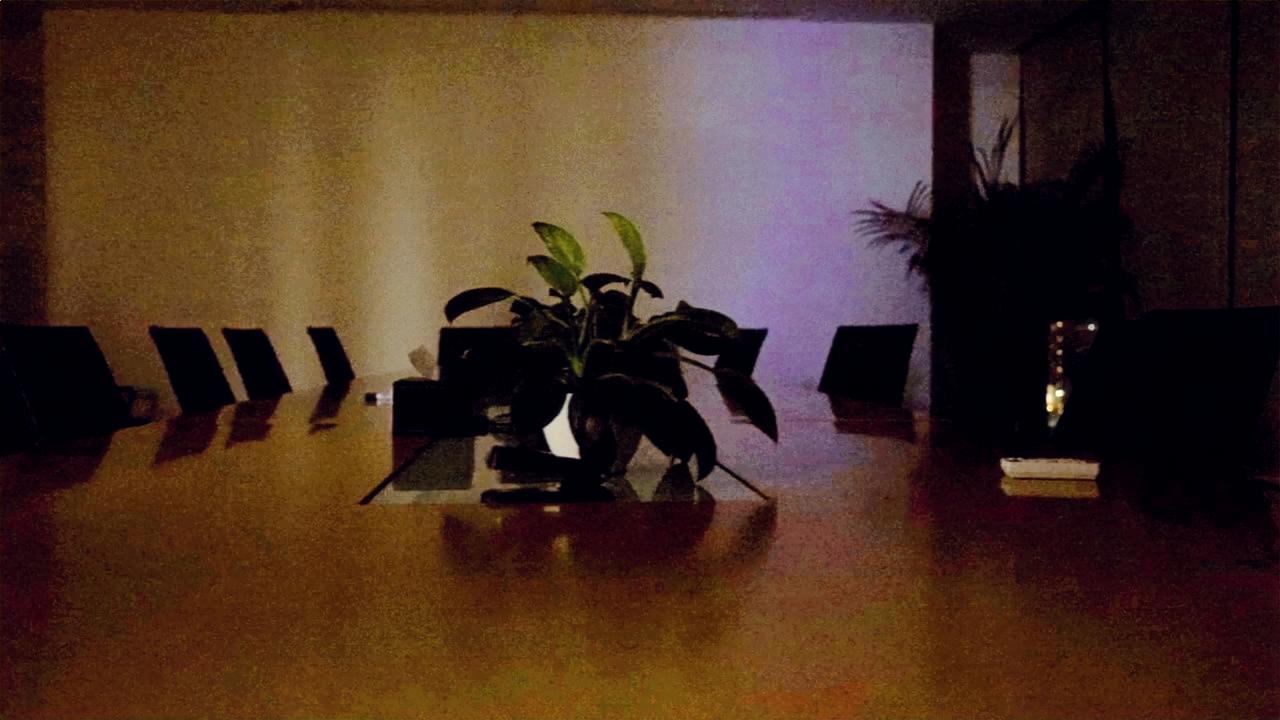}
        \end{minipage}
        }
\\
\\
\\
\hline
\end{tabular}
\end{table}

To demonstrate that our Light-VQA+ can serve as an effective metric for facilitating the development of exposure correction algorithms, we utilize the Light-VQA+ to fine-tune a recent exposure correction algorithm, FEC-Net~\cite{algoFECNET}. We leverage the released FEC-Net model that is pre-trained on SICE dataset, which consists of two subsets that collect the low-light and over-exposed images respectively.

Subsequently, the Light-VQA+ is utilized to fine-tune the FEC-Net. During the fine-tuning phase, the result of Light-VQA+ is incorporated as part of the loss function. {After training and fine-tuning, we randomly choose 50 over-exposed and 50 low-light original videos from VEC-QA, and use the pre-trained and fine-tuned FEC-Net models to correct them. Finally, we assess the quality of corrected videos.} 

As shown in Tab. \ref{refinement}, the pre-trained FEC-Net model is capable of correcting the improper exposed videos. With the help of Light-VQA+, the performance of fine-tuned FEC-Net is further improved in exposure correction, which justifies the effectiveness of our quality assessment model.

\section{{Limitation}}

Although Light-VQA+ has demonstrated impressive performance on our VEC-QA dataset, there remains potential for further improvement. It is important to note that the CLIP is not specifically designed for VQA tasks. In general, modifying CLIP through fine-tuning could enhance its effectiveness for this specific application. 
However, the lack of suitable datasets for the quality assessment task hinders us from fine-tuning the CLIP. In the future, we plan to collect a specific dataset that can support the CLIP fine-tuning.

\section{{Conclusion}}

In this paper, we focus on the issue of evaluating the quality of
VEC algorithms. To facilitate our work, we construct a VEC-QA dataset containing 4,518 videos that feature diverse content with various brightness levels. Further, we propose an effective VQA model named Light-VQA+. Concretely, we integrates the exposure-sensitive CLIP-captured features into both spatial and temporal features. After the feature fusion, we perform a quality regression to obtain assessment scores for each video clip. Finally, we apply a weighted average on these results with the HVS weights to obtain the final quality assessment score. Our experimental results confirm the effectiveness of Light-VQA+ as a video quality assessment model for exposure correction. Besides, the Light-VQA+ is capable of boosting the performance of video exposure correction methods. Therefore, we envision that the Light-VQA+ would be widely considered while developing them.\\

\noindent\textbf{Data Availability.} The datasets used for experiments during the current study are all publicly available. The KoNViD-1k is available at \url{https://database.mmsp-kn.de/konvid-1k-database.html}. The VDPVE dataset is available at \url{https://github.com/YixuanGao98/VDPVE-VQA-Dataset-for-Perceptual-Video-Enhancement}. The YouTube-UGC dataset is available at \url{https://media.withyoutube.com/}. The LIVE-VQC is available at \url{https://live.ece.utexas.edu/research/LIVEVQC/index.html}. The SICE dataset is available at \url{https://github.com/KevinJ-Huang/FECNet}.\\

\noindent\textbf{Acknowledgement.} The work was supported in part by the National Natural Science Foundation of China under Grant 62301310, and in part by the Shanghai Pujiang Program under Grant 22PJ1406800.

\bibliography{main}

\end{document}